\documentclass[10pt,twocolumn,letterpaper]{article}

\usepackage{cvpr}
\usepackage{times}
\usepackage{epsfig}
\usepackage{graphicx}
\usepackage{amsmath}
\usepackage{amssymb}

\usepackage{booktabs}
\usepackage{pifont} %
\usepackage{subcaption}
\usepackage{algorithm,algorithmicx,algpseudocode}
\usepackage{multirow}

\makeatletter
\newcommand{\settitle}{
    \twocolumn[\@maketitle]}
\@namedef{ver@everyshi.sty}{}
\makeatother

\usepackage{pgfplots} %
\usepackage{pgfplotstable} %
\pgfplotsset{compat=newest} %
\usetikzlibrary{plotmarks} %

\usepackage[space]{grffile}

\usepackage[pagebackref=true,breaklinks=true,letterpaper=true,colorlinks,bookmarks=false]{hyperref}

\cvprfinalcopy %

\def\cvprPaperID{1528} %

\newcommand{\PAR}[1]{\vskip1pt \noindent {\bf #1~}}
\newcommand{\PARbegin}[1]{\noindent {\bf #1~}}

\ifcvprfinal\fi
\begin{document}

\title{\vspace{-4pt}Siam R-CNN: Visual Tracking by Re-Detection\vspace{-4pt}}

\addtocounter{footnote}{1}
\author{Paul Voigtlaender$^{1}$ \hspace{20pt} Jonathon Luiten$^{1,2,}$\thanks{Work performed both while at the RWTH Aachen University and on a research visit at the University of Oxford.}  \hspace{20pt} Philip H.S. Torr$^{2}$ \hspace{20pt} Bastian Leibe$^{1}$\\
$^{1}$RWTH Aachen University \hspace{20pt} $^{2}$University of Oxford\\
{\tt\small \{voigtlaender,luiten,leibe\}@vision.rwth-aachen.de \hspace{20pt} phst@robots.ox.ac.uk} \vspace{0pt}
}

\maketitle

\thispagestyle{empty}

\begin{abstract}
\vspace{-4pt}

We present Siam R-CNN, a Siamese re-detection architecture which unleashes the full power of two-stage object detection approaches for visual object tracking. We combine this with a novel tracklet-based dynamic programming algorithm, which takes advantage of re-detections of both the first-frame template and previous-frame predictions, to model the full history of both the object to be tracked and potential distractor objects.
This enables our approach to make better tracking decisions, as well as to re-detect tracked objects after long occlusion.
Finally, we propose a novel hard example mining strategy to improve Siam R-CNN's robustness to similar looking objects. Siam R-CNN achieves the current best performance on ten tracking benchmarks, with especially strong results for long-term tracking. We make our code and models available at \url{www.vision.rwth-aachen.de/page/siamrcnn}.
\vspace{-4pt}

\end{abstract}

\vspace{-10pt}
\section{Introduction}

We approach Visual Object Tracking using the paradigm of Tracking by Re-Detection. We present a powerful novel re-detector, Siam R-CNN, an adaptation of Faster R-CNN \cite{Ren15NIPS} with a Siamese architecture, which re-detects a template object anywhere in an image by determining if a region proposal is the same object as a template region, and regressing the bounding box for this object.  %
Siam R-CNN is robust against changes in object size and aspect ratio as the proposals are aligned to the same size, which is in contrast to the popular cross-correlation-based methods \cite{Li18CVPRSiamRPN}.

\begin{figure}[t!]
\centering
\includegraphics[width=\linewidth]{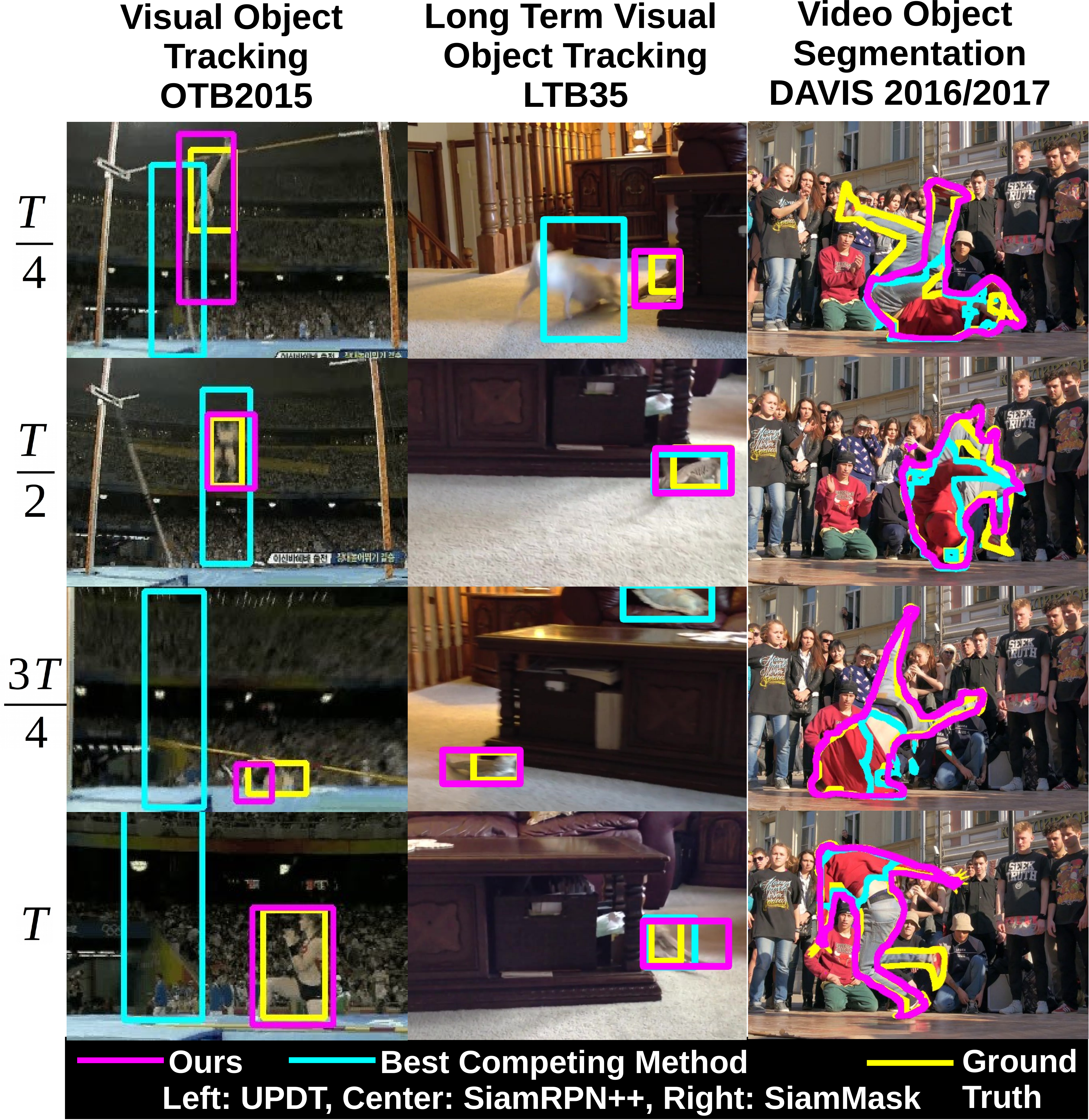}
\caption{Example results of Siam R-CNN on 3 different tracking tasks where it obtains new state-of-the-art results.} %
\label{fig:teaser}
\end{figure}

Tracking by re-detection has a long history, reaching back to the seminal work of Avidan~\cite{Avidan04PAMI} and Grabner \etal~\cite{Grabner06BMVC}.
Re-detection is challenging due to the existence of distractor objects that are very similar to the template object. In the past, the problem of distractors has mainly been approached by strong spatial priors from previous predictions \cite{Bertinetto2016ECCV,Li18CVPRSiamRPN, Li19CVPR}, or by online adaptation \cite{Avidan04PAMI, Grabner06BMVC, Babenko11PAMI, Saffari10CVPR, Hare15PAMI, Saffari09ICCVW, TLD}. Both of these strategies are prone to drift.

We instead approach the problem of distractors by making two novel contributions beyond our Siam R-CNN re-detector design. Firstly we introduce a novel hard example mining procedure which trains our re-detector specifically for difficult distractors.
Secondly we propose a novel Tracklet Dynamic Programming Algorithm (TDPA) which simultaneously tracks all potential objects, including distractor objects, by re-detecting all object candidate boxes from the previous frame, and grouping boxes over time into tracklets (short object tracks). It then uses dynamic programming to select the best object in the current timestep based on the complete history of all target object and distractor object tracklets. %
By explicitly modeling the motion and interaction of all potential objects and pooling similarity information from detections grouped into tracklets, Siam R-CNN is able to effectively perform long-term tracking, while being resistant to tracker drift, and being able to immediately re-detect objects after disappearance.
Our TDPA requires only a small set of new re-detections in each timestep, updating its tracking history iteratively online. This allows Siam R-CNN to run at 4.7 frames per second (FPS) and its speed-optimized variant to run at more than 15 FPS.%

We present evaluation results on a large number of datasets. Siam R-CNN outperforms all previous methods on six short-term tracking benchmarks %
 as well as on four long-term tracking benchmarks%
 , where it achieves especially strong results, up to $10$ percentage points higher than previous methods. By obtaining segmentation masks using an off-the-shelf box-to-segmentation network, Siam R-CNN also outperforms all previous Video Object Segmentation methods that only use the first-frame bounding box (without the mask) on four recent VOS benchmarks.

\section{Related Work}

\PARbegin{Visual Object Tracking (VOT).}
VOT is the task of tracking an object through a video given the first-frame bounding box of the object. VOT is commonly evaluated on benchmarks such as OTB \cite{Wu13CVPR,Wu15TPAMI}, VOT \cite{Kristan16TPAMI,Kristan18ECCVW}, and many more \cite{Muller18ECCV,Huang18Arxiv,Zhu18Arxiv,Mueller16ECCV,Galoogahi17ICCV}. Recently a number of long-term tracking benchmarks have been proposed \cite{lukevzivc2018now,Valmadre18ECCV,Fan19CVPRLASOT} which extend VOT to a more difficult and realistic setting, where objects must be tracked over many frames, with objects disappearing and reappearing.

Many classical methods use an online learned classifier to re-detect the object of interest over the full image \cite{Avidan04PAMI, Grabner06BMVC, Babenko11PAMI, Saffari10CVPR, Hare15PAMI, Saffari09ICCVW, TLD}. In contrast, Siam R-CNN learns the expected appearance variations by offline training instead of learning a classifier online.

Like our Siam R-CNN, many recent methods approach VOT using Siamese architectures. Siamese region proposal networks (SiamRPN \cite{Li18CVPRSiamRPN}) use a single-stage RPN \cite{Ren15NIPS} detector adapted to re-detect a template by cross-correlating the deep template features with the deep features of the current frame. Here, single-stage means directly classifying anchor boxes \cite{Liu16ECCV} which is in contrast to two-stage architectures \cite{Ren15NIPS} which first generate proposals, and then  align their features and classify them in the second stage.

Recent tracking approaches improve upon SiamRPN, making it distractor aware (DaSiamRPN \cite{Zhu18ECCV}), adding a cascade (C-RPN \cite{Fan19CVPRCRPN}), producing masks (SiamMask \cite{Wang19CVPR}), using deeper architectures (SiamRPN+ \cite{Zhang19CVPR} and SiamRPN++ \cite{Li19CVPR}) and maintaining a set of diverse templates (THOR \cite{Sauer19BMVC}). These (and many more \cite{Bolme10CVPR,Henriques15TPAMI,Ma15ICCV}) only search for the object within a small window of the previous prediction. %
DiMP \cite{Bhat19ICCV} follows this paradigm while meta-learning a robust target and background appearance model.

Other recent developments in VOT include using domain specific layers with online learning \cite{mdnet}, learning an adaptive spatial filter regularizer \cite{Dai19CVPR}, exploiting category-specific semantic information \cite{Tripathi19BMVC}, using continuous \cite{Danelljan16ECCV} or factorized \cite{Danelljan17CVPR} convolutions, and achieving accurate bounding box predictions using an overlap prediction network \cite{Danelljan19CVPR}. Huang \etal~\cite{Huang19ICCV} propose a framework to convert any detector into a tracker. Like Siam R-CNN, they also apply two-stage architectures, but their method relies on meta-learning and it achieves a much lower accuracy.

Long-term tracking is mainly addressed by enlarging the search window of these Siamese trackers when the detection confidence is low \cite{Zhu18ECCV, Li19CVPR}.
In contrast, we use a two-stage Siamese re-detector which searches over the whole image, producing stronger results %
across many benchmarks.%

\PAR{Video Object Segmentation (VOS).}
VOS is an extension of VOT where a set of template segmentation masks are given, and segmentation masks need to be produced in each frame.
Many methods perform fine-tuning on the template masks \cite{OSVOS,Maninis18TPAMI,voigtlaender17BMVC,Li18ECCV,Bao18CVPR,Luiten18ACCV}, which leads to strong results but is slow. Recently, several methods have used the first-frame masks without fine-tuning \cite{Chen18CVPR,Yang18CVPR,Cheng18CVPR,Hu18ECCV,Oh18CVPR,Xu18ECCV,Voigtlaender19CVPR,Oh19ICCV}, running faster but often not performing as well.

Very few methods \cite{Wang19CVPR,Yeo17CVPR} tackle the harder problem of producing mask tracking results while only using the given template bounding box and not the mask. We adapt our method to perform VOS in this setting by using a second network to produce masks for our box tracking results.

\begin{figure*}[t]
\centering
\includegraphics[width=\linewidth]{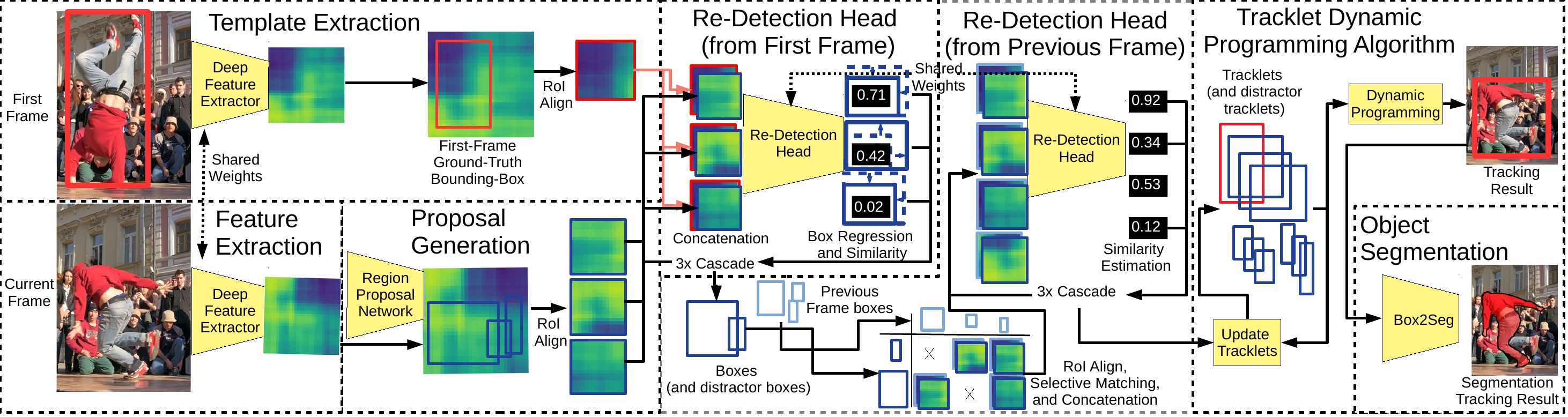}
\caption{Overview of Siam R-CNN. A Siamese R-CNN provides re-detections of the object given in the first-frame bounding box, which are used by our Tracklet Dynamic Programming Algorithm along with re-detections from the previous frame. The results are bounding box level tracks which can be converted to segmentation masks by the Box2Seg network.}
\label{fig:overview}
\end{figure*}

\section{Method}
Inspired by the success of Siamese trackers \cite{Kristan18ECCVW,Wu15TPAMI,Kristan16TPAMI}, we use a Siamese  architecture for our re-detector.
 Many recent trackers \cite{Zhu18ECCV,Wang19CVPR,Li19CVPR,Li18CVPRSiamRPN,Bhat19ICCV} adopt a single-stage detector architecture. %
 For the task of single-image object detection, two-stage detector networks such as Faster R-CNN \cite{Ren15NIPS} have been shown to outperform single-stage detectors. Inspired by this, we design our tracker as a Siamese two-stage detection network. %
 The second stage can directly compare a proposed Region of Interest (RoI) to a template region by concatenating their RoI aligned features. By aligning proposals and reference to the same size, Siam R-CNN achieves robustness against changes in object size and aspect ratio, which is hard to achieve when using the popular cross-correlation operation \cite{Li18CVPRSiamRPN}.
 Fig.~\ref{fig:overview} shows an overview of Siam R-CNN including the Tracklet Dynamic Programming Algorithm (TDPA).

\subsection{Siam R-CNN}
Siam R-CNN is a Siamese re-detector based on a two-stage detection architecture. %
Specifically, we take a Faster R-CNN network that has been pre-trained on the COCO \cite{coco} dataset for detecting 80 object classes. This network consists of a backbone feature extractor followed by two detection stages; first a category-agnostic RPN, followed by a category-specific detection head. We fix the weights of the backbone and the RPN and replace the category-specific detection head with our re-detection head.

We create input features for the re-detection head for each region proposed by the RPN by performing RoI Align \cite{He17ICCV} to extract deep features from this proposed region. We also take the RoI Aligned deep features of the initialization bounding box in the first frame, and then concatenate these together and feed the combined features into a $1\times1$ convolution which reduces the number of features channels back down by half.
These joined features are then fed into the re-detection head with two output classes; the proposed region is either the reference object or it is not.
Our re-detection head uses a three-stage cascade \cite{Cai18CVPR} without shared weights. The structure of the re-detection head is the same as the structure of the detection head of Faster R-CNN, except for using only two classes and for the way the input features for the re-detection head are created by concatenation.
The backbone and RPN are frozen %
 and only the re-detection head (after concatenation) is trained for tracking, using pairs of frames from video datasets. Here, an object in one frame is used as reference and the network is trained to re-detect the same object in another frame.

\subsection{Video Hard Example Mining}
\label{sec:hardexample}
During conventional Faster R-CNN training, the negative examples for the second stage are sampled from the regions proposed by the RPN in the target image. However, in many images there are only few relevant negative examples. In order to maximize the discriminative power of the re-detection head, we need to train it on hard negative examples. %
Mining hard examples for detection has been explored in previous works (\eg \cite{Felzenszwalb10PAMI,Shrivastava16CVPR}). %
 However, rather than finding general hard examples for detection, we find hard examples for re-detection conditioned on the reference object by retrieving objects from other videos.

\begin{figure}[t]
\begin{center}
\begin{subfigure}{0.47\columnwidth}
\includegraphics[width=\textwidth]{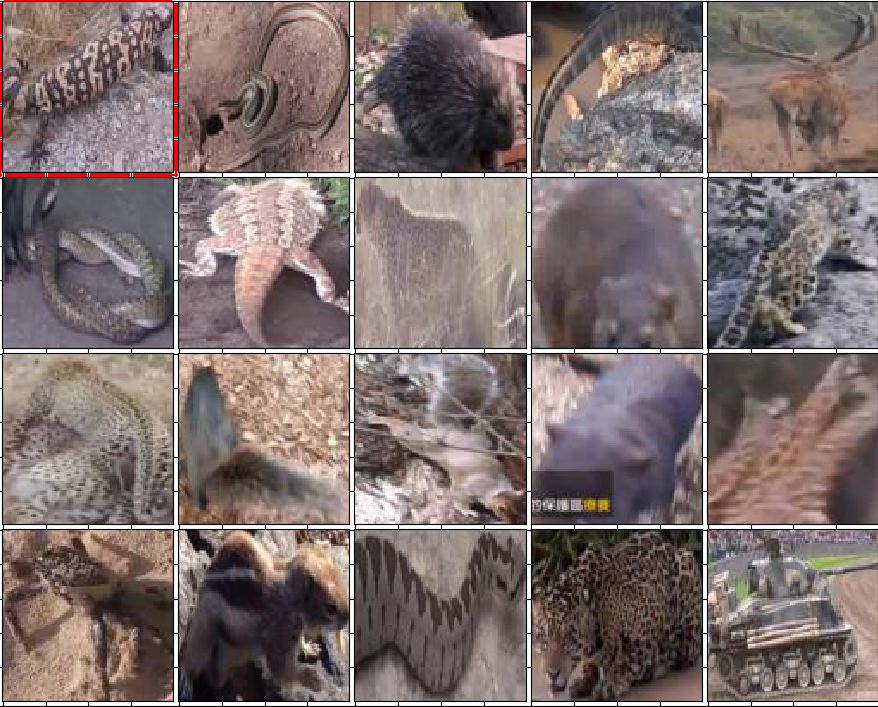}
\end{subfigure}
\hfill
\begin{subfigure}{0.47\columnwidth}
\includegraphics[width=\textwidth]{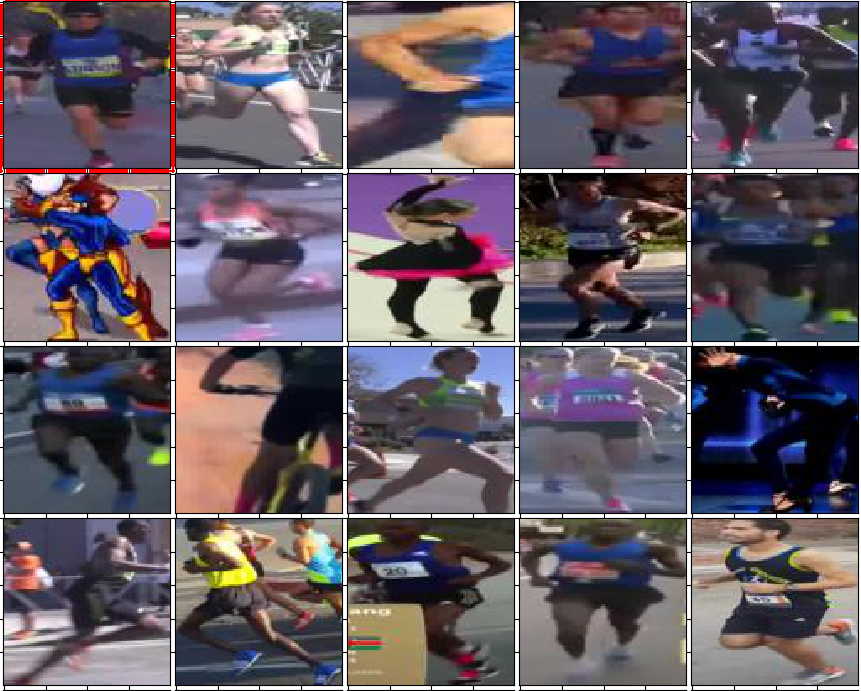}
\end{subfigure}
\end{center}
\vspace{-4mm}
\caption{\label{fig:hard-example-mining-example}Hard negative mining examples retrieved from other videos for the reference objects shown in red. %
}
\end{figure}

\PAR{Embedding Network.} A straightforward approach to selecting relevant videos from which to get hard negative examples for the current video, is taking videos in which an object has the same class as the current object \cite{Zhu18ECCV}. However, object class labels are not always available, and some objects of the same class could be easy to distinguish, while some objects of different classes could also be potentially hard negatives. Hence, we propose to use an embedding network, inspired by person re-identification, which extracts an embedding vector for every ground truth bounding box which represents the appearance of that object. We use the network from PReMVOS \cite{Luiten18ACCV}, which is trained with batch-hard triplet loss \cite{Hermans17Arxiv} to separate classes on COCO before being trained on YouTube-VOS to disambiguate between individual object instances. \Eg, two distinct persons should be far away in the embedding space, while two crops of the same person in different frames should be close.%

\PAR{Index Structure.} We next create an efficient indexing structure for approximate nearest neighbor queries (see supplemental material) %
 and use it to find nearest neighbors of the tracked object in the embedding space. %
 Fig.~\ref{fig:hard-example-mining-example} shows examples of the retrieved hard negative examples. As can be seen, most of the negative examples are very relevant and hard.

\PAR{Training Procedure.}
Evaluating the backbone on-the-fly on other videos to retrieve hard negative examples for the current video frame would be very costly. Instead, we pre-compute the RoI-aligned features for every ground truth box of the training data. %
 For each training step, as usual, a random video and object in this video is selected and then a random reference and a random target frame. Afterwards, we use the indexing structure to retrieve for the reference box  the 10,000 nearest neighbor bounding boxes from other videos and sample 100 of them as additional negative training examples. %
More details about video hard example mining can be found in the supplemental material.

\subsection{Tracklet Dynamic Programming Algorithm}
\label{subsec:trackingalg}
Our Tracklet Dynamic Programming Algorithm (TDPA) implicitly tracks both the object of interest and potential similar-looking distractors using spatio-temporal cues. In this way, distractors can be consistently suppressed, which would not be possible using only visual similarity.
To this end, TDPA maintains a set of tracklets, \ie, short sequences of detections which almost certainly belong to the same object. It then uses a dynamic programming based scoring algorithm to select the most likely sequence of tracklets for the template object between the first and the current frame.

Each detection is part of exactly one tracklet and it is defined by a bounding box, a re-detection score, and its RoI-aligned features. A tracklet %
 is defined by a set of detections, exactly one for each time step from its start to its end time.%

\begin{algorithm}[t]
\small
\algnewcommand{\LineComment}[1]{\State \(\triangleright\) #1}
\caption{\label{alg:tracklet}Update $\mathrm{tracklets}$ for one time-step $t$}
\begin{algorithmic}[1]
\State \textbf{Inputs} $\mathrm{ff\_gt\_feats, tracklets, image_t, dets_{t-1}}$
\State $\mathrm{backbone\_feats} \gets \mathrm{backbone}(\mathrm{image}_t)$ \label{trackletalg:backbone}
\State $\mathrm{RoIs} \gets \mathrm{RPN}(\mathrm{backbone\_feats}) \cup \mathrm{dets}_{t-1}$ \label{trackletalg:rois}
\State $\mathrm{dets}_t \gets \mathrm{redetection\_head}(\mathrm{RoIs}, \mathrm{ff\_gt\_feats})$ \label{trackletalg:detect}
\LineComment{scores are set to $-\infty$ if spatial distance is $>\gamma$}
\State $\mathrm{scores} \gets \mathrm{score\_pairwise\_redetection}(\mathrm{dets}_t, \mathrm{dets}_{t-1}, \gamma)$ \label{trackletalg:scores}
\For{$d_t \in \mathrm{dets}_t$} \label{trackletalg:matching-start}
  \State $s_1 \gets \max_{d_{t-1} \in \mathrm{dets}_{t-1}} \mathrm{scores}[d_{t}, d_{t-1}]$
  \State $\hat{d}_{t-1} \gets \arg\!\max_{d_{t-1} \in \mathrm{dets}_{t-1}} \mathrm{scores}[d_{t}, d_{t-1}]$
  \LineComment{Max score of all other current detections}
  \State $s_2 \gets \max_{\tilde{d}_{t} \in \mathrm{dets}_{t}\setminus \{d_{t}\}} \mathrm{scores}[\tilde{d}_{t}, \hat{d}_{t-1}]$
  \LineComment{Max score of all other previous detections}
  \State $s_3 \gets \max_{d_{t-1} \in \mathrm{dets}_{t-1}\setminus \{\hat{d}_{t-1}\}} \mathrm{scores}[d_{t}, d_{t-1}]$
  \If{$s_1 > \alpha \land s_2 \leq s_1 - \beta \land s_3 \leq s_1 - \beta$}
      \State $\mathrm{tracklet}(\hat{d}_{t-1}).\mathrm{append}(d_t) \ \ \triangleright \text{Extend tracklet}$
  \Else{\hspace{3mm}\(\triangleright\) Start new tracklet}
      \State $\mathrm{tracklets} \gets \mathrm{tracklets} \cup \{\{d_t\}\}$
  \EndIf
\EndFor \label{trackletalg:matching-end}
\end{algorithmic}
\end{algorithm}

\PAR{Tracklet Building.}
We extract the RoI aligned features for the first-frame ground truth bounding box ($\mathrm{ff\_gt\_feats}$) and initialize a tracklet consisting of just this box. For each new frame, we update the set of tracklets as follows (\cf Algorithm \ref{alg:tracklet}):
We extract backbone features of the current frame and evaluate the region proposal network (RPN) to get regions of interest (RoIs, lines~\ref{trackletalg:backbone}--\ref{trackletalg:rois}). To compensate for potential RPN false negatives, the set of RoIs is extended by the bounding box outputs from the previous frame. We run the re-detection head (including bounding box regression) on these RoIs to produce a set of re-detections of the first-frame template (line~\ref{trackletalg:detect}).
Afterwards, we re-run the classification part of the re-detection head (line~\ref{trackletalg:scores}) on the current detections $\mathrm{dets}_{t}$, but this time with the detections $\mathrm{dets}_{t-1}$ from the previous frame as reference instead of the first-frame ground truth box, to calculate similarity scores ($\mathrm{scores}$) between each pair of detections.

To measure the spatial distance of two detections, we represent their bounding boxes by their center coordinates $x$ and $y$, and their width $w$ and height $h$, of which $x$ and $w$ are normalized with the image width, and $y$ and $h$ are normalized with the image height, so that all values are between $0$ and $1$. The spatial distance between two bounding boxes $(x_1,y_1,w_1,h_1)$ and $(x_2,y_2,w_2,h_2)$ is then given by the $L_{\infty}$ norm%
, \ie, $\max(|x_1-x_2|, |y_1-y_2|, |w_1-w_2|, |h_1-h_2|)$. In order to save computation and to avoid false matches, we calculate the pairwise similarity scores only for pairs of detections where this spatial distance is less than $\gamma$ and set the similarity score to $-\infty$ otherwise.

We extend the tracklets from the previous frame by the current frame detections (lines~\ref{trackletalg:matching-start}--\ref{trackletalg:matching-end}) when the similarity score to a new detection is high ($>\!\!\alpha$) and there is no ambiguity, \ie, there is no other detection which has an almost as high similarity (less than $\beta$ margin) with that tracklet, and there is no other tracklet which has an almost as high similarity (less than $\beta$ margin) with that detection.
Whenever there is any ambiguity, we start a new tracklet which initially consists of a single detection. The ambiguities will then be resolved in the tracklet scoring step.

\PAR{Scoring.}
A track $A=(a_1,\dots,a_N)$ is a sequence of $N$ non-overlapping tracklets, \ie, $\mathrm{end}(a_i)<\mathrm{start}(a_{i+1})\  \forall i\in\{1,\dots,N-1\}$, where $\mathrm{start}$ and $\mathrm{end}$ denote the start and end times of a tracklet, respectively.
The total score of a track consists of a unary score measuring the quality of the individual tracklets, and of a location score which penalizes spatial jumps between tracklets, \ie
\vspace{-4pt}
\begin{align}
\label{eq:track-score}
\mathrm{score}(A)=&\sum_{i=1}^{N}\mathrm{unary}(a_{i})+\sum_{i=1}^{N-1}w_\mathrm{loc}\mathrm{loc\_score}(a_{i},a_{i+1}).
\\
\mathrm{unary}(a_{i}) =&\sum_{t=\mathrm{start}(a_{i})}^{\mathrm{end}(a_{i})}w_{\mathrm{ff}}\mathrm{ff\_score}(a_{i,t})\\\nonumber
 &+(1-w_{\mathrm{ff}})\mathrm{ff\_tracklet\_score}(a_{i,t}),
\end{align}
where $\mathrm{ff\_score}$ denotes the re-detection confidence for the detection $a_{i,t}$ of tracklet $a_i$ at time $t$ from the re-detection head using the first-frame ground truth bounding box as reference. There is always one tracklet which contains the first-frame ground truth bounding box, which we denote as the first-frame tracklet $a_{\mathrm{ff}}$. All detections in a tracklet have a very high chance of being a correct continuation of the initial detection of this tracklet, because in cases of ambiguities tracklets are terminated. Hence, the most recent detection of the first-frame tracklet is also the most recent observation that is almost certain to be the correct object. Thus, we use this
 detection
 as an additional reference for re-detection producing a score denoted by $\mathrm{ff\_tracklet\_score}$ which is linearly combined with the $\mathrm{ff\_score}$.

The location score between two tracklets $a_i$ and $a_j$ is given by the negative $L_1$ norm of the difference between the bounding box $(x,y,w,h)$ of the last detection of $a_i$ and the bounding box of the first detection of $a_j$, \ie
\vspace{-4pt}
\begin{eqnarray}
\mathrm{loc\_score}(a_{i},a_{j})=-|\mathrm{end\_bbox}(a_{i})-\mathrm{start\_bbox}(a_{j})|_{1}. \nonumber
\\[-15pt]\nonumber
\end{eqnarray}

\begin{figure}[t]
\scalebox{1.02}{
\includegraphics[width=0.49\columnwidth]{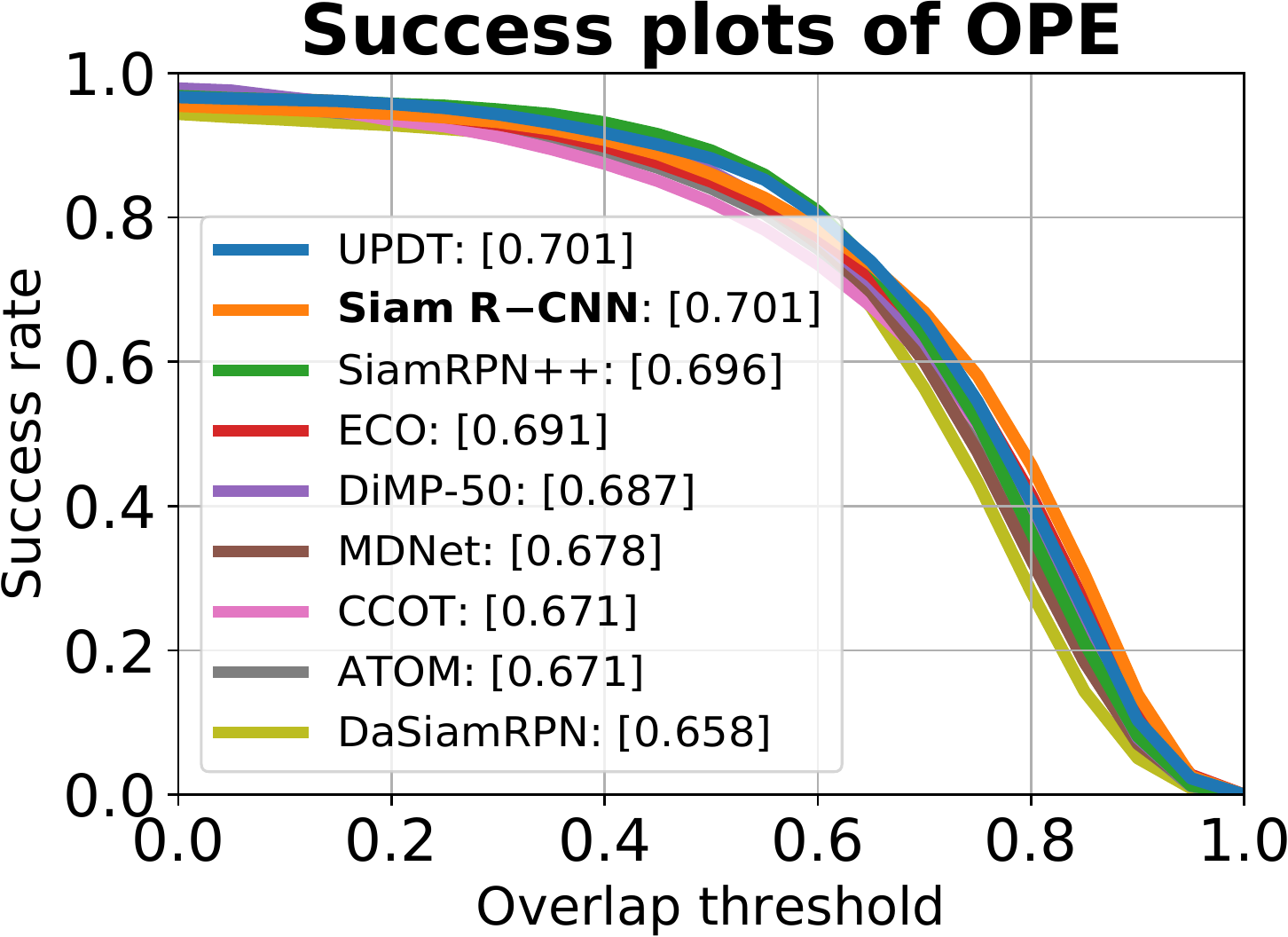}
\includegraphics[width=0.49\columnwidth]{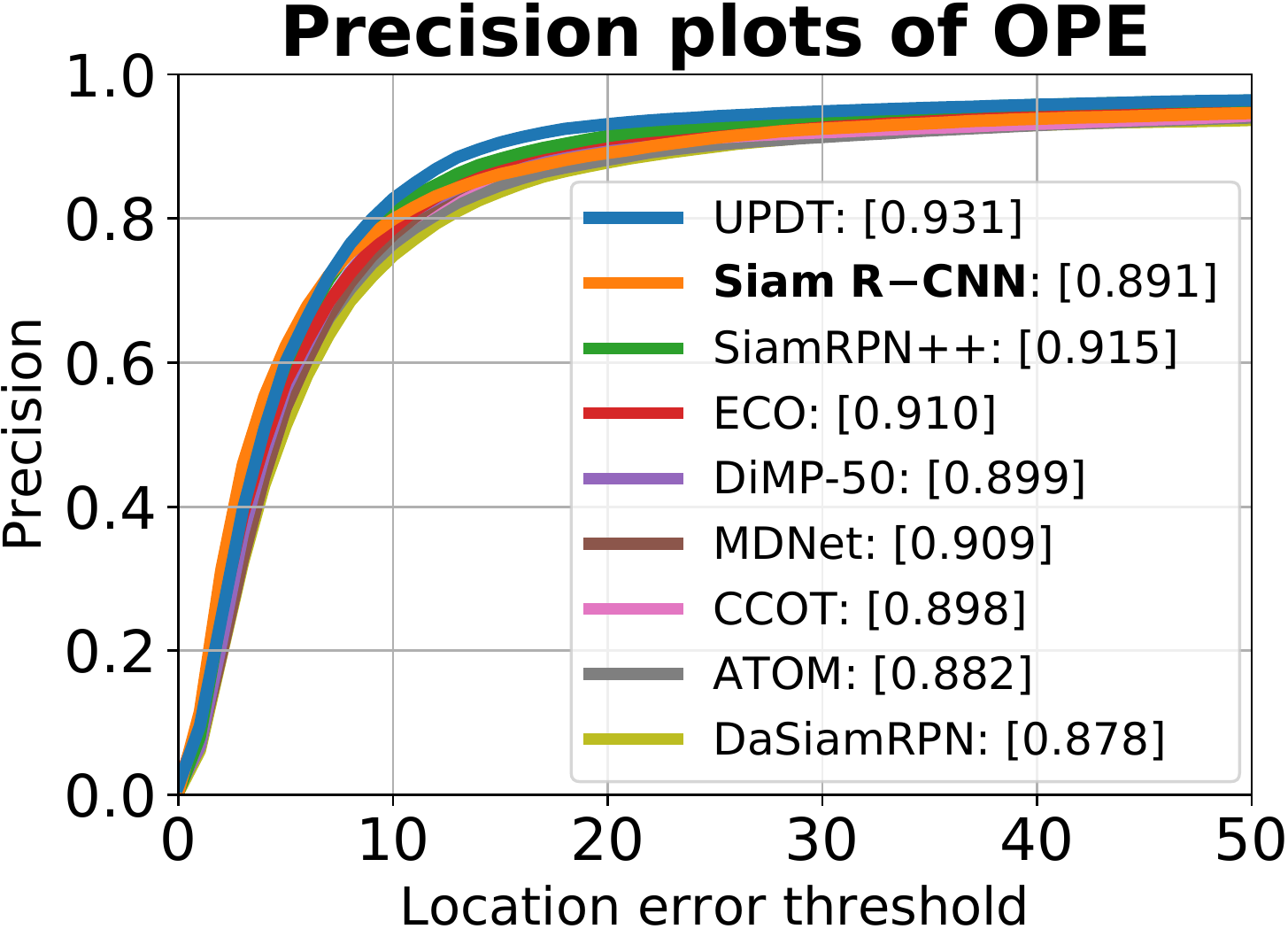}
}
\caption{Results on OTB2015 \cite{Wu15TPAMI}.}
\label{fig:otb-res}
\end{figure}

\PARbegin{Online Dynamic Programming.}
We efficiently find the sequence of tracklets with the maximum total score (Eq.~\ref{eq:track-score}) by maintaining an array $\theta$ which for each tracklet $a$ stores the total score $\theta[a]$ of the optimal sequence of tracklets which starts with the first-frame tracklet and ends with $a$.

Once a tracklet is not extended, it is terminated. Thus, for each new frame only the scores for tracklets which have been extended or newly created need to be newly computed.
For a new time-step, first we set $\theta[a_{\mathit{ff}}]=0$ for the first-frame tracklet $a_{\mathit{ff}}$, since all tracks have to start with that tracklet. Afterwards, for every tracklet $a$ which has been updated or newly created, $\theta[a]$ is calculated as
\vspace{-4pt}
\begin{eqnarray}
  \theta[a] = \mathrm{unary}(a) +\!\!\!\! \max_{\tilde{a}: \mathrm{end}(\tilde{a})<\mathrm{start}(a)} \!\!\!\theta[\tilde{a}] + w_\mathrm{loc} \mathrm{loc\_score}(\tilde{a}, a). \nonumber
 \\[-14pt]\nonumber
\end{eqnarray}
To retain efficiency for very long sequences, we allow a maximum temporal gap between two tracklets of $1500$ frames, %
 which is long enough for most applications.%

After updating $\theta$ for the current frame, we select the tracklet $\hat{a}$ with the highest dynamic programming score, \ie
  $\hat{a} = \arg\max_{a} \theta[a]$.
If the selected tracklet does not contain a detection in the current frame, then our algorithm has indicated that the object is not present. For benchmarks that require a prediction in every frame we use the most recent box from the selected tracklet, and assign it a score of $0$.

\subsection{Box2Seg}
To produce segmentation masks for the VOS task, we use an off-the-shelf bounding-box-to-segmentation (Box2Seg) network from PReMVOS~\cite{Luiten18ACCV}. Box2Seg is a fully convolutional DeepLabV3+ \cite{Chen18ECCV} network with an Xception-65 \cite{CholletCVPR17} backbone. It has been trained on Mapillary \cite{Neuhold17ICCV} and COCO \cite{coco} to output the mask for a bounding box crop. Box2Seg is fast, running it after tracking only requires 0.025 seconds per object per frame.
We combine overlapping masks such that masks with less pixels are on top.

\subsection{Training Details}
Siam R-CNN is built upon the Faster R-CNN \cite{Ren15NIPS} implementation %
 from \cite{Wu16tensorpack}, with a ResNet-101-FPN backbone \cite{resnet, Lin17CVPR}, group normalization \cite{Wu18ECCV} and cascade \cite{Cai18CVPR}. It has been pre-trained from scratch \cite{He19ICCV} on COCO \cite{coco}.
Except where specified otherwise, we train Siam R-CNN on the training sets of multiple tracking datasets simultaneously:
 ImageNet VID \cite{imagenet} (4000 videos), YouTube-VOS 2018  \cite{Xu18ECCV} (3471 videos), GOT-10k \cite{Huang18Arxiv} (9335 videos) and LaSOT \cite{Fan19CVPRLASOT} (1120 videos). In total, we use 18k videos and 119k static images from COCO, which is a significant amount of data, but it is actually less than what previous methods used, \eg SiamRPN++ uses 384k videos and 1867k static images.
More details about the amount of training data are in the supplemental material.

During training, we use motion blur \cite{Zhu18ECCV}, grayscale, gamma, flip, and scale augmentations.

\begin{figure}[t]
\centering
\includegraphics[width=0.49\columnwidth]{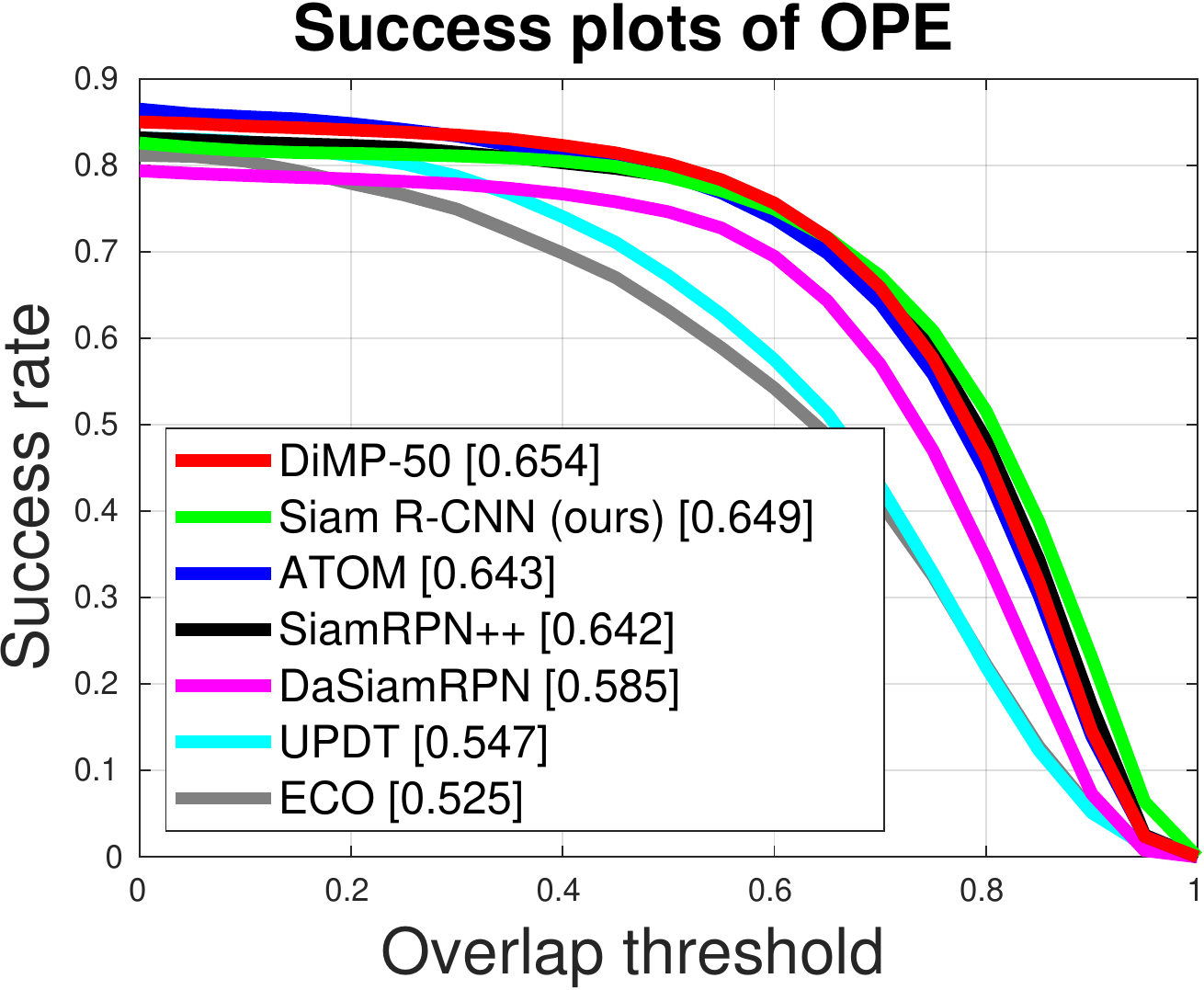}
\includegraphics[width=0.49\columnwidth]{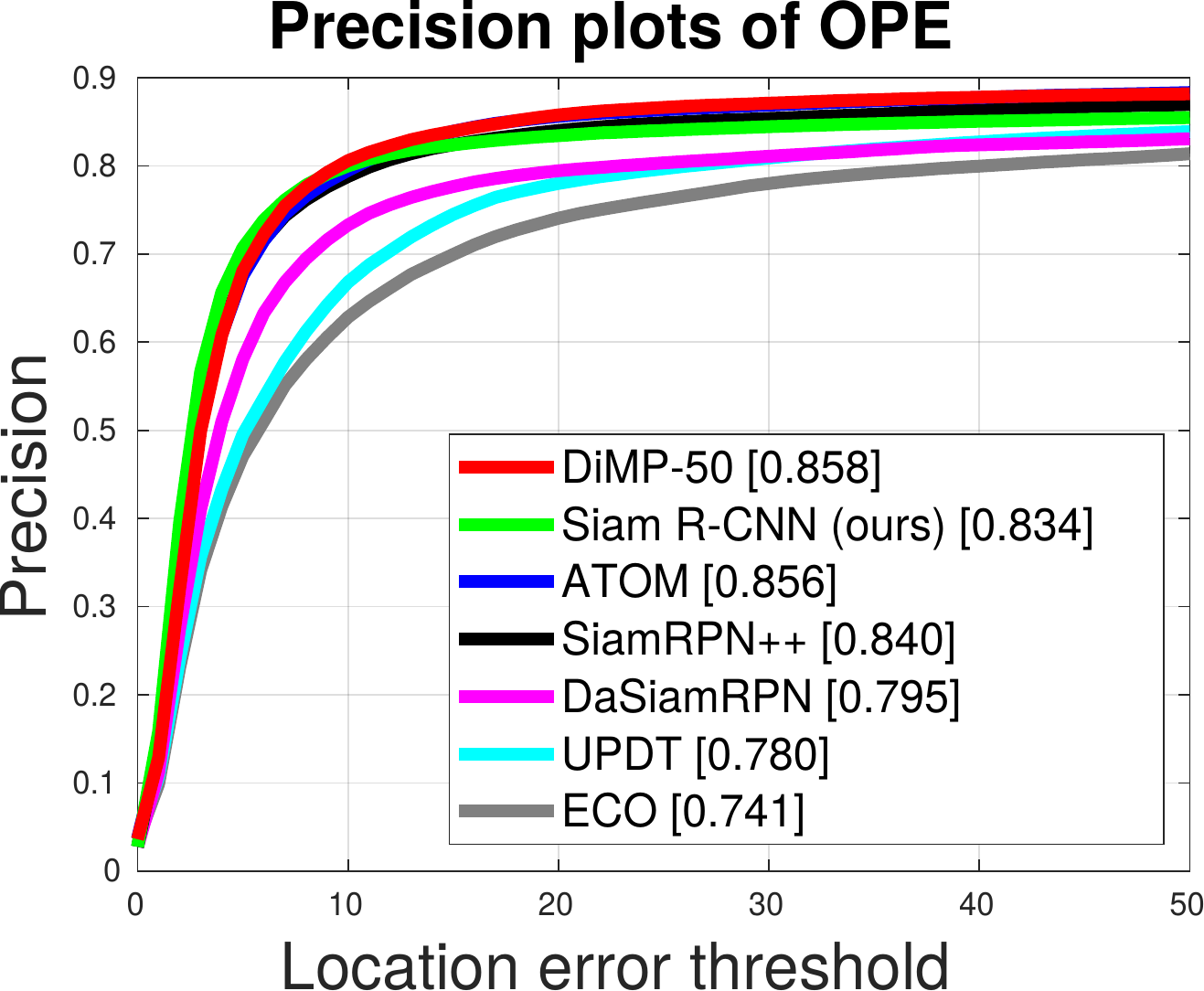}
\caption{Results on UAV123 \cite{Mueller16ECCV}.}
\label{fig:uav123-res}
\end{figure}

\section{Experiments}

We evaluate Siam R-CNN for standard visual object tracking, for long-term tracking, and on VOS benchmarks.
We tune a single set of hyper-parameters for our Tracklet Dynamic Programming Algorithm (\cf Section \ref{subsec:trackingalg}) on the DAVIS 2017 training set, as this is a training set that we did not use to train our re-detector. We present results using these hyper-parameters on all benchmarks, rather than tuning the parameters separately for each one. %

\subsection{Short-Term Visual Object Tracking Evaluation}
We evaluate short-term VOT on six benchmarks, and on five further benchmarks in the supplemental material.

\PAR{OTB2015.} We evaluate on OTB2015 \cite{Wu15TPAMI} (100 videos, 590 frames average length), calculating the success and precision over varying overlap thresholds. Methods are ranked by the area under the success curve (AUC). Fig.~\ref{fig:otb-res} compares our results to eight state-of-the-art (SOTA) trackers \cite{Bhat18ECCV, Li19CVPR, Danelljan17CVPR, Bhat19ICCV, mdnet, Danelljan16ECCV, Danelljan19CVPR, Zhu18ECCV}.
Siam R-CNN achieves $70.1\%$ AUC, which equals the previous best result by UPDT \cite{Bhat18ECCV}.

\PAR{UAV123.} Fig.~\ref{fig:uav123-res} shows our results on UAV123 \cite{Mueller16ECCV} (123 videos, 915 frames average length) on the same metrics as OTB2015 compared to six SOTA approaches \cite{Bhat19ICCV, Danelljan19CVPR, Li19CVPR, Zhu18ECCV, Bhat18ECCV, Danelljan17CVPR}. We achieve an AUC of $64.9\%$, which is close to the previous best result of DiMP-50 \cite{Bhat19ICCV} with $65.4\%$.

\PAR{NfS.} Tab.~\ref{tab:nfs-res} shows our results on the NfS dataset \cite{Galoogahi17ICCV} (30FPS, 100 videos, 479 frames average length) compared to five SOTA approaches. Siam R-CNN achieves a success score of $63.9\%$, which is $1.9$ percentage points higher than the previous best result by DiMP-50 \cite{Bhat19ICCV}.

\PAR{TrackingNet.}
Tab.~\ref{tab:trackingnet} shows our results on the TrackingNet test set \cite{Muller18ECCV} (511 videos, 442 frames average length), compared to five SOTA approaches. Siam R-CNN achieves a success score of $81.2\%$, \ie, $7.2$ percentage points higher than the previous best result of DiMP-50 \cite{Bhat19ICCV}. In terms of precision the gap is more than $10$ percentage points.

\begin{figure}[t]
\centering
\includegraphics[width=0.6\columnwidth]{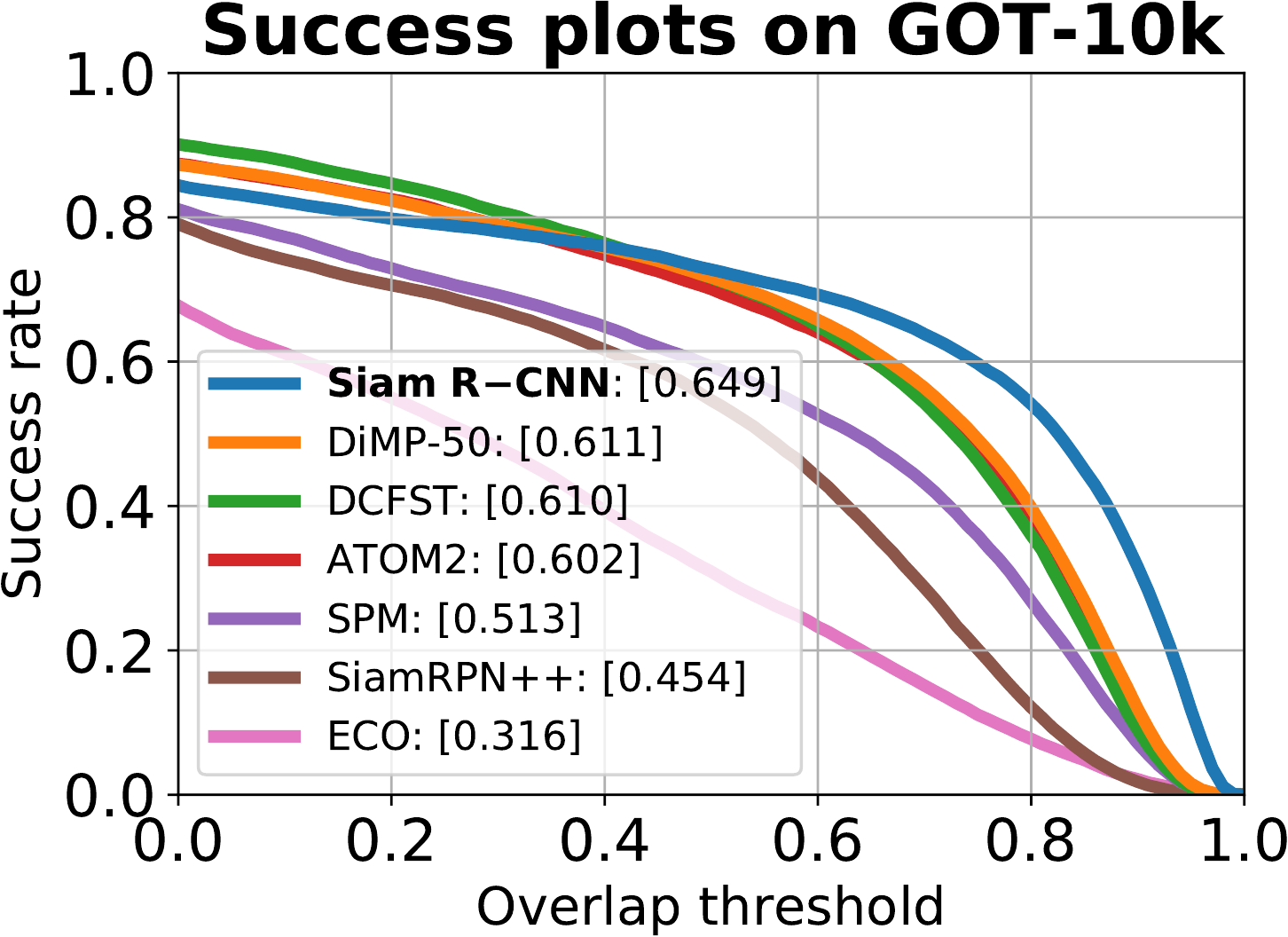}
\caption{Results on GOT-10k \cite{Huang18Arxiv}.}
\label{fig:got10k-res}
\end{figure}

\PAR{GOT-10k.} Fig.~\ref{fig:got10k-res} shows our results on the GOT-10k \cite{Huang18Arxiv} test set (180 videos, 127 frames average length) compared to six SOTA approaches \cite{Bhat19ICCV, zheng2019learning, Danelljan19CVPR, wang2019spm, Li19CVPR, Danelljan17CVPR}.
On this benchmark, methods are only allowed to use the GOT-10k training set as video data for training. Therefore we train a new model starting from COCO pre-training, and train only on GOT-10k. We achieve a success rate of $64.9\%$ which is $3.8$ percentage points higher than the previous best result from DiMP-50 \cite{Bhat19ICCV}. This shows that Siam R-CNN's advantage over all previous methods is not just due to different training data, but from the tracking approach itself.

\begin{table}[t]
\centering{}%
\footnotesize
\setlength{\tabcolsep}{2pt}
\begin{tabular}{lcccccc}
\toprule 
 & {\footnotesize{}Huang \etal} & {\footnotesize{}UPDT} & {\footnotesize{}ATOM} & {\footnotesize{}Tripathi \etal} & {\footnotesize{}DiMP-50} & {\footnotesize{}Siam}\tabularnewline
 & {\footnotesize{}\cite{Huang19ICCV}} & {\footnotesize{}\cite{Bhat18ECCV}} & {\footnotesize{}\cite{Danelljan19CVPR}} & {\footnotesize{}\cite{Tripathi19BMVC}} & {\footnotesize{}\cite{Bhat19ICCV}} & R-CNN\tabularnewline
\midrule 
{\footnotesize{}Success} & {\footnotesize{}51.5} & {\footnotesize{}53.7} & {\footnotesize{}58.4} & {\footnotesize{}60.5} & {\footnotesize{}62.0} & {\footnotesize{}\textbf{63.9}}\tabularnewline
\bottomrule
\end{tabular}
\caption{\label{tab:nfs-res}Results on NfS \cite{Bhat19ICCV}.}
\end{table}

\begin{table}[t]
\centering{}%
\setlength{\tabcolsep}{2pt}
\footnotesize
\begin{tabular}{lcccccc}
\toprule
 & {\footnotesize{}DaSiamRPN} & {\footnotesize{}UPDT} & {\footnotesize{}ATOM} & {\footnotesize{}SiamRPN++} & {\footnotesize{}DiMP-50} & {\footnotesize{}Siam}\tabularnewline
 & {\footnotesize{}\cite{Zhu18ECCV}} & {\footnotesize{}\cite{Bhat18ECCV}} & {\footnotesize{}\cite{Danelljan19CVPR}} & {\footnotesize{}\cite{Li19CVPR}} & {\footnotesize{}\cite{Bhat19ICCV}} & {\footnotesize{}R-CNN} \tabularnewline 
\midrule
Precision & {\footnotesize{}59.1} & {\footnotesize{}55.7} & {\footnotesize{}64.8} & {\footnotesize{}69.4} & {\footnotesize{}68.7} & {\footnotesize{}\textbf{80.0}}\tabularnewline
Norm. Prec. & {\footnotesize{}73.3} & {\footnotesize{}70.2} & {\footnotesize{}77.1} & {\footnotesize{}80.0} & {\footnotesize{}80.1} & {\footnotesize{}\textbf{85.4}}\tabularnewline
Success & {\footnotesize{}63.8} & {\footnotesize{}61.1} & {\footnotesize{}70.3} & {\footnotesize{}73.3} & {\footnotesize{}74.0} & {\footnotesize{}\textbf{81.2}}\tabularnewline
\bottomrule
\end{tabular}
\caption{\label{tab:trackingnet}Results on TrackingNet \cite{Muller18ECCV}.}
\end{table}

\begin{table}[t]
\centering{}%
\setlength{\tabcolsep}{1pt}
\footnotesize
\begin{tabular}{lcccccccc}
\toprule 
 & {\footnotesize{}LADCF } & {\footnotesize{}ATOM } & {\footnotesize{}SiamRPN++ } & THOR & {\footnotesize{}DiMP-50 } & {\footnotesize{}Ours } & {\footnotesize{}Ours} & \tabularnewline
& {\footnotesize{}\cite{Xu19TIP} } & {\footnotesize{}\cite{Danelljan19CVPR} } & {\footnotesize{}\cite{Li19CVPR} } & {\footnotesize{}\cite{Bhat19ICCV} } & {\footnotesize{}\cite{Sauer19BMVC}} & & {\scriptsize{}{(short-t.)}}  \tabularnewline
\midrule 
{\footnotesize{}EAO } & {\footnotesize{}0.389 } & {\footnotesize{}0.401 } & {\footnotesize{}0.414 } & {\footnotesize{}0.416} & \textbf{\footnotesize{}0.440}{\footnotesize{} } & {\footnotesize{}0.140 } & {\footnotesize{}0.408} & \tabularnewline
{\footnotesize{}Accuracy } & {\footnotesize{}0.503 } & {\footnotesize{}0.590 } & {\footnotesize{}0.600 } & {\footnotesize{}0.582} & {\footnotesize{}0.597 } & \textbf{\footnotesize{}0.624}{\footnotesize{} } & {\footnotesize{}0.609} & \tabularnewline
{\footnotesize{}Robustn. } & {\footnotesize{}0.159 } & {\footnotesize{}0.204 } & {\footnotesize{}0.234 } & {\footnotesize{}0.234} & \textbf{\footnotesize{}0.153 } & {\footnotesize{}1.039 } & {\footnotesize{}0.220} & \tabularnewline
\midrule 
{\footnotesize{}AO } & {\footnotesize{}0.421 } & {\footnotesize{}- } & \textbf{\footnotesize{}0.498}{\footnotesize{} } & {\footnotesize{}-} & {\footnotesize{}- } & {\footnotesize{}0.476 } & {\footnotesize{}0.462} & \tabularnewline
\bottomrule
\end{tabular}
\caption{Results on VOT2018 \cite{Kristan18ECCVW}.}
\label{tab:vot2018}
\end{table}

\PAR{VOT2018.} Tab.~\ref{tab:vot2018} shows our results on VOT2018 \cite{Kristan18ECCVW} (60 videos, 356 frames average length), where a reset-based evaluation is used. Once the object is lost, the tracker is restarted with the ground truth box five frames later and receives a penalty. The main evaluation criterion is the Expected Average Overlap (EAO) \cite{Kristan15ICCVW}.
This extreme short-term tracking scenario is not what Siam R-CNN with the TDPA was designed for. It often triggers resets, which without reset-based evaluation Siam R-CNN could automatically recover from, resulting in an EAO of $0.140$. For this setup, we created a simple short-term version of Siam R-CNN which %
 averages the predictions of re-detecting the first-frame reference and re-detecting the previous prediction and combines them with a strong spatial prior. %
 With $0.408$ EAO this variant is competitive with many SOTA approaches. Notably, both versions of Siam R-CNN achieve the highest accuracy scores. The last row shows the average overlap (AO), when using the normal (non-reset) evaluation. %
 When estimating rotated bounding boxes from segmentation masks produced by Box2Seg, Siam R-CNN's EAO increases to 0.423 and the accuracy greatly improves to 0.684. More details on rotated boxes and on the short-term tracking algorithm are in the supplemental material.

\subsection{Long-Term Visual Object Tracking Evaluation}
We evaluate Siam R-CNN's ability to perform long-term tracking on three benchmarks, LTB35 \cite{lukevzivc2018now}, LaSOT \cite{Fan19CVPRLASOT} and OxUvA \cite{Valmadre18ECCV}. In the supplemental material we also evaluate on UAV20L \cite{Mueller16ECCV}. In long-term tracking, sequences are much longer, and objects may disappear and reappear again (LTB35 has on average 12.4 disappearances per video, each one on average 40.6 frames long).
Siam R-CNN significantly outperforms all previous methods on all of these benchmarks, indicating the strength of our tracking by re-detection approach. By searching globally over the whole image rather than within a local window of a previous prediction, our method is more resistant to drift, and can easily re-detect a target after disappearance.

\begin{figure}[t]
\centering
 \scalebox{1.02}{
\includegraphics[scale=0.3]{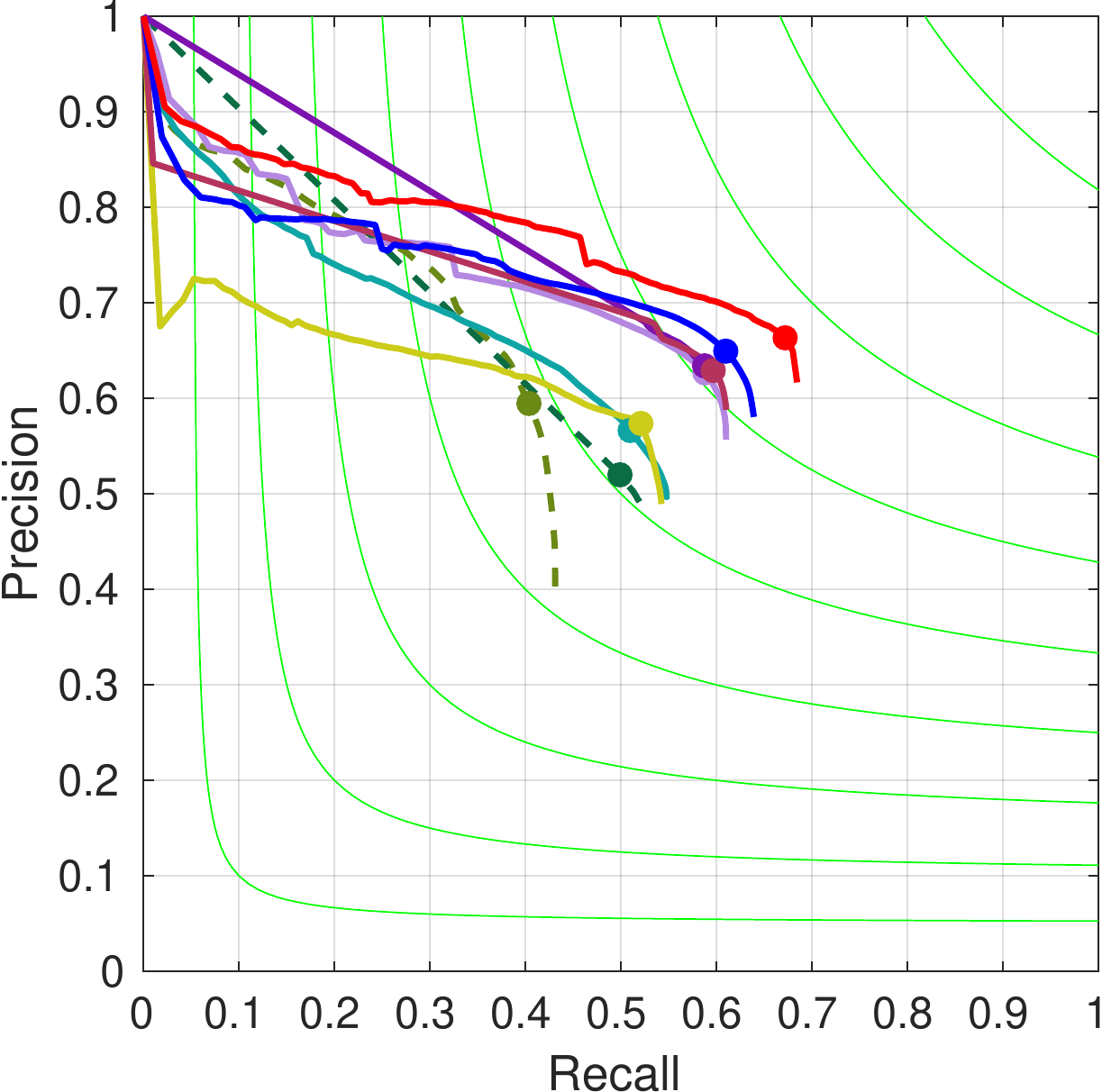}
\includegraphics[scale=0.3]{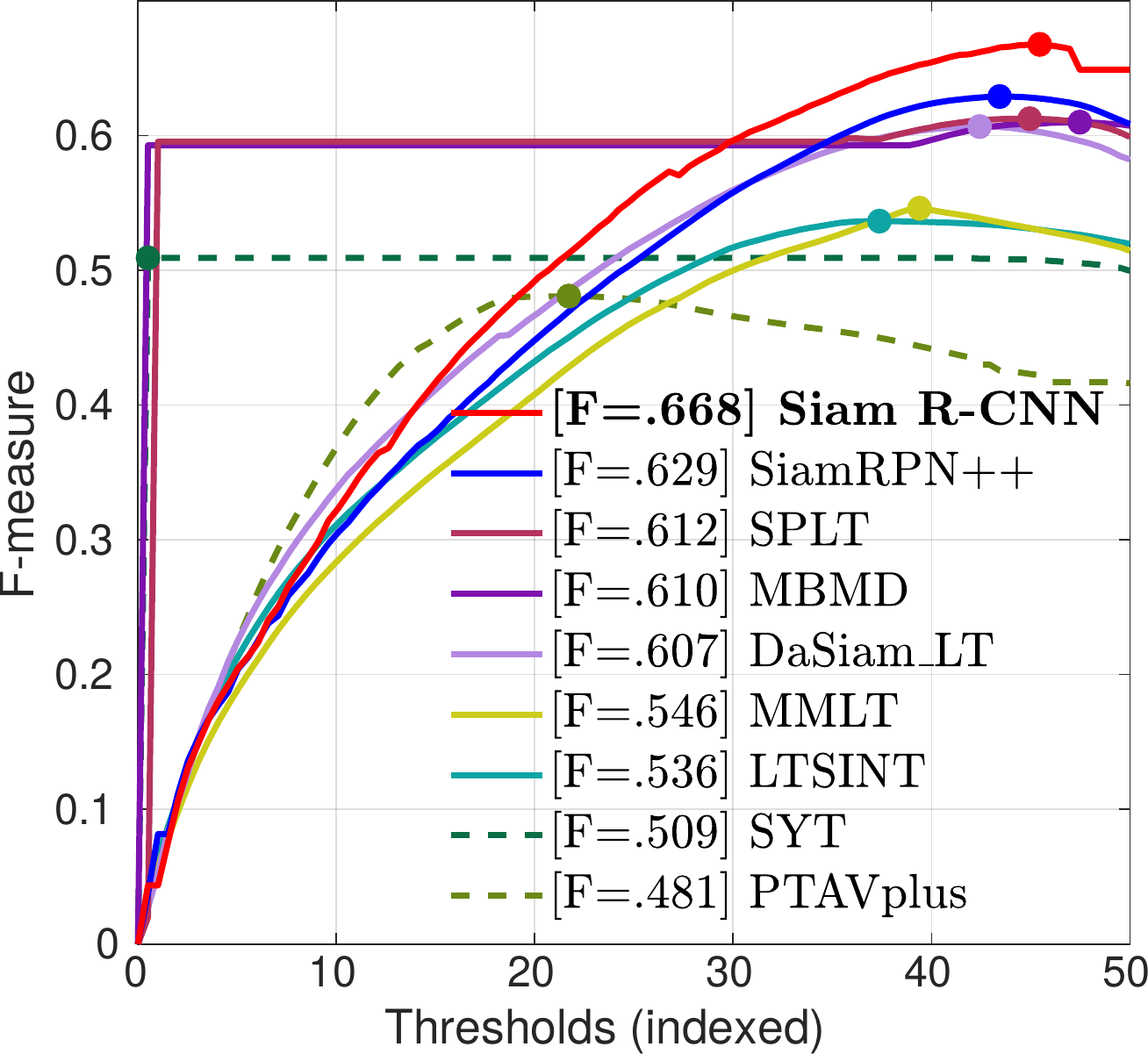}
}
\caption{Results on LTB35 \cite{lukevzivc2018now} (VOT18-LongTerm).} %
\label{fig:ltb35-res}
\end{figure}

\PAR{LTB35.}
Fig.~\ref{fig:ltb35-res} shows the results of our method on the LTB35 benchmark (also known as VOT18-LT) \cite{lukevzivc2018now} (35 videos, 4200 frames average length) compared to eight SOTA approaches.
Trackers are required to output a confidence of the target being present for the prediction in each frame. Precision (Pr) and Recall (Re) are evaluated for a range of confidence thresholds, and the $F$-score is calculated as $F = \frac{2PrRe}{Pr+Re}$. Trackers are ranked by the maximum $F$-score over all thresholds.
We compare to the 6 best-performing methods in the 2018 VOT-LT challenge \cite{Kristan18ECCVW} and to SiamRPN++ \cite{Li19CVPR} and SPLT \cite{Yan19ICCV}.
Siam R-CNN outperforms all previous methods with an $F$-score of $66.8\%$, \ie, $3.9$ percentage points higher than the previous best result.

\begin{figure}[t]
\centering
\includegraphics[width=0.49\columnwidth]{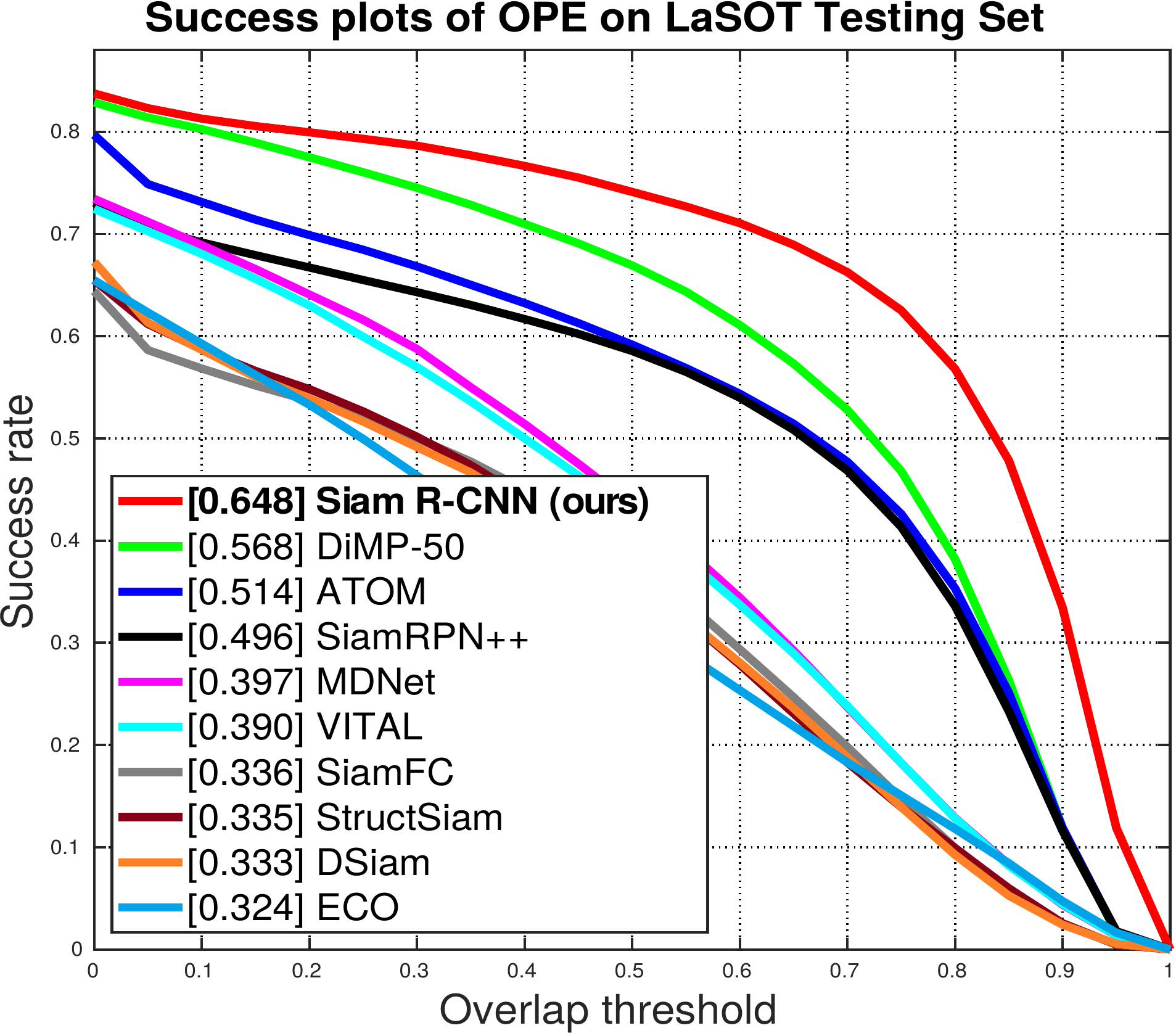}
\includegraphics[width=0.49\columnwidth]{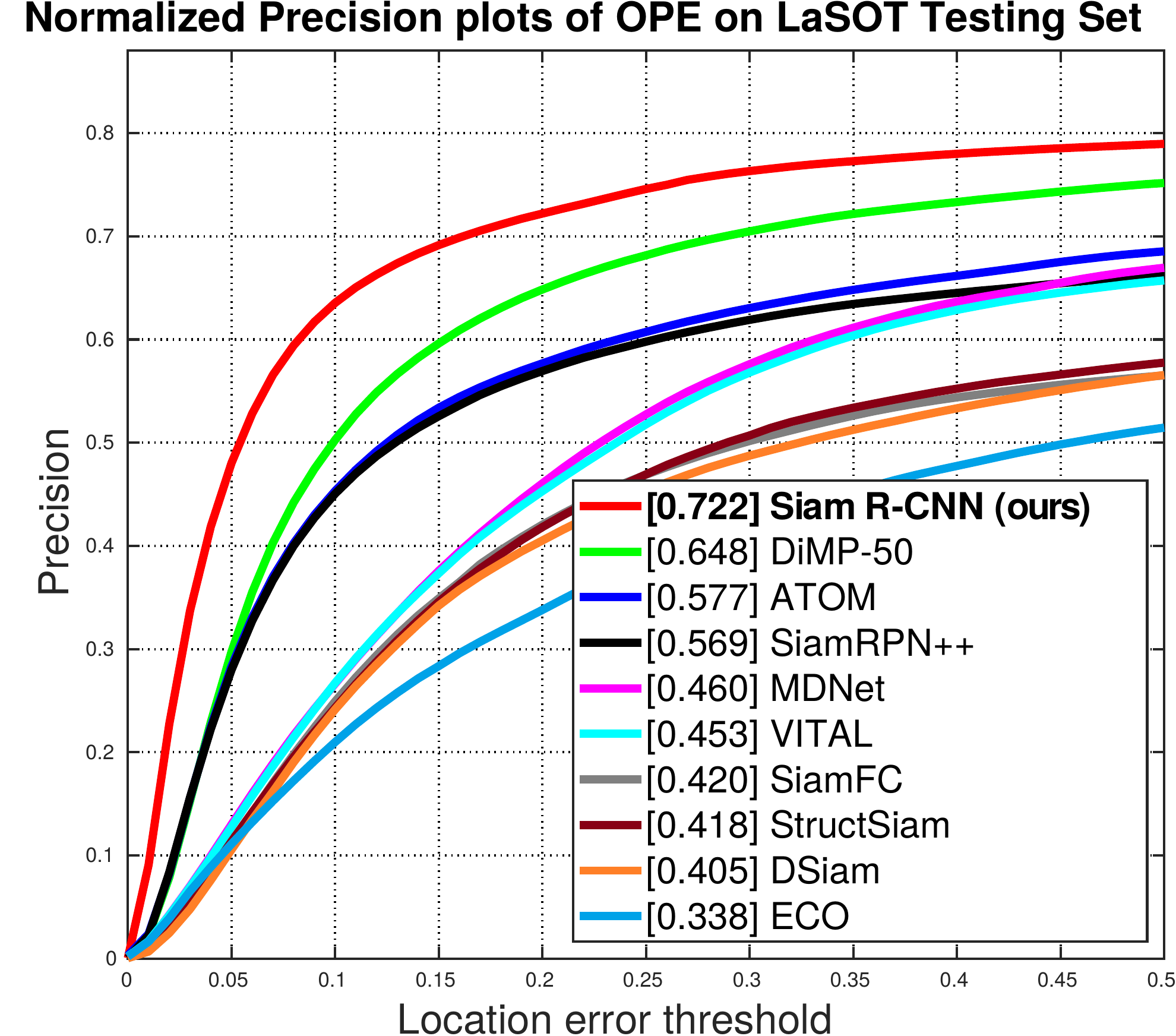}
\caption{Results on LaSOT \cite{Fan19CVPRLASOT}.}
\label{fig:lasot-res}
\end{figure}

\PAR{LaSOT.}
Fig.~\ref{fig:lasot-res} shows results on the LaSOT test set \cite{Fan19CVPRLASOT} (280 videos, 2448 frames average length) compared to nine SOTA methods \cite{Bhat19ICCV, Danelljan19CVPR, Li19CVPR, mdnet, Song18CVPR, Bertinetto2016ECCV, zhang2018structured, guo2017learning, Danelljan17CVPR}. Siam R-CNN achieves an unprecedented result with a success rate of $64.8\%$ and $72.2\%$ normalized precision. This is $8$ percentage points higher in success and $7.4$ points higher in normalized precision than the previous best method.

\PAR{OxUvA.}
Tab.~\ref{tab:oxuva} shows results on the OxUvA test set \cite{Valmadre18ECCV} (166 videos, 3293 frames average length) compared to five SOTA methods. Trackers must make a hard decision each frame whether the object is present. We do this by comparing the detector confidence to a threshold tuned on the dev set. Methods are ranked by the maximum geometric mean (MaxGM) of the true positive rate (TPR) and the true negative rate (TNR). Siam R-CNN achieves a MaxGM more than 10 percentage points higher than all previous methods.

\begin{table}[t]
\centering{}%
\setlength{\tabcolsep}{2pt}
\footnotesize
\begin{tabular}{lcccccc}
\toprule 
 & {\footnotesize{}DaSiam\_LT} & {\footnotesize{}TLD} & {\footnotesize{}SiamFC+R} & {\footnotesize{}MBMD} & {\footnotesize{}SPLT} & {\footnotesize{}Siam}\tabularnewline
& {\footnotesize{}\cite{Kristan18ECCVW}} & {\footnotesize{}\cite{TLD}} & {\footnotesize{}\cite{Valmadre18ECCV}} & {\footnotesize{}\cite{Zhang18Arxiv}} & {\footnotesize{}\cite{Yan19ICCV}} & {\footnotesize{}R-CNN} \tabularnewline 
\midrule 
{\footnotesize{}MaxGM} & {\footnotesize{}41.5} & {\footnotesize{}43.1} & {\footnotesize{}45.4} & {\footnotesize{}54.4} & {\footnotesize{}62.2} & {\footnotesize{}\textbf{72.3}}\tabularnewline
{\footnotesize{}TPR} & {\footnotesize{}68.9} & {\footnotesize{}20.8} & {\footnotesize{}42.7} & {\footnotesize{}60.9} & {\footnotesize{}49.8} & {\footnotesize{}\textbf{70.1}}\tabularnewline
{\footnotesize{}TNR} & {\footnotesize{}0} & {\footnotesize{}\textbf{89.5}} & {\footnotesize{}48.1} & {\footnotesize{}48.5} & {\footnotesize{}77.6} & {\footnotesize{}74.5}\tabularnewline
\bottomrule
\end{tabular}
\caption{\label{tab:oxuva}Results on OxUvA \cite{Valmadre18ECCV}.}
\end{table}

\begin{table}[t]
\footnotesize
\centering{}{\footnotesize{}}%
\scalebox{0.92}{
\setlength{\tabcolsep}{2pt}
\begin{tabular}{lcccccccc}
\toprule 
{\footnotesize{}Init}\hspace{-4mm} & {\footnotesize{}Method} & {\footnotesize{}FT} & {\footnotesize{}M} & {\footnotesize{}$\mathcal{J}$\&$\mathcal{F}$} & {\footnotesize{}$\mathcal{J}$} & {\footnotesize{}$\mathcal{F}$} & {\footnotesize{}$\mathcal{J}_{box}$} & {\footnotesize{}t(s)}\tabularnewline

\midrule

\parbox[t]{2mm}{\multirow{5}{*}{\rotatebox[origin=c]{90}{bbox}}} &
{\footnotesize{}\textbf{Siam R-CNN (ours)}} & {\footnotesize{}\ding{55}} & {\footnotesize{}\ding{55}} & {\footnotesize{}$\mathbf{70.6}$} & {\footnotesize{}$66.1$} & {\footnotesize{}$\mathbf{75.0}$} & {\footnotesize{}$\mathbf{78.3}$} & {\footnotesize{}$0.32$}\tabularnewline

&{\footnotesize{}Siam R-CNN (fastest)} & {\footnotesize{}\ding{55}} & {\footnotesize{}\ding{55}} & {\footnotesize{}$70.5$} & {\footnotesize{}$\mathbf{66.4}$} & {\footnotesize{}$74.6$} & {\footnotesize{}$76.9$} & {\footnotesize{}$0.12$}\tabularnewline

&
{\footnotesize{}SiamMask \cite{Wang19CVPR}} & {\footnotesize{}\ding{55}} & {\footnotesize{}\ding{55}} & {\footnotesize{}$55.8$} & {\footnotesize{}$54.3$} & {\footnotesize{}$58.5$} &  {\footnotesize{}$64.3$} & {\footnotesize{}$\mathbf{0.06^{\dagger}}$}\tabularnewline

&{\footnotesize{}SiamMask \cite{Wang19CVPR} (Box2Seg)} & {\footnotesize{}\ding{55}} & {\footnotesize{}\ding{55}} & {\footnotesize{}$63.3$} & {\footnotesize{}$59.5$} & {\footnotesize{}$67.3$} &  {\footnotesize{}$64.3$} & {\footnotesize{}$0.11$}\tabularnewline

&{\footnotesize{}SiamRPN++ \cite{Li19CVPR} (Box2Seg)} & {\footnotesize{}\ding{55}} & {\footnotesize{}\ding{55}} & {\footnotesize{}$61.6$} & {\footnotesize{}$56.8$} & {\footnotesize{}$66.3$} &  {\footnotesize{}$64.0$} & {\footnotesize{}$0.11$}\tabularnewline

&{\footnotesize{}DiMP-50 \cite{Bhat19ICCV} (Box2Seg)} & {\footnotesize{}\ding{55}} & {\footnotesize{}\ding{55}} & {\footnotesize{}$63.7$} & {\footnotesize{}$60.1$} & {\footnotesize{}$67.3$} &  {\footnotesize{}$65.6$} & {\footnotesize{}$0.10$}\tabularnewline

\midrule

\parbox{2mm}{\multirow{3}{*}{\rotatebox[origin=c]{90}{mask}}}&{\footnotesize{}STM-VOS \cite{Oh19ICCV}} & {\footnotesize{}\ding{55}} & {\footnotesize{}\ding{51}} & {\footnotesize{}$\mathbf{81.8}$} & {\footnotesize{}$\mathbf{79.2}$} & {\footnotesize{}$\mathbf{84.3}$} & {\footnotesize{}$-$} & {\footnotesize{}$0.32^{\dagger}$}\tabularnewline

&{\footnotesize{}FEELVOS \cite{Voigtlaender19CVPR}} & {\footnotesize{}\ding{55}} & {\footnotesize{}\ding{51}} & {\footnotesize{}$71.5$} & {\footnotesize{}$69.1$} & {\footnotesize{}$74.0$} & {\footnotesize{}$71.4$} & {\footnotesize{}$0.51$}\tabularnewline

&{\footnotesize{}RGMP \cite{Oh18CVPR}} & {\footnotesize{}\ding{55}} & {\footnotesize{}\ding{51}} & {\footnotesize{}$66.7$} & {\footnotesize{}$64.8$} & {\footnotesize{}$68.6$} & {\footnotesize{}66.5} & \textbf{\footnotesize{}$\mathbf{0.28^{\dagger}}$}\tabularnewline

\midrule

\parbox[t]{2mm}{\multirow{3}{*}{\rotatebox[origin=c]{90}{mask+ft}}}&{\footnotesize{}PReMVOS \cite{Luiten18ACCV}} & {\footnotesize{}\ding{51}} & {\footnotesize{}\ding{51}} & {\footnotesize{}$\mathbf{77.8}$} & {\footnotesize{}$\mathbf{73.9}$} & {\footnotesize{}$\mathbf{81.7}$} & {\footnotesize{}$\mathbf{81.4}$} & {\footnotesize{}$37.6$}\tabularnewline

&{\footnotesize{}\textbf{Ours (Fine-tun. Box2Seg)}} & {\footnotesize{}\ding{51}} & {\footnotesize{}\ding{51}} & {\footnotesize{}$74.8$} & {\footnotesize{}$69.3$} & {\footnotesize{}$80.2$} & {\footnotesize{}$78.3$} & {\footnotesize{}$\mathbf{1.0}$}\tabularnewline

&{\footnotesize{}DyeNet \cite{Li18ECCV}} & {\footnotesize{}\ding{51}} & {\footnotesize{}\ding{51}} & {\footnotesize{}$74.1$} & {\footnotesize{}$-$} & {\footnotesize{}$-$} & {\footnotesize{}$-$} & {\footnotesize{}$9.32^{\dagger}$}\tabularnewline

\midrule

&{\footnotesize{}GT boxes (Box2Seg)} & {\footnotesize{}\ding{55}} & {\footnotesize{}\ding{55}} & {\footnotesize{}$82.6$} & {\footnotesize{}$79.3$} & {\footnotesize{}$85.8$} & {\footnotesize{}$100.0$} & {\footnotesize{}$-$}\tabularnewline

&{\footnotesize{}GT boxes (Fine-t. Box2Seg)} & {\footnotesize{}\ding{51}} & {\footnotesize{}\ding{51}} & {\footnotesize{}$86.2$} & {\footnotesize{}$81.8$} & {\footnotesize{}$90.5$} &  {\footnotesize{}$100.0$} & {\footnotesize{}$-$}\tabularnewline

\bottomrule
\end{tabular}}{\footnotesize{}\caption{\label{tab:results-davis17}Results on the DAVIS 2017
validation set. FT: fine-tuning, M: using the first-frame masks, t(s): time per frame in seconds. $\dagger$: timing extrapolated from DAVIS 2016. %
An extended table is in the supplemental material. Siam R-CNN (fastest) denotes Siam R-CNN with ResNet-50 backbone, half input resolution, and 100 RoIs, see Section \ref{subsec:faster}.
}
}{\footnotesize \par}
\end{table}

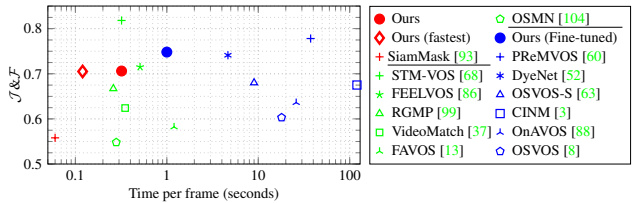
\begin{figure}
\centering
\resizebox{\linewidth}{!}{\begin{tikzpicture}[/pgfplots/width=1\linewidth, /pgfplots/height=0.6\linewidth, /pgfplots/legend pos=south east]
    \begin{axis}[ymin=0.50,ymax=0.85,xmin=0.05,xmax=130,enlargelimits=false,
        xlabel=Time per frame (seconds),
        ylabel=$\mathcal{J}$\&$\mathcal{F}$,
		font=\small,%
        grid=both,
		grid style=dotted,
        xlabel shift={-2pt},
        ylabel shift={-5pt},
        xmode=log,
        legend columns=2,
        minor ytick={0,0.025,...,1.1},
        ytick={0,0.1,...,1.1},
	    yticklabels={0,0.1,0.2,0.3,0.4,0.5,0.6,0.7,0.8,0.9,1},
	    xticklabels={0.01,0.1,1,10,100},
        legend pos= outer north east,
        legend cell align={left}
        ]

	\addplot[red,mark=*,only marks,line width=0.75, mark size=3.0] coordinates{(0.32,0.706)};
        \addlegendentry{\hphantom{i}Ours}
        
        \addplot[green,mark=pentagon, only marks, line width=0.75, mark size=2.5] coordinates{(0.28,0.548)};
        \addlegendentry{\underline{\hphantom{i}OSMN \cite{Yang18CVPR}\phantom{iiiiiiii}}}

	\addplot[red,mark=diamond,only marks,line width=1.5, mark size=3.5] coordinates{(0.12,0.705)};
        \addlegendentry{\hphantom{i}Ours (fastest)}
        
  	\addplot[blue,mark=*,only marks,line width=0.75, mark size=3.0] coordinates{(1.0,0.748)};
        \addlegendentry{\hphantom{i}Ours (Fine-tuned)}
        
        \addplot[red,mark=+, only marks, line width=0.75, mark size=2.5] coordinates{(0.06,0.558)};
        \addlegendentry{\underline{SiamMask \cite{Wang19CVPR}\phantom{iii}}}
        
        \addplot[blue,mark=+,only marks,line width=0.75, mark size=2.5] coordinates{(37.4,0.778)};
        \addlegendentry{\hphantom{i}PReMVOS \cite{Luiten18ACCV}}

        \addplot[green,mark=+, only marks, line width=0.75, mark size=2.5] coordinates{(0.32,0.818)};
        \addlegendentry{\hphantom{i}STM-VOS \cite{Oh19ICCV}}
        
       \addplot[blue,mark=asterisk, only marks, line width=0.75, mark size=2.5]coordinates{(4.66,0.741)};
        \addlegendentry{\hphantom{i}DyeNet \cite{Li18ECCV}}
        
        \addplot[green,mark=star,only marks,line width=0.75, mark size=2.5] coordinates{(0.51,0.715)};
        \addlegendentry{\hphantom{i}FEELVOS \cite{Voigtlaender19CVPR}}
        
        \addplot[blue,mark=triangle, only marks, line width=0.75, mark size=2.5] coordinates{(9,0.68)};
        \addlegendentry{\hphantom{i}OSVOS-S \cite{Maninis18TPAMI}}
        
        \addplot[green,mark=triangle, only marks, line width=0.75, mark size=2.5] coordinates{(0.26,0.667)};
        \addlegendentry{\hphantom{i}RGMP \cite{Oh18CVPR}}
        
        \addplot[blue,mark=square, only marks, line width=0.75, mark size=2.5] coordinates{(120,0.675)};
        \addlegendentry{\hphantom{i}CINM \cite{Bao18CVPR}} 
        
        \addplot[green,mark=square,only marks,line width=0.75, mark size=2.0] coordinates{(0.35,0.624)};
        \addlegendentry{\hphantom{i}VideoMatch \cite{Hu18ECCV}}

        \addplot[blue,mark=Mercedes star,only marks,line width=0.75, mark size=2.5] coordinates{(26,	0.636)};
        \addlegendentry{\hphantom{i}OnAVOS \cite{voigtlaender17BMVC}}
        
        \addplot[green,mark=Mercedes star, only marks, line width=0.75, mark size=2.5] coordinates{(1.2,0.582)};
        \addlegendentry{\hphantom{i}FAVOS \cite{Cheng18CVPR}}  
        
        \addplot[blue,mark=pentagon, only marks, line width=0.75, mark size=2.5] coordinates{(18,0.603)};
        \addlegendentry{\hphantom{i}OSVOS \cite{OSVOS}}

    \end{axis}
\end{tikzpicture}}
\vspace{-5mm}
   \caption{Quality versus timing on DAVIS 2017 (validation set). Only SiamMask \cite{Wang19CVPR} and our method (red) can work without the first-frame ground truth mask and require just the bounding box. Methods shown in blue fine-tune on the first-frame mask. Ours (fastest) denotes Siam R-CNN with ResNet-50, half resolution, and 100 RoIs, see Section \ref{subsec:faster}.}
   \label{fig:davis17-speedplot}
\end{figure}

\subsection{Video Object Segmentation (VOS) Evaluation}
We further evaluate the ability to track multiple objects and to segment them on VOS datasets %
 using the $\mathcal{J}$ metric (mask intersection over union (IoU)), the $\mathcal{F}$ metric (mask boundary similarity), %
 and the bounding box IoU $\mathcal{J}_{box}$. %

\begin{table}[t]
\centering{}{\footnotesize{}}%
\footnotesize
\scalebox{0.92}{
\setlength{\tabcolsep}{3pt}
\begin{tabular}{lcccccccc}
\toprule 
{\footnotesize{}Init}\hspace{-4mm} & {\footnotesize{}Method} & {\footnotesize{}FT} & {\footnotesize{}M} & {\footnotesize{}$\mathcal{O}$} & {\footnotesize{}$\mathcal{J}_{seen}$} & {\footnotesize{}$\mathcal{J}_{unseen}$} & {\footnotesize{}t(s)}\tabularnewline
\midrule

\parbox[t]{2mm}{\multirow{3}{*}{\rotatebox[origin=c]{90}{bbox}}} &
{\footnotesize{}\textbf{Siam R-CNN (ours)}} & {\footnotesize{}\ding{55}} & {\footnotesize{}\ding{55}} & {\footnotesize{}$\mathbf{68.3}$} & {\footnotesize{}$\mathbf{69.9}$} & \textbf{\footnotesize{}$\mathbf{61.4}$} & {\footnotesize{}$0.32$}\tabularnewline

&{\footnotesize{}\textbf{Siam R-CNN (fastest)}} & {\footnotesize{}\ding{55}} & {\footnotesize{}\ding{55}} & {\footnotesize{}$66.2$} & {\footnotesize{}$69.2$} & \textbf{\footnotesize{}$57.7$} & {\footnotesize{}$0.12$}\tabularnewline

&{\footnotesize{}SiamMask \cite{Wang19CVPR}} & {\footnotesize{}\ding{55}} & {\footnotesize{}\ding{55}} & {\footnotesize{}$52.8$} & {\footnotesize{}$60.2$} & \textbf{\footnotesize{}$45.1$} &  {\footnotesize{}$\mathbf{0.06}$}\tabularnewline

\midrule

\parbox{2mm}{\multirow{2}{*}{\rotatebox[origin=c]{90}{mask}}}&
{\footnotesize{}STM-VOS \cite{Oh19ICCV}} & {\footnotesize{}\ding{55}} & {\footnotesize{}\ding{51}} & {\footnotesize{}$\mathbf{79.4}$} & {\footnotesize{}$\mathbf{79.7}$} & {\footnotesize{}$\mathbf{72.8}$} & {\footnotesize{}$0.30^{\dagger}$}\tabularnewline

&{\footnotesize{}RGMP \cite{Oh18CVPR}} & {\footnotesize{}\ding{55}} & {\footnotesize{}\ding{51}} & \textbf{\footnotesize{}$53.8$} & {\footnotesize{}$59.5$} & \textbf{\footnotesize{}$45.2$} & \textbf{\footnotesize{}$\mathbf{0.26}^{\dagger}$}\tabularnewline

\midrule 

\parbox[t]{2mm}{\multirow{4}{*}{\rotatebox[origin=c]{90}{mask+ft}}}&
{\footnotesize{}\textbf{Ours (Fi.-tu. Box2Seg)}} & {\footnotesize{}\ding{51}} & {\footnotesize{}\ding{51}} & {\footnotesize{}{$\mathbf{73.2}$}} & \textbf{\footnotesize{}{$\mathbf{73.5}$}} & \textbf{\footnotesize{}{$\mathbf{66.2}$}} & \textbf{\footnotesize{}{$\mathbf{0.65}$}}\tabularnewline

&{\footnotesize{}PReMVOS \cite{Luiten18ACCV,Luiten18ECCVW}} & {\footnotesize{}\ding{51}} & {\footnotesize{}\ding{51}} & {\footnotesize{}$66.9$} & {\footnotesize{}$71.4$} & {\footnotesize{}$56.5$} & {\footnotesize{}$6$}\tabularnewline

&{\footnotesize{}OnAVOS \cite{voigtlaender17BMVC}} & {\footnotesize{}\ding{51}} & {\footnotesize{}\ding{51}} & {\footnotesize{}$55.2$} & {\footnotesize{}$60.1$} & {\footnotesize{}$46.6$} & {\footnotesize{}$24.5$}\tabularnewline

&{\footnotesize{}OSVOS \cite{OSVOS}} & {\footnotesize{}\ding{51}} & {\footnotesize{}\ding{51}} & {\footnotesize{}$58.8$} & {\footnotesize{}$59.8$} & {\footnotesize{}$54.2$} & {\footnotesize{}$17^{\dagger}$}\tabularnewline

\bottomrule
\end{tabular}}{\footnotesize{}\caption{\label{tab:results-youtubevos}Results on the YouTube-VOS 2018 \cite{Xu18ECCV} validation set. %
 The notation is explained in the caption of Tab.~\ref{tab:results-davis17}.
}
}
\end{table}

\PARbegin{DAVIS 2017.}
Tab.~\ref{tab:results-davis17} and Fig.~\ref{fig:davis17-speedplot} show results on the DAVIS 2017 validation set (30 videos, 2.03 objects and 67.4 frames average length per video). %
Methods are ranked by the mean of %
$\mathcal{J}$ and $\mathcal{F}$.
Siam R-CNN significantly outperforms the previous best method that only uses the first-frame bounding boxes, SiamMask \cite{Wang19CVPR}, by $14.8$ percentage points. To evaluate how much of this improvement comes from Box2Seg and how much from our tracking, we applied Box2Seg to the output of SiamMask. This does improve the results while still being $7.3$ percentage points worse than our method. We also run SiamRPN++ \cite{Li19CVPR} and DiMP-50 \cite{Bhat19ICCV} with Box2Seg for comparison. As a reference for the achievable performance for our tracker, we ran Box2Seg on the ground truth boxes which resulted in a score of $82.6\%$.

Even without using the first-frame mask, Siam R-CNN outperforms many methods that use the mask such as RGMP \cite{Oh18CVPR} and VideoMatch \cite{Hu18ECCV}, and even some methods like OSVOS-S \cite{Maninis18TPAMI} that perform slow first-frame fine-tuning. %
Our method is also more practical, as it is far more tedious to create a perfect first-frame segmentation mask by hand than a bounding box initialization. If the first-frame mask is available, then we are able to fine-tune Box2Seg on this, improving results by $4.2$ percentage points at the cost of speed. We evaluate on the DAVIS 2017 test-dev benchmark and on DAVIS 2016 \cite{DAVIS2016} in the supplemental material.

\PAR{YouTube-VOS.}
Tab.~\ref{tab:results-youtubevos} %
shows results on %
 YouTube-VOS 2018 %
 \cite{Xu18ECCV} (474 videos, 1.89 objects and 26.6 frames average length per video). %
Methods are ranked by the %
mean $\mathcal{O}$ of the $\mathcal{J}$ and $\mathcal{F}$ metrics over classes in the training set (\textit{seen}) and unseen classes. Siam R-CNN again outperforms all methods which do not use the first-frame mask (by $15.5$ percentage points), and also outperforms PReMVOS \cite{Luiten18ACCV,Luiten18ECCVW} and all other previous methods except for STM-VOS \cite{Oh19ICCV}. %

\begin{table}[t]
\centering{}%
\footnotesize
\setlength{\tabcolsep}{3pt}

\begin{tabular}{ccccc}
\toprule 
Dataset & Speed & OTB2015 & LaSOT & LTB35\tabularnewline
Eval measure & FPS & AUC & AUC & F\tabularnewline
\midrule
Siam R-CNN & 4.7 & 70.1 & \textbf{64.8} & 66.8\tabularnewline
\midrule
No hard ex. mining & 4.7 & 68.4 & 63.2 & 66.5\tabularnewline
\midrule
Argmax & 4.9 & 63.8 & 62.9 & 65.5\tabularnewline
Short-term & 4.6 & 67.2 & 55.7 & 57.2\tabularnewline
\midrule
$\ensuremath{\frac{1}{2}}$res. + 100 RoIs & 13.6 & 69.1 & 63.2 & 66.0\tabularnewline
ResNet-50 & 5.1 & 68.0 & 62.3 & 64.4\tabularnewline
ResNet-50 + $\ensuremath{\frac{1}{2}}$res. + 100 RoIs & \textbf{15.2} & 67.7 & 61.1 & 63.7\tabularnewline
\bottomrule

\end{tabular}\caption{\label{tab:ablations}Ablation and timing analysis of Siam R-CNN.%
}
\end{table}

\subsection{Ablation and Timing Analysis}
Tab.~\ref{tab:ablations} shows a number of ablations of Siam R-CNN on three datasets together with their speed (using a V100 GPU). Siam R-CNN runs at 4.7 frames per second (FPS) using a ResNet-101 backbone, $1000$ RPN proposals per frame, and TDPA. %
The row ``No hard ex. mining" shows the results without hard example mining (\cf Sec.~\ref{sec:hardexample}). Hard example mining improves results on all datasets, by up to $1.7$ percentage points.
We compare TDPA to using just the highest scoring re-detection in each frame (``Argmax") and the short-term algorithm we used for the reset-based VOT18 evaluation (``Short-term"). TDPA outperforms both of these on all datasets. A per-attribute analysis of the influence of TDPA can be found in the supplemental material.
For the long-term datasets, Argmax significantly outperforms both the short-term variant and even all previous methods.%

\subsection{Making Siam R-CNN Even Faster}
\label{subsec:faster}
Tab.~\ref{tab:ablations} also shows the result of three changes aimed at increasing the speed of Siam R-CNN (smaller backbone, smaller input resolution, and less RoI proposals). More details and analyses are in the supplemental material.

When evaluating with a ResNet-50 backbone, Siam R-CNN performs slightly faster and still achieves SOTA results ($62.3$ on LaSOT, compared to $56.8$ for DiMP-50 with the same backbone). This shows that the results are not only due to a larger backbone. %
When using half input resolution and only 100 RoIs from the RPN, the speed increases from 4.7 FPS to 13.6 FPS, or even 15.2 FPS in the case of ResNet-50. These setups still show very strong results, especially for long-term tracking. Note that even the fastest variant is not real-time and our work focuses on accuracy achieving much better results, especially for long-term tracking, while still running at a reasonable speed.

\begin{figure}[t]
\centering
 \scalebox{1.0}{
\includegraphics[width=0.49\columnwidth]{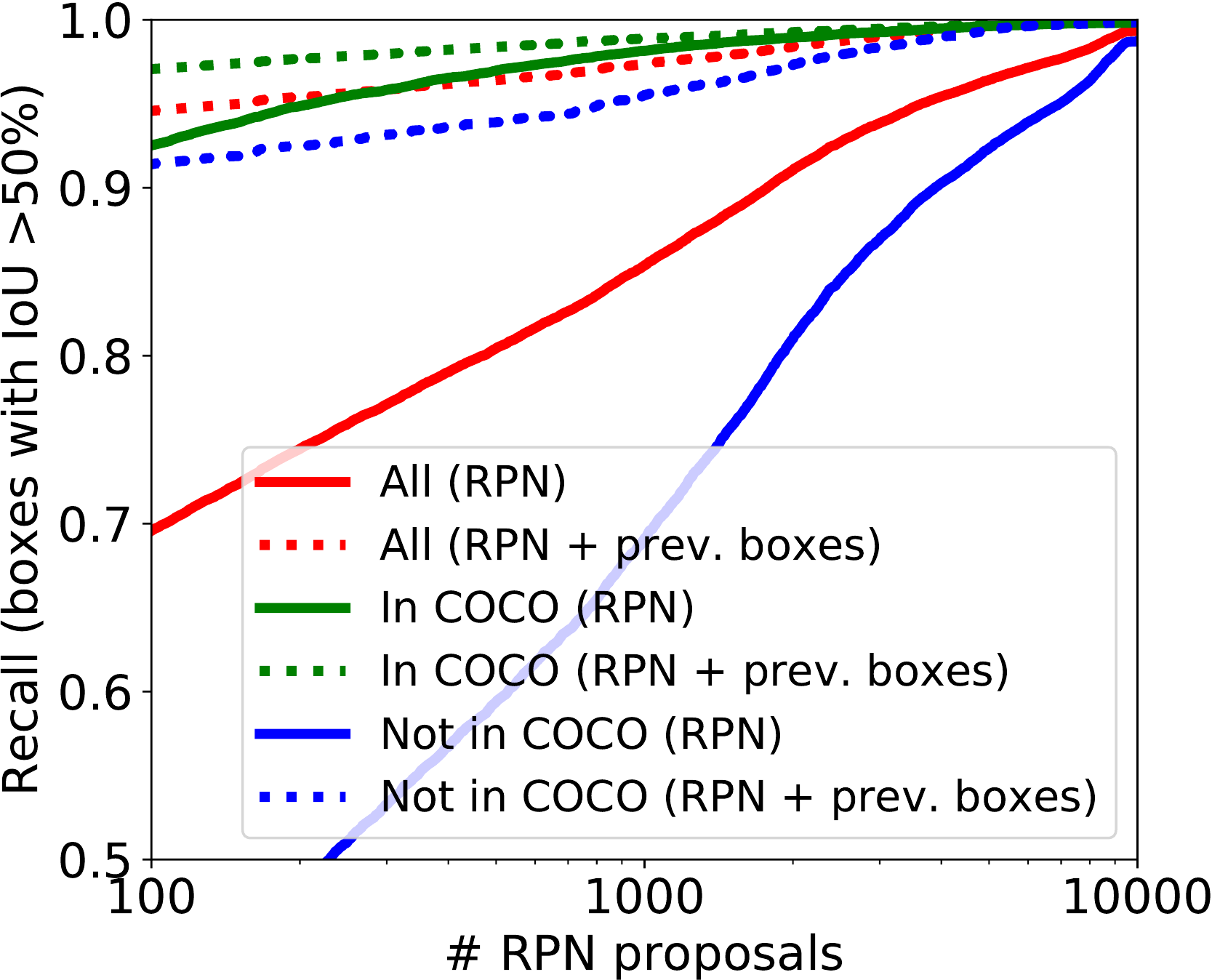}
\includegraphics[width=0.49\columnwidth]{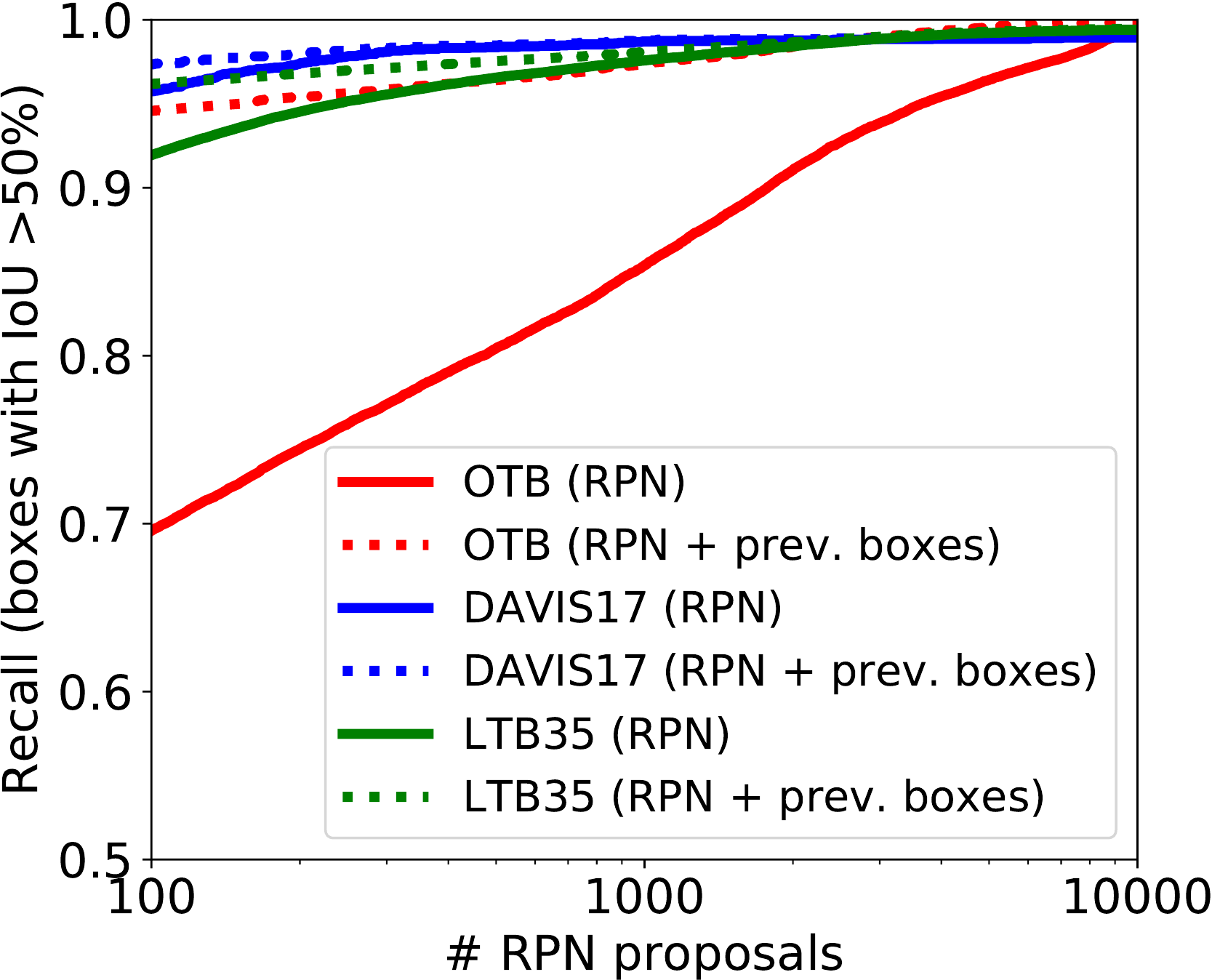}
}
\caption{%
RPN recall with varying number of proposals. Dotted lines have up to 100 re-detections from the previous frame added. Left: comparison on COCO/non-COCO classes of OTB2015. Right: comparison over three datasets.}
\label{fig:objectness}
\end{figure}

\subsection{Generic Object Tracking Analysis}
Siam R-CNN should be able to track any generic object. However, its backbone and RPN have been trained only on 80 object classes in COCO and have then been frozen. %
In Fig.~\ref{fig:objectness}, we investigate the recall of our RPN on the $44\%$ of OTB2015 sequences that contain objects not in COCO, compared to the rest. With the default of $1000$ proposals, the RPN achieves only $69.1\%$ recall for unknown objects, compared to $98.2\%$ for known ones. One solution is to increase the number of proposals used. When using $10,000$ proposals the RPN achieves $98.7\%$ recall for unknown objects but causes Siam R-CNN to run much slower (around $1$ FPS). Our solution is to instead include the previous-frame re-detections (up to $100$) as additional proposals. This increases the recall to $95.5\%$ for unknown objects when using $1000$ RPN proposals. %
 This shows why Siam R-CNN is able to outperform all previous methods on OTB2015, even though almost half of the objects are not from COCO classes. We also run a recall analysis on the DAVIS 2017 and LTB35 datasets where
  most objects belong to COCO classes and we achieve excellent recall (see Fig.~\ref{fig:objectness} right).

\vspace{-2pt}
\section{Conclusion}
We introduce Siam R-CNN as a Siamese two-stage full-image re-detection architecture with a Tracklet Dynamic Programming Algorithm. Siam R-CNN outperforms all previous methods on ten tracking benchmarks, with especially strong results for long-term tracking. We hope that our work will inspire future work on using two-stage architectures and full-image re-detection for tracking.

\footnotesize \PAR{Acknowledgements:} For partial funding of this project, PV, JL and BL would like to acknowledge the ERC Consolidator Grant DeeViSe (ERC-2017-COG-773161) and a Google Faculty Research Award. PHST would like to acknowledge CCAV project Streetwise and EPSRC/MURI grant
EP/N019474/1. The authors would like to thank Sourabh Swain, Yuxin Wu, Goutam Bhat, and Bo Li for helpful discussions.
\normalsize

{\small
\bibliographystyle{ieee_fullname}
\bibliography{abbrev_short,paper}
}

\newpage
\vskip .375in
\twocolumn[\begin{center}
{\Large \bf Supplemental Material for\\Siam R-CNN: Visual Tracking by Re-Detection \par}
\vskip 1.5em
\end{center}]
\appendix

\normalsize
\begin{abstract}
We provide more details of Siam RCNN's training procedure, a more detailed description of the video hard example mining procedure, of the short-term tracking algorithm, and of the estimation of rotated bounding boxes for VOT2018.
Additionally, we explain and analyze different methods to speed up Siam R-CNN in more detail and we analyze the amount of training data used by previous methods compared to our method. Moreover, we conduct a per-attribute analysis  of variants of Siam R-CNN on OTB2015 and also analyze the effect of fine-tuning Box2Seg.
Finally, we present additional quantitative and qualitative results for short-term tracking, long-term tracking and video object segmentation.
\end{abstract}

\section{Further Method Details}
In the following, we provide further details on the training procedure, video hard example mining, the short-term-tracking algorithm, and on rotated bounding box estimation.

\subsection{Training}

We train Siam R-CNN with random image scale sampling by scaling the small edge of the image to a random value between 640 and 800 pixels, while keeping the aspect ratio. In cases where this resizing would result in a larger edge size of more than 1333 pixels, it is resized to a larger edge size of 1333 pixels instead. During test time we resize the image to a smaller edge length of 800 pixels, again keeping the longer edge size no larger than 1333 pixels. Note that these settings are the default of the Mask R-CNN implementation we used.

We train our network with two NVIDIA GTX 1080 Ti GPUs for 1 million steps (5.8 days) with a learning rate of 0.01, and afterwards for 120,000 steps and 80,000 steps with learning rates of 0.001 and 0.0001, respectively. The hard example training is done afterwards with a single GPU for 160,000 steps and a learning rate of 0.001. A batch size of one pair of images (reference and target) per GPU is used.

\subsection{Video Hard Example Mining}

\PAR{Index Structure.} For the indexing structure, we use the ``Approximate Nearest Neighbors Oh Yeah'' library for approximate nearest neighbor queries\footnote{https://github.com/spotify/annoy}.

\PAR{Feature Pre-computation.}
During normal training without hard negative examples, the RoIs given to the second stage are generated automatically by the RPN and are thus not always perfectly aligned to the object boundaries. In the three cascade stages, the RoI will then be successively refined by bounding box regression. However, when naively pre-computing the features for the ground truth bounding boxes, the network might overfit to these perfect boxes.

In order for the network to learn to handle imperfect bounding boxes, we add random Gaussian noise to the ground truth bounding boxes before pre-computing the features, and afterwards run these  jittered RoIs through the cascade stages and also pre-compute the re-aligned features after every cascade stage.

In particular, we take the ground truth bounding box $(x_0, y_0, x_1, y_1)$ and independently add to each component noise sampled from a clipped Gaussian with mean $0$ and standard deviation $0.25$, \ie, for $i \in \{0, 1, 2, 3\}$, we have
\begin{equation}
  \tilde{r}_i \sim \mathcal{N}(0, 0.25),
\end{equation}
\begin{equation}
  r_i = \mathrm{clip}(\tilde{r}_i, -0.25, 0.25).
\end{equation}
The jittered box is then given by
\begin{equation}
(x_0 + r_0, y_0 + r_1, x_1 + r_2, y_1 + r_3).
\end{equation}

\PAR{Training Procedure.}
Having pre-computed the RoI-aligned features for each object and each cascade stage, the training procedure now works as follows. 
For each training step, as usual, a random video and object in this video is selected and then a random reference and a random target frame. Afterwards, we use the indexing structure to retrieve the 10,000 nearest neighbor bounding boxes from other videos (50,000 boxes for LaSOT because of the  long sequences). Note that the nearest neighbors are searched for over all training datasets, regardless where the reference comes from. 
Since the nearest neighbors are found per frame, often a few videos will dominate the set of nearest neighbors. To get more diverse negative examples, we create a list of all videos (excluding the reference) in which nearest neighbor boxes were found and randomly select 100 of these videos. For each of the 100 videos, we then randomly select 1 of the boxes which were retrieved as nearest neighbors from this video and add them as negative examples for the re-detection head.

Adding only additional negative examples creates an imbalance in the training data. Hence, we also retrieve the features of the ground truth bounding boxes of 30 randomly selected frames of the current reference video as additional positive examples.

\subsection{Short-term Tracking Algorithm}
For the VOT2018 dataset \cite{Kristan18ECCVW}%
, it is standard to use a reset-based evaluation, where once the object is lost (0 IoU between predicted and ground truth bounding box), the tracker is restarted with the ground truth box five frames later and receives a penalty. %
This extreme short-term tracking scenario is not what Siam R-CNN with the Tracklet Dynamic Programming Algorithm (TDPA) was designed for. It often triggers resets, which normally (without reset-based evaluation) Siam R-CNN could have automatically recovered from.%

Since VOT2018 is an important tracking benchmark, we created a short-term version of the Siam R-CNN tracking algorithm (see Alg.~\ref{alg:short-term}).
\begin{algorithm}[t]
\small
\algnewcommand{\LineComment}[1]{\State \(\triangleright\) #1}
\caption{\label{alg:short-term}Perform Short-term Tracking for time-step $t$}
\begin{algorithmic}[1]
\State \textbf{Inputs} $\mathrm{ff\_gt\_feats, image_t, det_{t-1}}$ \label{sota:input}
\State $\mathrm{backbone\_feats} \gets \mathrm{backbone}(\mathrm{image}_t)$ 
\State $\mathrm{RoIs} \gets \mathrm{RPN}(\mathrm{backbone\_feats})$ \label{sota:rpn}
\LineComment{Add shifted versions of $\mathrm{det}_\mathrm{t-1}$ to RoIs}
\For{$\mathrm{shift}_x \in \{-1.5, -1.0, -0.5, 0.0, 0.5, 1.0, 1.5\}$} \label{sota:shiftbegin}
  \For{$\mathrm{shift}_y \in \{-1.5, -1.0, -0.5, 0.0, 0.5, 1.0, 1.5\}$}
    \State $\mathrm{RoIs} \gets \mathrm{RoIs} \cup \{\mathrm{shift}(\mathrm{det}_{t-1}, \mathrm{shift}_x, \mathrm{shift}_y)\}$
  \EndFor
\EndFor \label{sota:shiftend}
\State $\mathrm{dets}_t, \mathrm{det\_scores}_t \gets \mathrm{redetection\_head}(\mathrm{RoIs}, \mathrm{ff\_gt\_feats})$ \label{sota:det}
\State $\mathrm{prev\_scores}_t \gets \mathrm{score\_redetection}(\mathrm{dets}_t, \mathrm{det}_{t-1})$ \label{sota:prev-score}
\State $\mathrm{loc\_scores}_t \gets |\mathrm{bbox}(\mathrm{dets}_t) - \mathrm{bbox}(\mathrm{det}_{t-1})|_1$ \label{sota:loc-score}
\State $\mathrm{scores}_t \gets \mathrm{det\_scores}_t + \mathrm{prev\_scores}_t + \delta \cdot \mathrm{loc\_scores}_t$ \label{sota:score-combine}
\LineComment{Filter out detections with too large spatial distance}
\For{$\mathrm{det}_t \in \mathrm{dets}_t$} \label{sota:filterbegin}
  \If{$\lVert\mathrm{det}_t - \mathrm{det}_{t-1}\rVert_{\infty} > \xi$}
    \State $\mathrm{scores}[\mathrm{det}_t] \gets -\infty$
  \EndIf
\EndFor \label{sota:filterend}
\State \Return $\arg\max_{\mathrm{det}_t \in \mathrm{dets}_t} \mathrm{scores}_t[\mathrm{det}_t]$ \label{sota:res}
\end{algorithmic}
\end{algorithm}
Given the RoI Aligned features $\mathrm{ff\_gt\_feats}$ of the first-frame bounding box and the previous-frame tracking result $\mathrm{det}_{t-1}$, we first extract the backbone features of the current image and afterwards generate regions of interest (RoIs) using the region proposal network (RPN, lines~\ref{sota:input}--\ref{sota:rpn}). 

Note, that we know for sure that the previous-frame predicted box $\mathrm{det}_{t-1}$ has a positive IoU with the ground truth box, as otherwise a reset would have been triggered. Hence, the object to be tracked should be located close to the previous-frame predicted box.
In order to exploit this, and to compensate for potential false negatives of the RPN, we add shifted versions of the previous-frame prediction as additional RoIs (lines~\ref{sota:shiftbegin}--\ref{sota:shiftend}). Here, the function $\mathrm{shift}(\cdot)$ shifts the previous-frame box $\mathrm{det}_{t-1}$ by factors of its width and height, \eg, if $\mathrm{shift}_x=0.5$ and $\mathrm{shift}_y=1.0$, the box is shifted by half its width in x-direction and by its full height in y-direction.

Afterwards, we use the $\mathrm{redetection\_head}$ to produce detections $\mathrm{dets}_t$ with detection scores $\mathrm{det\_scores}_t$ for the current frame $t$ (line~\ref{sota:det}). We then additionally compute previous-frame scores $\mathrm{prev\_scores}_t$ by using the previous-frame box as a reference to score the current detections (line~\ref{sota:prev-score}). 
To exploit spatial consistency, we also compute location scores ($\mathrm{loc\_scores}_t$, line~\ref{sota:loc-score}), given by the $L_1$-norm of the pairwise differences between the current detection boxes and the previous-frame predicted box.
All three scores are then combined (line~\ref{sota:score-combine}) by a linear combination, where the current-frame and previous-frame scores have equal weight.

Even when using a location score, it can happen, that a distractor object appears far away from the object to be tracked and gets a high combined score because it looks very similar to that object. However, since we know that the previous-frame box has positive overlap with the ground truth box, a far-away detection cannot be the object to be tracked.
Hence, we explicitly filter out detections which have a large spatial distance (measured by the $L_{\infty}$-norm) to the previous-frame predicted box (lines~\ref{sota:filterbegin}--\ref{sota:filterend}).

Finally, we report the detection with the highest combined score as the result for the current frame (line~\ref{sota:res}), or in case there is no valid detection, we repeat the previous-frame result as result for the current frame.

\subsection{Rotated Bounding Box Estimation}
\begin{table}
\begin{centering}
\setlength{\tabcolsep}{3pt}
\begin{tabular}{lccc}
\toprule 
{\footnotesize{}Method} & {\footnotesize{}EAO $\uparrow$} & {\footnotesize{}Accuracy $\uparrow$} & {\footnotesize{}Robustn. $\downarrow$}\tabularnewline
\midrule 
{\footnotesize{}DiMP-50 \cite{Bhat19ICCV}} & \textbf{\footnotesize{}0.440} & {\footnotesize{}0.597} & {\footnotesize{}0.153}\tabularnewline
{\footnotesize{}SiamRPN++ \cite{Li19CVPR}} & {\footnotesize{}0.414} & {\footnotesize{}0.440} & {\footnotesize{}0.234}\tabularnewline
{\footnotesize{}ATOM \cite{Danelljan19CVPR}} & {\footnotesize{}0.401} & {\footnotesize{}0.590} & {\footnotesize{}0.204}\tabularnewline
\midrule
{\footnotesize{}Ours (short-term)} & {\footnotesize{}0.408} & {\footnotesize{}0.609} & {\footnotesize{}0.220}\tabularnewline
{\footnotesize{}+Rotated Boxes} & {\footnotesize{}0.409} & \textbf{\footnotesize{}0.686} & {\footnotesize{}0.272}\tabularnewline
{\footnotesize{}+Rotated Boxes +Mask Dens. Filt.} & \textbf{\footnotesize{}0.423} & {\footnotesize{}0.684} & \textbf{\footnotesize{}0.248}\tabularnewline
\bottomrule
\end{tabular}\caption{\label{tab:rotated-bboxes} Results using rotated bounding boxes on VOT2018. Mask Dens. Filt. denotes using a mask density filter.}
\par\end{centering}
\end{table}
For VOT2018, the ground truth is given as rotated bounding boxes which were automatically estimated by an optimization procedure based on hand-annotated segmentation masks \cite{Kristan16ECCVW}. Nevertheless, most methods produce axis-aligned bounding boxes and then evaluate against the rotated bounding box ground truth. 

As an extension of Siam R-CNN, we used Box2Seg to produce segmentation masks and then also ran the optimization procedure which was used to create the ground truth to generate rotated bounding boxes. Note that this optimization procedure (implemented in MATLAB) is very slow, slowing down our whole method to around 0.23 FPS. Note that the speed of the optimization might strongly depend on the hardware/software setup. However, here we do not aim for a good run-time but instead want to analyze the achievable performance using rotated bounding boxes. 

Table~\ref{tab:rotated-bboxes} shows our results on VOT2018 with rotated bounding boxes. When creating a rotated bounding box for each frame, the overall EAO stays almost the same and only increases from 0.408 to 0.409. However, the accuracy which measures the average intersection-over-union of the bounding box with the ground truth while disregarding resets, increases strongly from 0.609 to 0.686. At the same time, the number of resets strongly increases which can be seen by the robustness degrading from 0.220 to 0.272. 

A manual inspection of the results revealed that in some cases the estimated segmentation mask from Box2Seg was almost empty and the resulting rotated bounding box is hence of very poor quality and can easily trigger a reset. 

To avoid these cases, we apply a mask density filter, which means that in cases where the estimated segmentation mask fills less than 10\% of the bounding box which was used to generate it, we stick to the original axis-aligned bounding box in this frame instead of reporting the rotated bounding box.
In this setup, the EAO significantly increases to 0.423, while keeping a high accuracy of 0.684. With the mask density filter, Siam R-CNN achieves a robustness of 0.248 which is still worse than the robustness of the axis-aligned version, but significantly better than the version without the filter.

\section{Further Analyses}
In the following, we conduct further analyses of the speed-accuracy trade-off of Siam R-CNN and of the dependence of Siam R-CNN and other methods on the used training data. Additionally, we conduct a per-attribute analysis on OTB2015, and an analysis of the fine-tuning of Box2Seg.

\subsection{Speed-Accuracy Trade-off}
\begin{table}[t]
\centering{}%
\footnotesize
\setlength{\tabcolsep}{3pt}
\begin{tabular}{ccccc}
\toprule 
{\footnotesize{}Dataset} & {\footnotesize{}Speed} & {\footnotesize{}OTB2015} & {\footnotesize{}LaSOT} & {\footnotesize{}LTB35}\tabularnewline
{\footnotesize{}Eval measure} & {\footnotesize{}FPS} & {\footnotesize{}AUC} & {\footnotesize{}AUC} & {\footnotesize{}F}\tabularnewline
\midrule
{\footnotesize{}Siam R-CNN} & {\footnotesize{}4.7} & {\footnotesize{}70.1} & {\footnotesize{}\textbf{64.8}} & {\footnotesize{}66.8}\tabularnewline
\midrule
{\footnotesize{}ResNet-50} & {\footnotesize{}5.1} & {\footnotesize{}68.0} & {\footnotesize{}62.3} & {\footnotesize{}64.4}\tabularnewline
{\footnotesize{}$\frac{1}{2}$ res.} & {\footnotesize{}5.7} & {\footnotesize{}\textbf{70.2}} & {\footnotesize{}62.9} & {\footnotesize{}65.6}\tabularnewline
{\footnotesize{}100 RoIs} & {\footnotesize{}8.7} & {\footnotesize{}68.7} & {\footnotesize{}64.1} & {\footnotesize{}\textbf{67.3}}\tabularnewline
{\footnotesize{}100 RoIs + $\frac{1}{2}$ res.} & {\footnotesize{}13.6} & {\footnotesize{}69.1} & {\footnotesize{}63.2} & {\footnotesize{}66.0}\tabularnewline
{\footnotesize{}ResNet-50 + 100 RoIs} & {\footnotesize{}10.3} & {\footnotesize{}66.9} & {\footnotesize{}61.5} & {\footnotesize{}65.9}\tabularnewline
{\footnotesize{}ResNet-50 + $\frac{1}{2}$ res.} & {\footnotesize{}6.1} & {\footnotesize{}68.6} & {\footnotesize{}61.2} & {\footnotesize{}64.0}\tabularnewline
{\footnotesize{}ResNet-50 + 100 RoIs + $\frac{1}{2}$ res.} & {\footnotesize{}\textbf{15.2}} & {\footnotesize{}67.7} & {\footnotesize{}61.1} & {\footnotesize{}63.7}\tabularnewline
\bottomrule
\end{tabular}\caption{\label{tab:ablations-extended}Extended timing analysis of Siam R-CNN using a V100 GPU.}
\end{table}

Tab.~\ref{tab:ablations-extended} extends the timing analysis of the main paper. Here, we evaluate three changes aimed at increasing the speed of Siam R-CNN (smaller backbone, smaller input resolution, and fewer RoI proposals) in more detail.

When evaluating with a ResNet-50 backbone, Siam R-CNN performs slightly faster and still achieves state-of-the-art (SOTA) results ($62.3$ on LaSOT, compared to $56.8$ for DiMP-50 with the same backbone). This shows that the strong results are not only due to a larger backbone, but due to our tracking by re-detection approach.

We also evaluate reducing the image input size to a smaller image edge length of 400 pixels instead of 800 (row ``$\frac{1}{2}$ res"). This also results in only a slight decrease in performance in two benchmarks, and a slight increase in performance on OTB2015.

The row ``100 RoIs" shows the results of using 100 RoIs from the RPN, instead of 1000. This almost doubles the speed as most compute occurs in the re-detection head. This results in only a small score decrease on two  benchmarks, while improving results on LTB35. This shows that Siam R-CNN can run very quickly, even though it is based on a two-stage detection architecture, as very few RoIs are required.

The fastest setup with all three of these speed improvements (ResNet-50 + 100 RoIs + $\frac{1}{2}$ res.) achieves 15.2 frames per second with a V100 GPU, but still achieves strong results, especially for long-term tracking. The same setup with a ResNet-101 backbone instead of ResNet-50 (100 RoIs + $\frac{1}{2}$ res.) runs at $13.6$ frames per second and achieves excellent results and loses at most $1.6$ percentage points over these three datasets compared to the standard Siam R-CNN, while running almost three times as fast.

\subsection{Training Data Dependence} 
\begin{table*}[ht]
\centering
\setlength{\tabcolsep}{3pt}
\footnotesize
\begin{tabular}{ccccccccccccc}
\toprule 
\multirow{3}{*}{\rotatebox{90}{Eval.}} & \multirow{3}{*}{Method} & \multicolumn{6}{c}{Videos} & \multicolumn{3}{c}{Additional Images} & \multicolumn{1}{c}{Total} &
\tabularnewline
 &  & GOT-10k & ImageNet-Vid & LaSOT & YT-VOS & TrackingNet & YT-BB & COCO & ImageNet & ImageNet-Det & Videos + Add. Imgs \tabularnewline
 &  & 9k & 4k & 1k & 3k & 30k & 380k & 119k & 1281k & 457k & \tabularnewline
\midrule 
\multirow{3}{*}{\rotatebox{90}{ALL}} & Siam R-CNN & \checkmark & \checkmark & \checkmark & \checkmark &  &  & \checkmark &  &  & 18k+119k\tabularnewline
 & DiMP \& ATOM & \checkmark &  & \checkmark &  & \checkmark &  & \checkmark & \checkmark &  & 41k+1400k\tabularnewline
 & SiamRPN++ &  & \checkmark &  &  &  & \checkmark & \checkmark & \checkmark & \checkmark & 384k+1867k\tabularnewline
\midrule
\multirow{3}{*}{\rotatebox{90}{GOT}} & Siam R-CNN & \checkmark &  &  &  &  &  & \checkmark &  &  & 9k+119k\tabularnewline
 & DiMP \& ATOM & \checkmark &  &  &  &  &  &  & \checkmark &  & 9k+1281k\tabularnewline
 & SiamRPN++ & \checkmark &  &  &  &  &  & \checkmark & \checkmark &  & 9k+1400k\tabularnewline
\bottomrule
\end{tabular}
\caption{\label{tab:data} Training data used compared to some important recent methods \cite{Bhat19ICCV, Danelljan19CVPR, Li19CVPR}. Videos + Add. Imgs: Number of videos plus number of additional images not in videos. Eval.: Evaluation benchmark setup. ALL: All benchmarks except GOT-10k. GOT: Evaluate on GOT-10k, use only GOT-10k training data in addition to static images. ImageNet-Vid: ImageNet Video, YT-VOS: YouTube-VOS, YT-BB: YouTube BoundingBoxes.}
\end{table*}
\begin{table*}[ht]
\centering
\footnotesize
\begin{tabular}{ccccccccccccc}
\toprule 
 & ALL & BC & DEF & FM & IPR & IV & LR & MB & OCC & OPR & OV & SV\tabularnewline
\midrule
Siam R-CNN (ours) & \textbf{70.1} & \textbf{69.1} & 64.5 & \textbf{71.0} & \textbf{69.9} & 71.6 & 71.1 & \textbf{74.2} & \textbf{66.6} & \textbf{68.6} & \textbf{67.9} & \textbf{72.1}\tabularnewline
\midrule
No hard ex. min. & -1.7 & -3.0 & -1.3 & -2.2 & -2.2 & -2.3 & -5.3 & -2.1 & -1.7 & -2.1 & -1.2 & -1.7\tabularnewline
No TDPA (Argmax) & -6.3 & -7.9 & -1.5 & -5.6 & -9.5 & -3.3 & -14.0 & -5.8 & -4.6 & -8.4 & -6.8 & -7.2 \tabularnewline
No TDPA (Short-term) & -2.8 & -6.5 & -2.8 & -3.5 & -1.6 & -2.5 & -6.2 & -8.4 & -4.8 & -3.7 & -14.5 & -5.3 \tabularnewline
0 RoIs & -12.9 & -14.0 & -10.5 & -20.5 & -10.5 & -12.2 & -20.3 & -25.8 & -14.9 & -12.1 & -21.8 & -13.8 \tabularnewline
100 RoIs & -1.4 & -1.8 & \textbf{+1.2} & -3.6 & -3.1 & -0.8 & -7.8 & -4.5 & -1.6 & -2.2 & -4.2 & -2.1\tabularnewline
10000 RoIs & -0.5 & -0.8 & 0.0 & -0.1 & -0.6 & \textbf{+0.4} & \textbf{+0.8} & 0.0 & -0.7 & -0.7 & -0.3 & -0.6\tabularnewline
\midrule 
DiMP-50 \cite{Bhat19ICCV} & -1.3 & -5.3 & \textbf{+3.7} & -2.6 & -1.4 & -2.2 & -8.1 & -4.3 & -0.2 & -0.8 & -5.4 & -2.8\tabularnewline
SiamRPN++ \cite{Li19CVPR} & -0.4 & -0.0 & +1.7 & -2.0 & -0.5 & -0.3 & -5.1 & -3.0 & -0.3 & -0.3 & -3.1 & -2.4\tabularnewline
\bottomrule
\end{tabular}
\caption{\label{tab:attr}Per attribute ablation for Success (AUC) on OTB2015. The first row shows the performance of the full model, and all other rows show the absolute difference to the full model. All: All Videos, BC: Background Clutters, DEF: Deformation, FM: Fast Motion, IPR: In-Plane Rotation, IV: Illumination Variation, LR: Low Resolution, MB: Motion Blur, OCC: Occlusion, OPR: Out-of-Plane Rotation, OV: Out-of-View, SV: Scale Variation.}
\end{table*}
We show how our training data compares to the data used by some important recent methods in Table~\ref{tab:data}. We use more video `datasets', but actually use far less data. DiMP-50 \cite{Bhat19ICCV} and ATOM \cite{Danelljan19CVPR} both use 2.28 times more videos. SiamRPN++ \cite{Li19CVPR} uses 21.3 times more. All three use ImageNet and train on COCO by creating artificial videos using augmentations (we use COCO without this complex kind of augmentation). The only dataset we use which is not used by any of the other considered methods is YouTube-VOS \cite{Xu18ECCV}, as we also evaluate on VOS benchmarks. For GOT-10k \cite{Huang18Arxiv} evaluation, where only GOT-10k video training data is allowed, all other methods also use static images from ImageNet. For SiamRPN++, we use COCO but not ImageNet. This shows that our strong results are not due to the amount of training data.

\subsection{Per-Attribute Analysis}
Table~\ref{tab:attr} shows a per-attribute ablation on OTB2015. Hard example mining improves results over all attributes and is particularly helpful for low resolution (LR) and background clutter (BC).

Our Tracklet Dynamic Programming Algorithm (TDPA) models spatio-temporal consistency cues only where it is likely that there are consistent predictions, by building up tracklets.  It corrects itself immediately after disappearance by tracking all objects simultaneously, and determining the most likely set of previous tracklets for the object online using dynamic programming. For the Out-of-View (OV) attribute (the target disappears in the video), TDPA significantly outperforms Short-term, which is unable to rely on spatio-temporal consistency cues during disappearance and thus often fails and performs worse than Argmax. %
TDPA tackles this problem of object disappearance, by using a dynamic and robust type of spatio-temporal consistency cues and also increases robustness against distractors, improving results across all attributes.

\subsection{Fine-tuning Analysis}
\begin{figure}
\includegraphics[width=0.9\columnwidth]{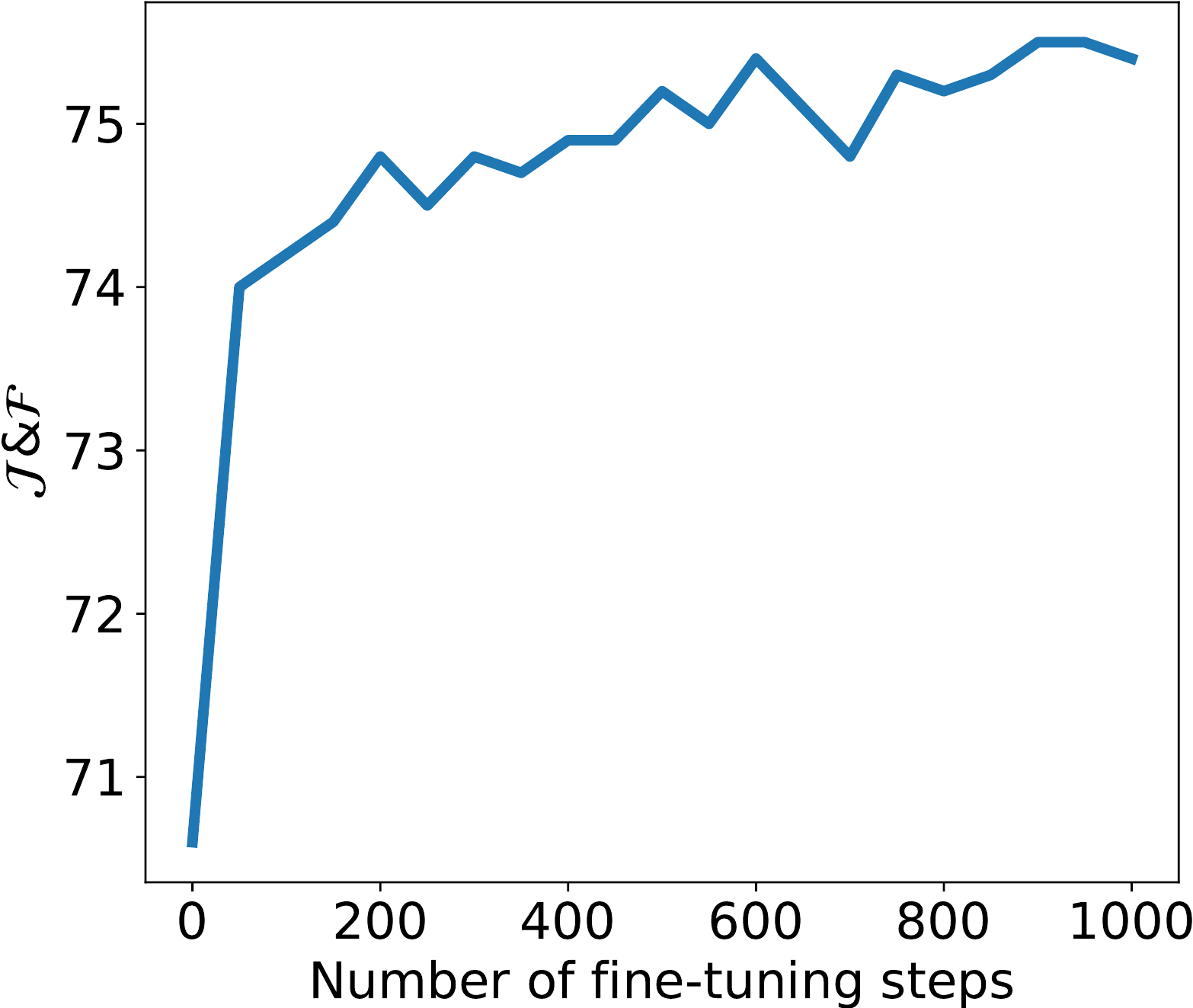}
\caption{\label{fig:ft}Segmentation quality on the DAVIS 2017 validation set depending on the number of fine-tuning steps.}
\end{figure}
Figure~\ref{fig:ft} shows the result of Siam R-CNN with a different number of fine-tuning steps for Box2Seg. For the fine-tuned Box2Seg variant in the main paper, we used 300 steps which yields a speed-accuracy trade-off of $74.8 \ \mathcal{J}\&\mathcal{F}$ with a run-time of 1 FPS (compared to $70.6 \ \mathcal{J}\&\mathcal{F}$ and 3.1 FPS without fine-tuning). Note that here the timing is per frame, and not per object, so that the run-time without fine-tuning is longer than for single-object tracking scenarios. When increasing the number of fine-tuning steps to 1000, Siam R-CNN achieves a $\mathcal{J}\&\mathcal{F}$ score of $75.4$ with a run-time of 0.38 FPS.

\section{Further Experimental Results}
In addition to the $11$ benchmarks that we presented in the main paper, we present here results on eight further benchmarks, of which five are short-term tracking benchmarks, one is a long-term tracking benchmark and two are video object segmentation benchmarks. For all benchmarks (except for the three VOT benchmarks) we use exactly the same tracking parameters. This is in contrast to many other methods which have parameters explicitly tuned for each dataset. This shows the generalization ability of our tracker to many different scenarios. For the three VOT datasets we use the short-term variant of the tracking parameters (with the same parameters across these three benchmarks).

\subsection{Further Short-Term Tracking Evaluation}
In the main paper we presented short-term tracking results on OTB2015 \cite{Wu15TPAMI}, UAV123 \cite{Mueller16ECCV}, NfS \cite{Galoogahi17ICCV}, TrackingNet \cite{Muller18ECCV}, VOT2018 (the same as VOT2017) \cite{Kristan18ECCVW}, and GOT-10k \cite{Huang18Arxiv}. Here in the supplemental material we present results on five further short-term tracking benchmarks. 
These further benchmarks are OTB-50 \cite{Wu15TPAMI}, OTB2013 \cite{Wu13CVPR}, VOT2015 \cite{Kristan15ICCVW}, VOT2016 \cite{Kristan16ECCVW}, and TempleColor128 (TC128) \cite{Liang15TIP}.

\PAR{OTB-50.} We evaluate on the OTB-50 benchmark \cite{Wu15TPAMI} (50 videos, 539 frames average length). This dataset is a subset of OTB2015 using exactly half of the sequences. It is evaluated with the same evaluation measures as OTB2015. Tab.~\ref{tab:otb50} compares our results to five state-of-the-art trackers.
Siam R-CNN achieves $66.3$ AUC, which outperforms the previous best published results by ACT \cite{Chen18ECCV_Actor} by $0.6$ percentage points. %

\PAR{OTB2013.} We evaluate on the OTB2013 benchmark \cite{Wu13CVPR} (51 videos, 578 frames average length). This dataset is a predecessor to OTB2015 and is evaluated with the same evaluation measures. Tab.~\ref{tab:otb2013} compares our results to five state-of-the-art trackers.
Siam R-CNN achieves $70.4$ AUC, which is comparable to state-of-the-art trackers while being $1.8$ percentage points behind the best published results by GFS-DCF \cite{Xu19ICCV}. 

\PAR{TC128.} We evaluate on the TempleColor128 (TC128) benchmark \cite{Liang15TIP} (128 videos, 429 frames average length). This dataset is also evaluated using the OTB2015 evaluation measures. Tab.~\ref{tab:tc128} compares our results to five state-of-the-art trackers.
Siam R-CNN achieves $60.1$ AUC, which is comparable to state-of-the-art trackers while being slightly inferior to the best published results by UPDT \cite{Bhat18ECCV} by $2.1$ percentage points. 

\begin{table}[t]
\centering{}%
\setlength{\tabcolsep}{2pt}
\footnotesize
\begin{tabular}{lcccccc}
\toprule
 & {\footnotesize{}SA-Siam} & {\footnotesize{}SINT++} & {\footnotesize{}RTINet} & {\footnotesize{}SPM} & {\footnotesize{}ACT} & {\footnotesize{}Siam}\tabularnewline
 & {\footnotesize{}\cite{He18CVPR_Twofold}} & {\footnotesize{}\cite{Wang18CVPR_Adversarial}} & {\footnotesize{}\cite{Yao18ECCV}} & {\footnotesize{}\cite{wang2019spm}} & {\footnotesize{}\cite{Chen18ECCV_Actor}} & {\footnotesize{}R-CNN} \tabularnewline 
\midrule
Success AUC & {\footnotesize{}61.0} & {\footnotesize{}62.4} & {\footnotesize{}63.7} & {\footnotesize{}65.3} & {\footnotesize{}65.7} & {\footnotesize{}\textbf{66.3}}\tabularnewline
\bottomrule
\end{tabular}
\caption{\label{tab:otb50}Results on OTB-50 \cite{Wu15TPAMI}.}
\end{table}

\begin{table}[t]
\centering{}%
\setlength{\tabcolsep}{2pt}
\footnotesize
\begin{tabular}{lcccccc}
\toprule
 & {\footnotesize{}RPCF} & {\footnotesize{}SACF} & {\footnotesize{}MCCT} & {\footnotesize{}DRT} & {\footnotesize{}GFS-DCF} & {\footnotesize{}Siam}\tabularnewline
 & {\footnotesize{}\cite{Sun19CVPR}} & {\footnotesize{}\cite{Zhang18ECCV_Aligned}} & {\footnotesize{}\cite{Wang18CVPR}} & {\footnotesize{}\cite{Sun18CVPR_Reliability}} & {\footnotesize{}\cite{Xu19ICCV}} & {\footnotesize{}R-CNN} \tabularnewline 
\midrule
Success AUC & {\footnotesize{}71.3} & {\footnotesize{}71.3} & {\footnotesize{}71.4} & {\footnotesize{}72.0} & {\footnotesize{}\textbf{72.2}} & {\footnotesize{}70.4}\tabularnewline
\bottomrule
\end{tabular}
\caption{\label{tab:otb2013}Results on OTB2013 \cite{Wu13CVPR}.}
\end{table}

\begin{table}[t]
\centering{}%
\setlength{\tabcolsep}{2pt}
\footnotesize
\begin{tabular}{lcccccc}
\toprule
 & {\footnotesize{}MCCT} & {\footnotesize{}STRCF} & {\footnotesize{}RTINet} & {\footnotesize{}ASRCF} & {\footnotesize{}UPDT} & {\footnotesize{}Siam}\tabularnewline
 & {\footnotesize{}\cite{Wang18CVPR}} & {\footnotesize{}\cite{Li18CVPR_Regularized}} & {\footnotesize{}\cite{Yao18ECCV}} & {\footnotesize{}\cite{Dai19CVPR}} & {\footnotesize{}\cite{Bhat18ECCV}} & {\footnotesize{}R-CNN} \tabularnewline 
\midrule
Success AUC & {\footnotesize{}59.6} & {\footnotesize{}60.1} & {\footnotesize{}60.2} & {\footnotesize{}60.3} & {\footnotesize{}\textbf{62.2}} & {\footnotesize{}60.1}\tabularnewline
\bottomrule
\end{tabular}
\caption{\label{tab:tc128}Results on TC128 \cite{Liang15TIP}.}
\end{table}

\begin{table}[t]
\centering{}%
\setlength{\tabcolsep}{2pt}
\footnotesize
\begin{tabular}{lcccccc}
\toprule
 & {\footnotesize{}FlowTrack} & {\footnotesize{}SACF} & {\footnotesize{}SiamRPN} & {\footnotesize{}SiamDW} & {\footnotesize{}DaSiamRPN} & {\footnotesize{}Ours}\tabularnewline
 & {\footnotesize{}\cite{Zhu18CVPR}} & {\footnotesize{}\cite{Zhang18ECCV_Aligned}} & {\footnotesize{}\cite{Li18CVPRSiamRPN}} & {\footnotesize{}\cite{Zhang19CVPR}} & {\footnotesize{}\cite{Zhu18ECCV}} & {\scriptsize{}(short-t.)} \tabularnewline 
\midrule
EAO & {\footnotesize{}34.1} & {\footnotesize{}34.3} & {\footnotesize{}35.8} & {\footnotesize{}38.0} & {\footnotesize{}44.6} & {\footnotesize{}\textbf{45.4}}\tabularnewline
\bottomrule
\end{tabular}
\caption{\label{tab:vot15}Results on VOT2015 \cite{Kristan15ICCVW}.}
\end{table}

\begin{table}[t]
\centering{}%
\setlength{\tabcolsep}{2pt}
\footnotesize
\begin{tabular}{lcccccc}
\toprule
 & {\footnotesize{}DaSiamRPN} & {\footnotesize{}SPM} & {\footnotesize{}DRT} & {\footnotesize{}SiamMask} & {\footnotesize{}UpdateNet} & {\footnotesize{}Ours}\tabularnewline
 & {\footnotesize{}\cite{Zhu18ECCV}} & {\footnotesize{}\cite{wang2019spm}} & {\footnotesize{}\cite{Sun18CVPR_Reliability}} & {\footnotesize{}\cite{Wang19CVPR}} & {\footnotesize{}\cite{Zhang19ICCV}} & {\scriptsize{}(short-t.)} \tabularnewline 
\midrule
EAO & {\footnotesize{}41.1} & {\footnotesize{}43.4} & {\footnotesize{}44.2} & {\footnotesize{}44.2} & {\footnotesize{}\textbf{48.1}} & {\footnotesize{}46.5}\tabularnewline
\bottomrule
\end{tabular}
\caption{\label{tab:vot16}Results on VOT2016 \cite{Kristan16ECCVW}.}
\end{table}

\begin{table}[t]
\centering{}%
\setlength{\tabcolsep}{2pt}
\footnotesize
\begin{tabular}{lcccccc}
\toprule 
 & {\footnotesize{}SiamFC} & {\footnotesize{}PTAV} & {\footnotesize{}ECO} & {\footnotesize{}SiamRPN} & {\footnotesize{}{}DaSiamRPN } & {\footnotesize{}{}Siam}\tabularnewline
 & {\footnotesize{}\cite{Bertinetto2016ECCV}} & {\footnotesize{}\cite{Fan17CVPR}} & {\footnotesize{}\cite{Danelljan17CVPR}} & {\footnotesize{}\cite{Li18CVPRSiamRPN}} & {\footnotesize{}{}\cite{Zhu18ECCV} } & {\footnotesize{}{}R-CNN }\tabularnewline
\midrule 
{\footnotesize{}Success AUC } & {\footnotesize{}39.9} & {\footnotesize{}42.3} & {\footnotesize{}43.5} & {\footnotesize{}45.4} & {\footnotesize{}{}61.7 } & {\footnotesize{}{}}\textbf{\footnotesize{}67.2}\tabularnewline
\bottomrule
\end{tabular}
\caption{\label{tab:uav20L}Results on UAV20L \cite{Mueller16ECCV}.}
\end{table}

\PAR{VOT2015.} We evaluate on the VOT2015 benchmark \cite{Kristan15ICCVW} (60 videos, 358 frames average length). This is evaluated with the same evaluation measures as VOT2018. Tab.~\ref{tab:vot15} compares our results to five state-of-the-art trackers.
The short-term version of Siam R-CNN achieves $45.4$ EAO, which outperforms the previous best published results by DaSiamRPN \cite{Zhu18ECCV} by $0.8$ percentage points. %

\PAR{VOT2016.} We evaluate on the VOT2016 benchmark \cite{Kristan16ECCVW} (60 videos, 358 frames average length). This is also evaluated with the same evaluation measures as VOT2018. Tab.~\ref{tab:vot16} compares our results to five state-of-the-art trackers.
The short-term version of Siam R-CNN achieves $46.5$ EAO, which outperforms all previous published results except those of UpdateNet \cite{Zhang19ICCV} which outperforms our results by $1.6$ percentage points. %

\subsection{Further Long-Term Tracking Evaluation}
We evaluate on one further long-term tracking dataset, in addition to the three benchmarks presented in the main paper. 

\PAR{UAV20L.} We evaluate on the UAV20L benchmark \cite{Mueller16ECCV} (20 videos, 2934 frames average length). This dataset contains 20 of the 123 sequences of UAV123, however each of these sequences extends for many more frames than the equivalent sequence in the UAV123 version. It is also evaluated with the same evaluation measures as OTB2015. Tab.~\ref{tab:uav20L} compares our results to five state-of-the-art trackers.
Siam R-CNN achieves $67.2$ AUC, which outperforms the previous best published results by DaSiamRPN \cite{Zhu18ECCV} by $5.5$ percentage points, which further highlights the ability of Siam R-CNN to perform long-term tracking. %

\subsection{Further Video Object Segmentation Evaluation}
\begin{table}[t]
\centering{}{\footnotesize{}}%
\scalebox{0.94}{
\setlength{\tabcolsep}{2pt}
\begin{tabular}{lcccccccc}
\toprule 
{\footnotesize{}Init}\hspace{-4mm} & {\footnotesize{}Method} & {\footnotesize{}FT} & {\footnotesize{}M} & {\footnotesize{}$\mathcal{J}$\&$\mathcal{F}$} & {\footnotesize{}$\mathcal{J}$} & {\footnotesize{}$\mathcal{F}$} & {\footnotesize{}$\mathcal{J}_{box}$} & {\footnotesize{}t(s)}\tabularnewline

\midrule

\parbox[t]{2mm}{\multirow{6}{*}{\rotatebox[origin=c]{90}{bbox}}} &
{\footnotesize{}\textbf{Siam R-CNN (ours)}} & {\footnotesize{}\ding{55}} & {\footnotesize{}\ding{55}} & {\footnotesize{}$\mathbf{70.6}$} & {\footnotesize{}$66.1$} & {\footnotesize{}$\mathbf{75.0}$} & {\footnotesize{}$\mathbf{78.3}$} & {\footnotesize{}$0.32$}\tabularnewline

& {\footnotesize{}\textbf{Siam R-CNN (fastest)}} & {\footnotesize{}\ding{55}} & {\footnotesize{}\ding{55}} & {\footnotesize{}$70.5$} & {\footnotesize{}$\mathbf{66.4}$} & {\footnotesize{}$74.6$} & {\footnotesize{}$76.9$} & {\footnotesize{}$0.12$}\tabularnewline

&
{\footnotesize{}SiamMask \cite{Wang19CVPR}} & {\footnotesize{}\ding{55}} & {\footnotesize{}\ding{55}} & {\footnotesize{}$55.8$} & {\footnotesize{}$54.3$} & {\footnotesize{}$58.5$} &  {\footnotesize{}$64.3$} & {\footnotesize{}$\mathbf{0.06^{\dagger}}$}\tabularnewline

&{\footnotesize{}SiamMask \cite{Wang19CVPR} (Box2Seg)} & {\footnotesize{}\ding{55}} & {\footnotesize{}\ding{55}} & {\footnotesize{}$63.3$} & {\footnotesize{}$59.5$} & {\footnotesize{}$67.3$} &  {\footnotesize{}$64.3$} & {\footnotesize{}$0.11$}\tabularnewline

&{\footnotesize{}SiamRPN++ \cite{Li19CVPR} (Box2Seg)} & {\footnotesize{}\ding{55}} & {\footnotesize{}\ding{55}} & {\footnotesize{}$61.6$} & {\footnotesize{}$56.8$} & {\footnotesize{}$66.3$} &  {\footnotesize{}$64.0$} & {\footnotesize{}$0.11$}\tabularnewline

&{\footnotesize{}DiMP-50 \cite{Bhat19ICCV} (Box2Seg)} & {\footnotesize{}\ding{55}} & {\footnotesize{}\ding{55}} & {\footnotesize{}$63.7$} & {\footnotesize{}$60.1$} & {\footnotesize{}$67.3$} &  {\footnotesize{}$65.6$} & {\footnotesize{}$0.10$}\tabularnewline

\midrule

\parbox{2mm}{\multirow{6}{*}{\rotatebox[origin=c]{90}{mask}}}&{\footnotesize{}STM-VOS \cite{Oh19ICCV}} & {\footnotesize{}\ding{55}} & {\footnotesize{}\ding{51}} & {\footnotesize{}$\mathbf{81.8}$} & {\footnotesize{}$\mathbf{79.2}$} & {\footnotesize{}$\mathbf{84.3}$} & {\footnotesize{}$-$} & {\footnotesize{}$0.32^{\dagger}$}\tabularnewline

&{\footnotesize{}FEELVOS \cite{Voigtlaender19CVPR}} & {\footnotesize{}\ding{55}} & {\footnotesize{}\ding{51}} & {\footnotesize{}$71.5$} & {\footnotesize{}$69.1$} & {\footnotesize{}$74.0$} & {\footnotesize{}$71.4$} & {\footnotesize{}$0.51$}\tabularnewline

&{\footnotesize{}RGMP \cite{Oh18CVPR}} & {\footnotesize{}\ding{55}} & {\footnotesize{}\ding{51}} & {\footnotesize{}$66.7$} & {\footnotesize{}$64.8$} & {\footnotesize{}$68.6$} & {\footnotesize{}66.5} & \textbf{\footnotesize{}$\mathbf{0.28^{\dagger}}$}\tabularnewline

&{\footnotesize{}VideoMatch \cite{Hu18ECCV}} & {\footnotesize{}\ding{55}} & {\footnotesize{}\ding{51}} & {\footnotesize{}$62.4$} & {\footnotesize{}$56.5$} & {\footnotesize{}$68.2$} &  {\footnotesize{}$-$} & {\footnotesize{}$0.35$}\tabularnewline

&{\footnotesize{}FAVOS \cite{Cheng18CVPR}} & {\footnotesize{}\ding{55}} & {\footnotesize{}\ding{51}} & {\footnotesize{}$58.2$} & {\footnotesize{}$54.6$} & {\footnotesize{}$61.8$} & {\footnotesize{}68.0} & {\footnotesize{}$1.2^{\dagger}$}\tabularnewline

&{\footnotesize{}OSMN \cite{Yang18CVPR}} & {\footnotesize{}\ding{55}} & {\footnotesize{}\ding{51}} &  {\footnotesize{}$54.8$} & {\footnotesize{}$52.5$} & {\footnotesize{}$57.1$} & {\footnotesize{}60.1} & \textbf{\footnotesize{}$\mathbf{0.28^{\dagger}}$}\tabularnewline

\midrule

\parbox[t]{2mm}{\multirow{7}{*}{\rotatebox[origin=c]{90}{mask+ft}}}&{\footnotesize{}PReMVOS \cite{Luiten18ACCV}} & {\footnotesize{}\ding{51}} & {\footnotesize{}\ding{51}} & {\footnotesize{}$\mathbf{77.8}$} & {\footnotesize{}$\mathbf{73.9}$} & {\footnotesize{}$\mathbf{81.7}$} & {\footnotesize{}$\mathbf{81.4}$} & {\footnotesize{}$37.6$}\tabularnewline

&{\footnotesize{}\textbf{Ours (Fine-t. Box2Seg)}} & {\footnotesize{}\ding{51}} & {\footnotesize{}\ding{51}} & {\footnotesize{}$74.8$} & {\footnotesize{}$69.3$} & {\footnotesize{}$80.2$} & {\footnotesize{}$78.3$} & {\footnotesize{}$\mathbf{1.0}$}\tabularnewline

&{\footnotesize{}DyeNet \cite{Li18ECCV}} & {\footnotesize{}\ding{51}} & {\footnotesize{}\ding{51}} & {\footnotesize{}$74.1$} & {\footnotesize{}$-$} & {\footnotesize{}$-$} & {\footnotesize{}$-$} & {\footnotesize{}$9.32^{\dagger}$}\tabularnewline

&{\footnotesize{}OSVOS-S \cite{Maninis18TPAMI}} & {\footnotesize{}\ding{51}} & {\footnotesize{}\ding{51}} & {\footnotesize{}$68.0$} & {\footnotesize{}$64.7$} & {\footnotesize{}$71.3$} & {\footnotesize{}$68.4$} &{\footnotesize{}$9^{\dagger}$}\tabularnewline

&{\footnotesize{}CINM \cite{Bao18CVPR}} & {\footnotesize{}\ding{51}} &  {\footnotesize{}\ding{51}} & {\footnotesize{}$67.5$} & {\footnotesize{}$64.5$} & {\footnotesize{}$70.5$} & {\footnotesize{}$72.9$} & {\footnotesize{}$>\!120$}\tabularnewline

&{\footnotesize{}OnAVOS \cite{voigtlaender17BMVC}} & {\footnotesize{}\ding{51}} & {\footnotesize{}\ding{51}} & {\footnotesize{}$63.6$} & {\footnotesize{}$61.0$} & {\footnotesize{}$66.1$} &  {\footnotesize{}$66.3$} & {\footnotesize{}$26$}\tabularnewline

&{\footnotesize{}OSVOS \cite{OSVOS}} & {\footnotesize{}\ding{51}} & {\footnotesize{}\ding{51}} & {\footnotesize{}$60.3$} & {\footnotesize{}$56.6$} & {\footnotesize{}$63.9$} & {\footnotesize{}$57.0$} & {\footnotesize{}$18^{\dagger}$}\tabularnewline

\midrule

&{\footnotesize{}GT boxes (Box2Seg)} & {\footnotesize{}\ding{55}} & {\footnotesize{}\ding{55}} & {\footnotesize{}$82.6$} & {\footnotesize{}$79.3$} & {\footnotesize{}$85.8$} & {\footnotesize{}$100.0$} & {\footnotesize{}$-$}\tabularnewline

&{\footnotesize{}GT boxes (Fine-t. Box2Seg)} & {\footnotesize{}\ding{51}} & {\footnotesize{}\ding{51}} & {\footnotesize{}$86.2$} & {\footnotesize{}$81.8$} & {\footnotesize{}$90.5$} &  {\footnotesize{}$100.0$} & {\footnotesize{}$-$}\tabularnewline

\bottomrule
\end{tabular}}{\footnotesize{}\caption{\label{tab:results-davis17-supp}Results on the DAVIS 2017
validation set. FT: fine-tuning, M: using the first-frame masks, t(s): time per frame in seconds. $\dagger$: timing extrapolated from DAVIS 2016 assuming
linear scaling in the number of objects. Siam R-CNN (fastest) denotes Siam R-CNN with ResNet-50 backbone, half input resolution, and 100 RoIs from the RPN%
.}
}{\footnotesize \par}
\end{table}

\begin{table}[t]
\centering{}{\footnotesize{}}%
\scalebox{0.94}{
\setlength{\tabcolsep}{2pt}
\begin{tabular}{lcccccccc}
\toprule 
{\footnotesize{}Init}\hspace{-4mm} & {\footnotesize{}Method} & {\footnotesize{}FT} & {\footnotesize{}M} & {\footnotesize{}$\mathcal{J}$\&$\mathcal{F}$} & {\footnotesize{}$\mathcal{J}$} & {\footnotesize{}$\mathcal{F}$} & {\footnotesize{}$\mathcal{J}_{box}$} & {\footnotesize{}t(s)}\tabularnewline
\midrule

\parbox[t]{2mm}{\multirow{4}{*}{\rotatebox[origin=c]{90}{bbox}}} & {\footnotesize{}\textbf{Siam R-CNN (ours)}} & {\footnotesize{}\ding{55}} & {\footnotesize{}\ding{55}} & \textbf{\footnotesize{}$78.6$} & {\footnotesize{}$76.8$} & {\footnotesize{}$80.4$} & {\footnotesize{}$\mathbf{86.6}$} & {\footnotesize{}$0.24$}\tabularnewline

& {\footnotesize{}\textbf{Siam R-CNN (fastest)}} & {\footnotesize{}\ding{55}} & {\footnotesize{}\ding{55}} &  \textbf{\footnotesize{}$\mathbf{79.0}$} & {\footnotesize{}$\mathbf{77.4}$} & {\footnotesize{}$\mathbf{80.6}$} & {\footnotesize{}$85.0$} & {\footnotesize{}$0.08$}\tabularnewline

& {\footnotesize{}SiamMask \cite{Wang19CVPR}} & {\footnotesize{}\ding{55}} & {\footnotesize{}\ding{55}} & \textbf{\footnotesize{}$69.8$} & {\footnotesize{}$71.7$} & {\footnotesize{}$67.8$} & {\footnotesize{}$73.3$} & {\footnotesize{}$\mathbf{0.03}$}\tabularnewline

& {\footnotesize{}SiamMask \cite{Wang19CVPR} (Box2Seg)} & {\footnotesize{}\ding{55}} & {\footnotesize{}\ding{55}} & {\footnotesize{}$75.9$} & {\footnotesize{}$75.6$} & {\footnotesize{}$76.3$} & {\footnotesize{}$73.3$} & {\footnotesize{}$0.06$}\tabularnewline

\midrule

\parbox[t]{2mm}{\multirow{7}{*}{\rotatebox[origin=c]{90}{mask}}} & {\footnotesize{}STM-VOS \cite{Oh19ICCV}} & {\footnotesize{}\ding{55}} & {\footnotesize{}\ding{51}} & \textbf{\footnotesize{}$\mathbf{89.3}$} & \textbf{\footnotesize{}$\mathbf{88.7}$} & {\footnotesize{}$\mathbf{89.9}$} & {\footnotesize{}$-$} & \textbf{\footnotesize{}$0.16$}\tabularnewline

& {\footnotesize{}RGMP \cite{Oh18CVPR}} & {\footnotesize{}\ding{55}} & {\footnotesize{}\ding{51}} & \textbf{\footnotesize{}$81.8$} & \textbf{\footnotesize{}$81.5$} & {\footnotesize{}$82.0$} & {\footnotesize{}$79.3$} & \textbf{\footnotesize{}$\mathbf{0.14}$}\tabularnewline

& {\footnotesize{}FEELVOS \cite{Voigtlaender19CVPR}} & {\footnotesize{}\ding{55}} & {\footnotesize{}\ding{51}} & \textbf{\footnotesize{}$81.7$} &{\footnotesize{}$81.1$} & {\footnotesize{}$82.2$} & {\footnotesize{}$80.2$} & {\footnotesize{}$0.45$}\tabularnewline

& {\footnotesize{}FAVOS \cite{Cheng18CVPR}} & {\footnotesize{}\ding{55}} & {\footnotesize{}\ding{51}} &  {\footnotesize{}$81.0$} & {\footnotesize{}$82.4$} & {\footnotesize{}$79.5$} & {\footnotesize{}$\mathbf{83.1}$} & {\footnotesize{}$0.6$}\tabularnewline

& {\footnotesize{}VideoMatch \cite{Hu18ECCV}} & {\footnotesize{}\ding{55}} & {\footnotesize{}\ding{51}} & {\footnotesize{}$80.9$} & {\footnotesize{}$81.0$} & {\footnotesize{}$80.8$} & {\footnotesize{}$-$} & {\footnotesize{}$0.32$}\tabularnewline

& {\footnotesize{}PML \cite{Chen18CVPR}} & {\footnotesize{}\ding{55}} & {\footnotesize{}\ding{51}} & {\footnotesize{}$77.4$} & {\footnotesize{}$75.5$} & {\footnotesize{}$79.3$} & {\footnotesize{}$75.9$} & {\footnotesize{}$0.28$}\tabularnewline

& {\footnotesize{}OSMN \cite{Yang18CVPR}} & {\footnotesize{}\ding{55}} & {\footnotesize{}\ding{51}} & {\footnotesize{}$73.5$} & {\footnotesize{}$74.0$} & {\footnotesize{}$72.9$} & {\footnotesize{}$71.8$} & {\footnotesize{}$\mathbf{0.14}$}\tabularnewline

\midrule 

\parbox[t]{2mm}{\multirow{7}{*}{\rotatebox[origin=c]{90}{mask+ft}}} & {\footnotesize{}\textbf{Ours (Fine-t. Box2Seg)}} & {\footnotesize{}\ding{51}} & {\footnotesize{}\ding{51}} &  {\footnotesize{}$\mathbf{87.1}$} & {\footnotesize{}$85.3$} & {\footnotesize{}$\mathbf{88.8}$} & {\footnotesize{}$86.6$} & {\footnotesize{}$\mathbf{0.56}$}\tabularnewline

& {\footnotesize{}PReMVOS \cite{Luiten18ACCV}} & {\footnotesize{}\ding{51}} & {\footnotesize{}\ding{51}} & {\footnotesize{}$86.8$} & {\footnotesize{}$84.9$} & {\footnotesize{}$88.6$} & {\footnotesize{}$\mathbf{89.9}$} & {\footnotesize{}$32.8$}\tabularnewline

& {\footnotesize{}DyeNet \cite{Li18ECCV}} & {\footnotesize{}\ding{51}} & {\footnotesize{}\ding{51}} & {\footnotesize{}$-$} & {\footnotesize{}$\mathbf{86.2}$} & {\footnotesize{}$-$} & {\footnotesize{}$-$} & {\footnotesize{}$4.66$}\tabularnewline

& {\footnotesize{}OSVOS-S \cite{Maninis18TPAMI}} & {\footnotesize{}\ding{51}} & {\footnotesize{}\ding{51}} & {\footnotesize{}$86.5$} & {\footnotesize{}$85.6$} & {\footnotesize{}$87.5$} & {\footnotesize{}$84.4$} & {\footnotesize{}$4.5$}\tabularnewline

& {\footnotesize{}OnAVOS \cite{voigtlaender17BMVC}} & {\footnotesize{}\ding{51}} & {\footnotesize{}\ding{51}} & {\footnotesize{}$85.0$} & {\footnotesize{}$85.7$} & {\footnotesize{}$84.2$} & {\footnotesize{}$84.1$} & {\footnotesize{}$13$}\tabularnewline

& {\footnotesize{}CINM \cite{Bao18CVPR}} & {\footnotesize{}\ding{51}} & {\footnotesize{}\ding{51}} & {\footnotesize{}$84.2$} & {\footnotesize{}$83.4$} & {\footnotesize{}$85.0$} & {\footnotesize{}$83.6$} & {\footnotesize{}$>120$}\tabularnewline

& {\footnotesize{}OSVOS \cite{OSVOS}} & {\footnotesize{}\ding{51}} & {\footnotesize{}\ding{51}} & {\footnotesize{}$80.2$} & {\footnotesize{}$79.8$} & {\footnotesize{}$80.6$} & {\footnotesize{}$76.0$} & {\footnotesize{}$9$}\tabularnewline

\midrule

& {\footnotesize{}GT boxes (Box2Seg)} & {\footnotesize{}\ding{55}} & {\footnotesize{}\ding{55}} & \textbf{\footnotesize{}$80.5$} & {\footnotesize{}$79.1$} & {\footnotesize{}$81.9$} & {\footnotesize{}$100.0$} & {\footnotesize{}$-$}\tabularnewline

& {\footnotesize{}GT boxes (Fine-t. Box2Seg)} & {\footnotesize{}\ding{51}} & {\footnotesize{}\ding{51}} & {\footnotesize{}$89.0$} & {\footnotesize{}$87.6$} & {\footnotesize{}$90.5$} & {\footnotesize{}$100.0$} & {\footnotesize{}$-$}\tabularnewline

\bottomrule
\end{tabular}}{\footnotesize{}\caption{\label{tab:results-davis16}Quantitative results on the DAVIS 2016
validation set. FT denotes fine-tuning, M denotes using the first-frame mask, and t(s) denotes time per frame
in seconds. Siam R-CNN (fastest) denotes Siam R-CNN with ResNet-50 backbone, half input resolution, and 100 RoIs from the RPN%
.}
}
\end{table}

\begin{table}[t]
\centering{}{\footnotesize{}{}}\scalebox{1.00}{ %
\setlength{\tabcolsep}{3pt}
\begin{tabular}{lccccccc}
\toprule 
{\footnotesize{}Init}\hspace{-4mm} & {\footnotesize{}Method} & {\footnotesize{}{}FT}  & {\footnotesize{}{}M}  & {\footnotesize{}{}$\mathcal{J}$\&$\mathcal{F}$}  & {\footnotesize{}{}$\mathcal{J}$}  & {\footnotesize{}{}$\mathcal{F}$}  & {\footnotesize{}{}t(s)}\tabularnewline
\midrule 

\parbox[t]{2mm}{\multirow{3}{*}{\rotatebox[origin=c]{90}{bbox}}} &{\footnotesize{}{}\textbf{Siam R-CNN (ours)}}  & {\footnotesize{}{}\ding{55}}  & {\footnotesize{}{}\ding{55}}  & {\footnotesize{}{}$\mathbf{53.3}$}  & {\footnotesize{}{}$\mathbf{48.1}$}  & {\footnotesize{}{}$\mathbf{58.6}$}  & {\footnotesize{}{}0.44}\tabularnewline

&{\footnotesize{}{}\textbf{Siam R-CNN (fastest)}}  & {\footnotesize{}{}\ding{55}}  & {\footnotesize{}{}\ding{55}}  & {\footnotesize{}{}$51.6$}  & {\footnotesize{}{}$46.3$}  & {\footnotesize{}{}$56.8$}  & {\footnotesize{}{}$0.16$}\tabularnewline

&{\footnotesize{}{}SiamMask \cite{Wang19CVPR}}  & {\footnotesize{}{}\ding{55}}  & {\footnotesize{}{}\ding{55}}  & {\footnotesize{}{}$43.2$}  & {\footnotesize{}{}$40.6$}  & {\footnotesize{}{}$45.8$}  & {\footnotesize{}{}$\mathbf{0.09}^{\dagger}$}\tabularnewline

\midrule 

\parbox[t]{2mm}{\multirow{3}{*}{\rotatebox[origin=c]{90}{mask}}} &
{\footnotesize{}{}STM-VOS \cite{Oh19ICCV}}  & {\footnotesize{}{}\ding{55}}  & {\footnotesize{}{}\ding{51}}  & {\footnotesize{}{}$\mathbf{72.3}$}  & {\footnotesize{}{}$\mathbf{69.3}$}  & {\footnotesize{}{}$\mathbf{75.2}$}  & {\footnotesize{}{}$\mathbf{0.48^{\dagger}}$}\tabularnewline

&{\footnotesize{}{}RGMP \cite{Oh18CVPR}}  & {\footnotesize{}{}\ding{55}}  & {\footnotesize{}{}\ding{51}}  & {\footnotesize{}{}$52.9$}  & {\footnotesize{}{}$51.4$}  & {\footnotesize{}{}$54.4$}  & {\footnotesize{}{}$0.42^{\dagger}$}\tabularnewline

&{\footnotesize{}{}FEELVOS \cite{Voigtlaender19CVPR}}  & {\footnotesize{}{}\ding{55}}  & {\footnotesize{}{}\ding{51}}  & {\footnotesize{}{}$57.8$}  & {\footnotesize{}{}$55.2$}  & {\footnotesize{}{}$60.5$}  & {\footnotesize{}{}$0.54$}\tabularnewline

\midrule 

\parbox[t]{2mm}{\multirow{3}{*}{\rotatebox[origin=c]{90}{mask+ft}}} &
{\footnotesize{}{}PReMVOS \cite{Luiten18ACCV}}  & {\footnotesize{}{}\ding{51}}  & {\footnotesize{}{}\ding{51}}  & {\footnotesize{}{}$\mathbf{71.6}$}  & {\footnotesize{}{}$\mathbf{67.5}$}  & {\footnotesize{}{}$\mathbf{75.7}$}  & {\footnotesize{}{}$41.3$}\tabularnewline

&{\footnotesize{}{}\textbf{Ours (Fine-t. Box2Seg)}}  & {\footnotesize{}{}\ding{51}}  & {\footnotesize{}{}\ding{51}}  & {\footnotesize{}{}$62.1$}  & {\footnotesize{}{}$57.3$}  & {\footnotesize{}{}$66.9$}  & {\footnotesize{}{}$\mathbf{1.48}$}\tabularnewline

&{\footnotesize{}{}OnAVOS \cite{voigtlaender17BMVC}}  & {\footnotesize{}{}\ding{51}}  & {\footnotesize{}{}\ding{51}}  & {\footnotesize{}{}$56.5$}  & {\footnotesize{}{}$53.4$}  & {\footnotesize{}{}$59.6$}  & {\footnotesize{}{}$39$}\tabularnewline

\bottomrule

\end{tabular}} {\footnotesize{}{}\caption{\label{tab:results-davis17-testdev}Quantitative results on the DAVIS
2017 test-dev set. FT denotes fine-tuning, M denotes using the first-frame
masks, and t(s) denotes time per frame in seconds. $\dagger$:
timing extrapolated from DAVIS 2016 assuming linear scaling in the
number of objects. Ours
(fastest) denotes Siam R-CNN with ResNet-50 backbone,
half input resolution, and 100 RoIs from the RPN.}
} 
\end{table}

Table~\ref{tab:results-davis17-supp} is an extended version of the results on the DAVIS 2017 \cite{DAVIS2017} validation set shown in the main paper. 

Table~\ref{tab:results-davis16} shows results on the DAVIS 2016 validation set \cite{DAVIS2016} (20 videos, 68.8 frames average length, 1 object per video) compared to 14  state-of-the-art methods. Methods are ranked by the mean of $\mathcal{J}$ and $\mathcal{F}$.
 Among methods which only use the first-frame bounding box (without the mask), Siam R-CNN achieves the strongest result with $78.6\%$ $\mathcal{J}\&\mathcal{F}$, which is $8.8$ percentage points higher than SiamMask \cite{Wang19CVPR}. When fine-tuning Box2Seg, our method achieves $87.1\%$ $\mathcal{J}\&\mathcal{F}$, which is close to the best result on DAVIS 2016 by STM-VOS \cite{Oh19ICCV} with $89.3\%$.

Table~\ref{tab:results-davis17-testdev} shows results on the DAVIS 2017 \cite{DAVIS2017} test-dev set (30 videos, 67.9 frames average length, 2.97 objects per video on average) compared to six state-of-the-art methods. Siam R-CNN achieves $53.3\%$ $\mathcal{J}\&\mathcal{F}$, which is more than $10$ percentage points higher than the result of SiamMask \cite{Wang19CVPR}. STM-VOS \cite{Oh19ICCV} performs significantly better with $72.3\%$, however it relies on the first-frame mask which makes it less usable in practice.

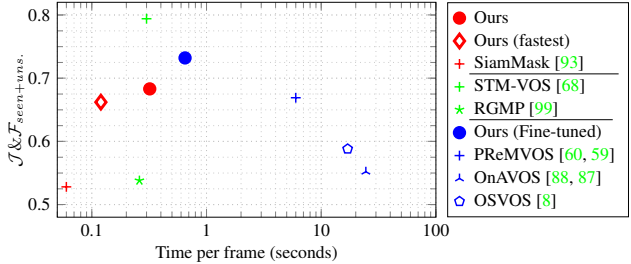
\begin{figure}
\centering
\resizebox{\linewidth}{!}{\begin{tikzpicture}[/pgfplots/width=1\linewidth, /pgfplots/height=0.65\linewidth, /pgfplots/legend pos=south east]
    \begin{axis}[ymin=0.48,ymax=0.82,xmin=0.05,xmax=100,enlargelimits=false,
        xlabel=Time per frame (seconds),
        ylabel=$\mathcal{J}\&\mathcal{F}_{seen+uns.}$,
		font=\small,%
        grid=both,
		grid style=dotted,
        xlabel shift={-2pt},
        ylabel shift={-5pt},
        xmode=log,
        legend columns=1,
        minor ytick={0,0.025,...,1.1},
        ytick={0,0.1,...,1.1},
	    yticklabels={0,0.1,0.2,0.3,0.4,0.5,0.6,0.7,0.8,0.9,1},
	    xticklabels={0.01,0.1,1,10,100},
        legend pos= outer north east,
        legend cell align={left}
        ]

	\addplot[red,mark=*,only marks,line width=0.75, mark size=3.0] coordinates{(0.32,0.683)};
        \addlegendentry{\hphantom{i}Ours}

	\addplot[red,mark=diamond,only marks,line width=1.5, mark size=3.5] coordinates{(0.12,0.662)};
        \addlegendentry{\hphantom{i}Ours (fastest)}
        
        \addplot[red,mark=+, only marks, line width=0.75, mark size=2.5] coordinates{(0.06,0.528)};
        \addlegendentry{\underline{\hphantom{i}SiamMask \cite{Wang19CVPR}\phantom{iiiiiii}}}        

        \addplot[green,mark=+, only marks, line width=0.75, mark size=2.5] coordinates{(0.3,0.794)};
        \addlegendentry{\hphantom{i}STM-VOS \cite{Oh19ICCV}}
        
        \addplot[green,mark=star, only marks, line width=0.75, mark size=2.5] coordinates{(0.26,0.538)};
        \addlegendentry{\underline{\hphantom{i}RGMP \cite{Oh18CVPR}\phantom{iiiiiiiiiii}}}
        
   	\addplot[blue,mark=*,only marks,line width=0.75, mark size=3.0] coordinates{(0.65,0.732)};
        \addlegendentry{\hphantom{i}Ours (Fine-tuned)}
        
        \addplot[blue,mark=+,only marks,line width=0.75, mark size=2.5] coordinates{(6,0.669)};
        \addlegendentry{\hphantom{i}PReMVOS \cite{Luiten18ACCV, Luiten18ECCVW}}
        
        \addplot[blue,mark=Mercedes star,only marks,line width=0.75, mark size=2.5] coordinates{(24.5,	0.552)};
        \addlegendentry{\hphantom{i}OnAVOS \cite{voigtlaender17BMVC, voigtlaender17DAVIS}}
        
        \addplot[blue,mark=pentagon, only marks, line width=0.75, mark size=2.5] coordinates{(17,0.588)};
        \addlegendentry{\hphantom{i}OSVOS \cite{OSVOS}}       
        
    \end{axis}
\end{tikzpicture}}
\vspace{-5mm}
   \caption{Quality versus timing on the YouTube-VOS 2018 \cite{Xu18ECCV} validation set. Only SiamMask \cite{Wang19CVPR} and our method (red) are able to work without the ground truth mask of the first frame and require just the bounding box. Methods shown in blue fine-tune on the first-frame mask. Ours (fastest) denotes Siam R-CNN with ResNet-50 backbone, half input resolution, and 100 RoIs from the RPN%
.}
   \label{fig:speedplot-ytbvos}
   \vspace{-1mm}
\end{figure}

Fig.~\ref{fig:speedplot-ytbvos} shows the speed-accuracy tradeoff of different methods for the YouTube-VOS 2018 validation set. Again, Siam R-CNN achieves a good speed/accuracy trade-off which is only beaten by STM-VOS \cite{Oh19ICCV} (which relies on the first-frame mask).

\subsection{Further Qualitative Evaluation}
\begin{figure*}[t]
  \includegraphics[width=\textwidth]{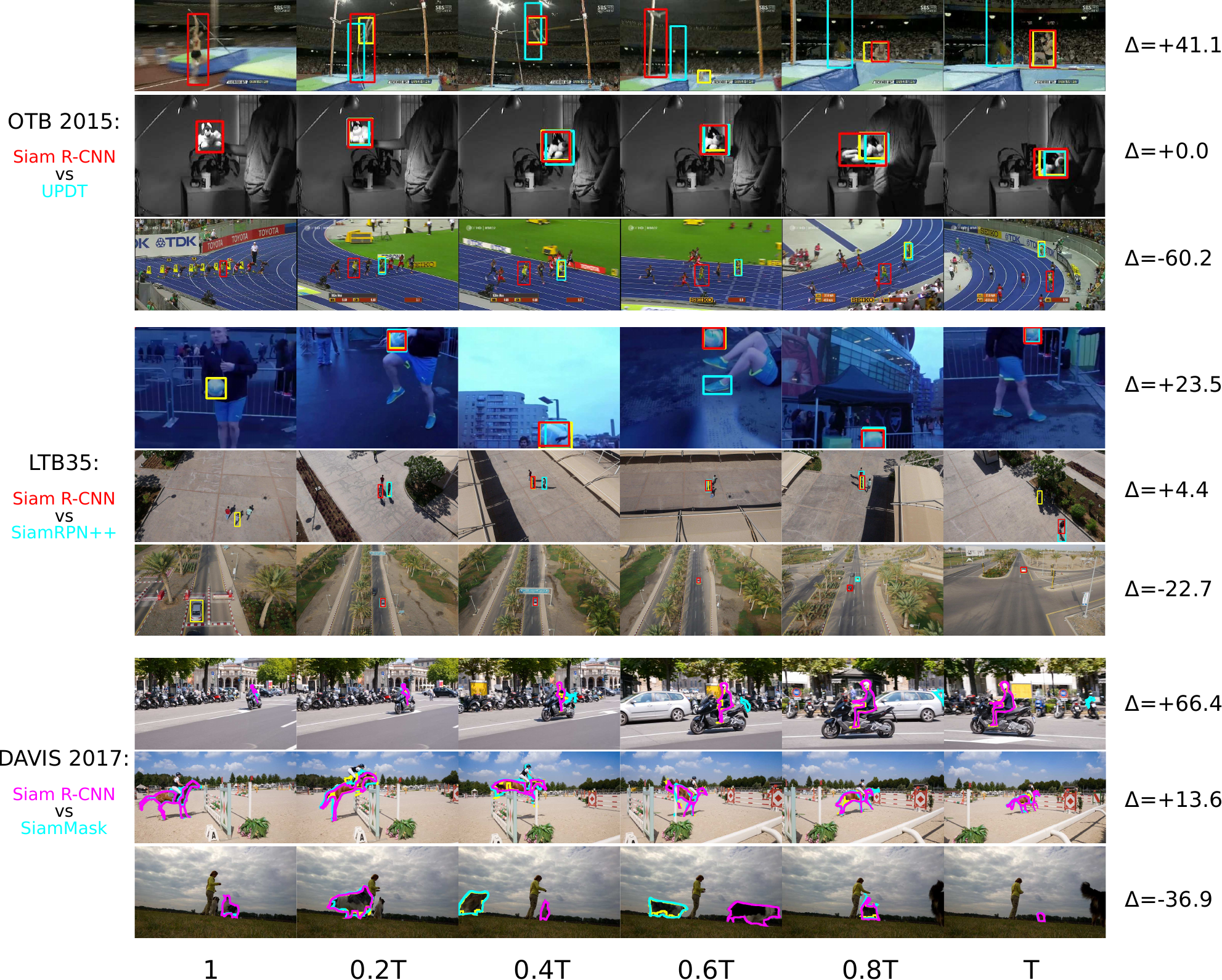}
  \caption{Qualitative results on OTB2015 \cite{Wu15TPAMI}, LTB35 \cite{lukevzivc2018now}, and DAVIS 2017 \cite{DAVIS2017} (validation set). We compare the results of Siam R-CNN to the best competing methods, which are UPDT \cite{Bhat18ECCV} for OTB 2015, SiamRPN++ \cite{Li19CVPR} for LTB35, and SiamMask \cite{Wang19CVPR} for DAVIS 2017. Siam R-CNN's result is shown in red (magenta for DAVIS 2017), the competing methods' results are shown in blue, and the ground truth in yellow. For each benchmark, the sequences with the best, worst and median relative performance ($\Delta$) between Siam R-CNN and the competing method are shown. Six frames spaced equally throughout each video are shown.}
  \label{fig:geil-ims}
\end{figure*}

In Figure \ref{fig:geil-ims} we present further qualitative results of our method on the OTB2015, LTB35 and DAVIS 2017 benchmarks. We present results of our method compared to the best competing method. We show sequences for which Siam R-CNN has the best and worst relative performance compared to the competing method, as well as the sequence with the median relative performance. 

\subsection{Thorough Comparison to Previous Methods}
Throughout the main paper and supplemental material we presented results on $11$ short-term tracking benchmarks and four long-term tracking benchmarks, however these comparisons are spread throughout a number of tables and figures. We provide a unified and thorough comparison of our results to previous methods across all of these benchmarks in Table \ref{tab:big_research}. We compare to the results of every paper that presents comparable tracking results from major vision conferences in 2018 and 2019. As well as including all results from all of these papers we also present additional results from some methods that were either taken from later papers or that we obtained by evaluating open-source code. Sometimes these additional results were different to those presented in the original papers, in which case both results are shown. By evaluating on all of these datasets, and comparing to all methods from these two years, we are able to present a complete and holistic evaluation of our method compared to previous works.

Our method outperforms all previous methods on six out of the $11$ evaluated short-term tracking benchmarks, sometimes by up to $7.2$ percentage points. On the remaining five benchmarks we achieve close to the best results, with only a few previous methods obtaining better results, and by not too large a margin. 

For long-term tracking, Siam R-CNN performs extremely well. Siam R-CNN outperforms all previous methods over all four benchmarks by between $3.9$ and $10.1$ percentage points.

\begin{table*}[t]
\centering{}%
\setlength{\tabcolsep}{1.5pt}
\renewcommand{\arraystretch}{0.4}
\scriptsize

\begin{tabular}{cccccccccccccccccc}
\cmidrule{3-13} \cmidrule{4-13} \cmidrule{5-13} \cmidrule{6-13} \cmidrule{7-13} \cmidrule{8-13} \cmidrule{9-13} \cmidrule{10-13} \cmidrule{11-13} \cmidrule{12-13} \cmidrule{13-13} \cmidrule{15-18} \cmidrule{16-18} \cmidrule{17-18} \cmidrule{18-18} 
 &  & \multicolumn{11}{c}{Short-Term Tracking} &  & \multicolumn{4}{c}{Long-Term Tracking}\tabularnewline
\cmidrule{3-13} \cmidrule{4-13} \cmidrule{5-13} \cmidrule{6-13} \cmidrule{7-13} \cmidrule{8-13} \cmidrule{9-13} \cmidrule{10-13} \cmidrule{11-13} \cmidrule{12-13} \cmidrule{13-13} \cmidrule{15-18} \cmidrule{16-18} \cmidrule{17-18} \cmidrule{18-18} 
 &  & Track.Net & GOT10k & NfS & VOT15 & OTB50 & OTB15 & UAV123 & VOT16 & OTB13 & TC128 & VOT17/18 &  & OxUVA & LaSOT & UAV20L & LTB35\tabularnewline
 &  & AUC & AUC & AUC & EAO & AUC & AUC & AUC & EAO & AUC & AUC & EAO &  & maxGM & AUC & AUC & F\tabularnewline
\cmidrule{1-13} \cmidrule{2-13} \cmidrule{3-13} \cmidrule{4-13} \cmidrule{5-13} \cmidrule{6-13} \cmidrule{7-13} \cmidrule{8-13} \cmidrule{9-13} \cmidrule{10-13} \cmidrule{11-13} \cmidrule{12-13} \cmidrule{13-13} \cmidrule{15-18} \cmidrule{16-18} \cmidrule{17-18} \cmidrule{18-18} 
\multirow{2}{*}{Ours} & $\Delta$ to SOTA & +7.2 & +3.8 & +1.9 & +0.8 & +0.6 & +0.0 & -0.5 & -1.6 & -1.8 & -2.1 & -3.2 &  & +10.1 & +7.9 & +5.5 & +3.9\tabularnewline
 & Siam R-CNN & \textcolor{red}{81.2} & \textcolor{red}{64.9}$^{^{*}}$ & \textcolor{red}{63.9} & \textcolor{red}{45.4}$^{\dagger}$ & \textcolor{red}{66.3} & \textcolor{red}{70.1} & \textcolor{green}{64.9} & \textcolor{green}{46.5}$^{\dagger}$ & 70.4 & \textcolor{black}{60.1} & \textcolor{blue}{40.8}$^{\dagger}$ &  & \textcolor{red}{72.3} & \textcolor{red}{64.8} & \textcolor{red}{67.2} & \textcolor{red}{66.8}\tabularnewline
\midrule
 & DiMP \cite{Bhat19ICCV} & \textcolor{green}{74.0} & \textcolor{green}{61.1} & \textcolor{green}{62.0} &  &  & 68.4 & \textcolor{red}{65.4} &  &  &  & \textcolor{red}{44.0} &  &  & \textcolor{green}{56.9 / 56.8}$^{\mathsection}$ &  & \tabularnewline
\cmidrule{2-18} \cmidrule{3-18} \cmidrule{4-18} \cmidrule{5-18} \cmidrule{6-18} \cmidrule{7-18} \cmidrule{8-18} \cmidrule{9-18} \cmidrule{10-18} \cmidrule{11-18} \cmidrule{12-18} \cmidrule{13-18} \cmidrule{14-18} \cmidrule{15-18} \cmidrule{16-18} \cmidrule{17-18} \cmidrule{18-18} 
 & UpdateNet \cite{Zhang19ICCV} & 67.7 &  &  &  &  &  &  & \textcolor{red}{48.1} &  &  & 39.3 &  &  & 47.5 &  & \tabularnewline
\cmidrule{2-18} \cmidrule{3-18} \cmidrule{4-18} \cmidrule{5-18} \cmidrule{6-18} \cmidrule{7-18} \cmidrule{8-18} \cmidrule{9-18} \cmidrule{10-18} \cmidrule{11-18} \cmidrule{12-18} \cmidrule{13-18} \cmidrule{14-18} \cmidrule{15-18} \cmidrule{16-18} \cmidrule{17-18} \cmidrule{18-18} 
 & GFS-DCF \cite{Xu19ICCV} & 60.9 &  &  &  &  & 69.3 &  &  & \textcolor{red}{72.2} &  &  &  &  &  &  & \tabularnewline
\cmidrule{2-18} \cmidrule{3-18} \cmidrule{4-18} \cmidrule{5-18} \cmidrule{6-18} \cmidrule{7-18} \cmidrule{8-18} \cmidrule{9-18} \cmidrule{10-18} \cmidrule{11-18} \cmidrule{12-18} \cmidrule{13-18} \cmidrule{14-18} \cmidrule{15-18} \cmidrule{16-18} \cmidrule{17-18} \cmidrule{18-18} 
ICCV & SPLT \cite{Yan19ICCV} &  &  &  &  &  &  &  &  &  &  &  &  & \textcolor{green}{62.2} &  &  & \textcolor{blue}{61.6}\tabularnewline
\cmidrule{2-18} \cmidrule{3-18} \cmidrule{4-18} \cmidrule{5-18} \cmidrule{6-18} \cmidrule{7-18} \cmidrule{8-18} \cmidrule{9-18} \cmidrule{10-18} \cmidrule{11-18} \cmidrule{12-18} \cmidrule{13-18} \cmidrule{14-18} \cmidrule{15-18} \cmidrule{16-18} \cmidrule{17-18} \cmidrule{18-18} 
2019 & fdKCF \cite{Zheng19ICCV} &  &  &  &  &  & 67.5 &  & 34.7 & 70.5 &  & 26.5 &  &  &  &  & \tabularnewline
\cmidrule{2-18} \cmidrule{3-18} \cmidrule{4-18} \cmidrule{5-18} \cmidrule{6-18} \cmidrule{7-18} \cmidrule{8-18} \cmidrule{9-18} \cmidrule{10-18} \cmidrule{11-18} \cmidrule{12-18} \cmidrule{13-18} \cmidrule{14-18} \cmidrule{15-18} \cmidrule{16-18} \cmidrule{17-18} \cmidrule{18-18} 
 & Bridging \cite{Huang19ICCV} &  &  & 51.5 &  &  & 64.7 & 58.6 &  & 65.6 &  &  &  &  &  &  & \tabularnewline
\cmidrule{2-18} \cmidrule{3-18} \cmidrule{4-18} \cmidrule{5-18} \cmidrule{6-18} \cmidrule{7-18} \cmidrule{8-18} \cmidrule{9-18} \cmidrule{10-18} \cmidrule{11-18} \cmidrule{12-18} \cmidrule{13-18} \cmidrule{14-18} \cmidrule{15-18} \cmidrule{16-18} \cmidrule{17-18} \cmidrule{18-18} 
 & GradNet \cite{Li19ICCV} &  &  &  &  &  & 63.9 &  &  &  & 55.6 & 24.7 &  &  & 36.5 &  & \tabularnewline
\cmidrule{2-18} \cmidrule{3-18} \cmidrule{4-18} \cmidrule{5-18} \cmidrule{6-18} \cmidrule{7-18} \cmidrule{8-18} \cmidrule{9-18} \cmidrule{10-18} \cmidrule{11-18} \cmidrule{12-18} \cmidrule{13-18} \cmidrule{14-18} \cmidrule{15-18} \cmidrule{16-18} \cmidrule{17-18} \cmidrule{18-18} 
 & MLT \cite{Choi19ICCV} &  &  &  &  &  & 61.1 &  &  & 62.1 &  &  &  &  & 36.8 &  & \tabularnewline
\cmidrule{2-18} \cmidrule{3-18} \cmidrule{4-18} \cmidrule{5-18} \cmidrule{6-18} \cmidrule{7-18} \cmidrule{8-18} \cmidrule{9-18} \cmidrule{10-18} \cmidrule{11-18} \cmidrule{12-18} \cmidrule{13-18} \cmidrule{14-18} \cmidrule{15-18} \cmidrule{16-18} \cmidrule{17-18} \cmidrule{18-18} 
 & ARCF \cite{Huang19ICCVAberrance} &  &  &  &  &  &  & 47.3 &  &  &  &  &  &  &  &  & \tabularnewline
\midrule
 & SiamRPN++ \cite{Li19CVPR} & \textcolor{blue}{73.3} & 45.4$^{\mathsection}$ &  &  &  & 69.6 & 61.3 / 64.2$^{\mathsection}$ &  &  &  & \textcolor{green}{41.4} &  &  & 49.6 &  & \textcolor{green}{62.9}\tabularnewline
\cmidrule{2-18} \cmidrule{3-18} \cmidrule{4-18} \cmidrule{5-18} \cmidrule{6-18} \cmidrule{7-18} \cmidrule{8-18} \cmidrule{9-18} \cmidrule{10-18} \cmidrule{11-18} \cmidrule{12-18} \cmidrule{13-18} \cmidrule{14-18} \cmidrule{15-18} \cmidrule{16-18} \cmidrule{17-18} \cmidrule{18-18} 
 & ATOM \cite{Danelljan19CVPR} & 70.3 &  & \textcolor{blue}{59 / 58.4}$^{\mathsection}$ &  &  & 67.1$^{\mathsection}$ & \textcolor{green}{65 / 64.3}$^{\mathsection}$ &  &  &  & 40.1 &  &  & \textcolor{blue}{51.5 / 51.4}$^{\mathsection}$ &  & \tabularnewline
\cmidrule{2-18} \cmidrule{3-18} \cmidrule{4-18} \cmidrule{5-18} \cmidrule{6-18} \cmidrule{7-18} \cmidrule{8-18} \cmidrule{9-18} \cmidrule{10-18} \cmidrule{11-18} \cmidrule{12-18} \cmidrule{13-18} \cmidrule{14-18} \cmidrule{15-18} \cmidrule{16-18} \cmidrule{17-18} \cmidrule{18-18} 
 & ASRCF \cite{Dai19CVPR} &  &  &  &  &  & 69.2 &  & 39.1 &  & \textcolor{green}{60.3} & 32.8 &  &  & 35.9 &  & \tabularnewline
\cmidrule{2-18} \cmidrule{3-18} \cmidrule{4-18} \cmidrule{5-18} \cmidrule{6-18} \cmidrule{7-18} \cmidrule{8-18} \cmidrule{9-18} \cmidrule{10-18} \cmidrule{11-18} \cmidrule{12-18} \cmidrule{13-18} \cmidrule{14-18} \cmidrule{15-18} \cmidrule{16-18} \cmidrule{17-18} \cmidrule{18-18} 
 & SPM \cite{wang2019spm} &  & \textcolor{blue}{51.3}$^{\mathsection}$ &  &  & \textcolor{blue}{65.3} & 68.7 &  & \textcolor{black}{43.4} & 69.3 &  & 33.8 &  &  &  &  & \tabularnewline
\cmidrule{2-18} \cmidrule{3-18} \cmidrule{4-18} \cmidrule{5-18} \cmidrule{6-18} \cmidrule{7-18} \cmidrule{8-18} \cmidrule{9-18} \cmidrule{10-18} \cmidrule{11-18} \cmidrule{12-18} \cmidrule{13-18} \cmidrule{14-18} \cmidrule{15-18} \cmidrule{16-18} \cmidrule{17-18} \cmidrule{18-18} 
CVPR & SiamMask \cite{Wang19CVPR} &  &  &  &  &  &  &  & \textcolor{blue}{44.2} &  &  & 38.7 &  &  &  &  & \tabularnewline
\cmidrule{2-18} \cmidrule{3-18} \cmidrule{4-18} \cmidrule{5-18} \cmidrule{6-18} \cmidrule{7-18} \cmidrule{8-18} \cmidrule{9-18} \cmidrule{10-18} \cmidrule{11-18} \cmidrule{12-18} \cmidrule{13-18} \cmidrule{14-18} \cmidrule{15-18} \cmidrule{16-18} \cmidrule{17-18} \cmidrule{18-18} 
2019 & SiamDW \cite{Zhang19CVPR} &  &  &  & \textcolor{blue}{38.0} &  & 67.3 &  & 37.0 & 66.6 &  & 30.1 &  &  &  &  & \tabularnewline
\cmidrule{2-18} \cmidrule{3-18} \cmidrule{4-18} \cmidrule{5-18} \cmidrule{6-18} \cmidrule{7-18} \cmidrule{8-18} \cmidrule{9-18} \cmidrule{10-18} \cmidrule{11-18} \cmidrule{12-18} \cmidrule{13-18} \cmidrule{14-18} \cmidrule{15-18} \cmidrule{16-18} \cmidrule{17-18} \cmidrule{18-18} 
 & RPCF \cite{Sun19CVPR} &  &  &  &  &  & 69.6 &  &  & 71.3 &  & 31.6 &  &  &  &  & \tabularnewline
\cmidrule{2-18} \cmidrule{3-18} \cmidrule{4-18} \cmidrule{5-18} \cmidrule{6-18} \cmidrule{7-18} \cmidrule{8-18} \cmidrule{9-18} \cmidrule{10-18} \cmidrule{11-18} \cmidrule{12-18} \cmidrule{13-18} \cmidrule{14-18} \cmidrule{15-18} \cmidrule{16-18} \cmidrule{17-18} \cmidrule{18-18} 
 & C-RPN \cite{Fan19CVPRCRPN} & 66.9 &  &  &  &  & 66.3 &  & 36.3 & 67.5 &  & 28.9 &  &  & 45.5 &  & \tabularnewline
\cmidrule{2-18} \cmidrule{3-18} \cmidrule{4-18} \cmidrule{5-18} \cmidrule{6-18} \cmidrule{7-18} \cmidrule{8-18} \cmidrule{9-18} \cmidrule{10-18} \cmidrule{11-18} \cmidrule{12-18} \cmidrule{13-18} \cmidrule{14-18} \cmidrule{15-18} \cmidrule{16-18} \cmidrule{17-18} \cmidrule{18-18} 
 & TADT \cite{Li19CVPR_TargetAware} &  &  &  & 32.7 &  & 66.0 &  & 29.9 & 68.0 & \textcolor{black}{56.2} &  &  &  &  &  & \tabularnewline
\cmidrule{2-18} \cmidrule{3-18} \cmidrule{4-18} \cmidrule{5-18} \cmidrule{6-18} \cmidrule{7-18} \cmidrule{8-18} \cmidrule{9-18} \cmidrule{10-18} \cmidrule{11-18} \cmidrule{12-18} \cmidrule{13-18} \cmidrule{14-18} \cmidrule{15-18} \cmidrule{16-18} \cmidrule{17-18} \cmidrule{18-18} 
 & GCT \cite{Gao19CVPR} &  &  &  &  &  & 64.8 & 50.8 &  & 67.0 &  & 27.4 &  &  &  &  & \tabularnewline
\cmidrule{2-18} \cmidrule{3-18} \cmidrule{4-18} \cmidrule{5-18} \cmidrule{6-18} \cmidrule{7-18} \cmidrule{8-18} \cmidrule{9-18} \cmidrule{10-18} \cmidrule{11-18} \cmidrule{12-18} \cmidrule{13-18} \cmidrule{14-18} \cmidrule{15-18} \cmidrule{16-18} \cmidrule{17-18} \cmidrule{18-18} 
 & UDT \cite{Wang19CVPR_DeepTracking} &  &  &  &  &  & 63.2 &  & 30.1 &  & 54.1 &  &  &  &  &  & \tabularnewline
\midrule
 & UPDT \cite{Bhat18ECCV} & 61.1$^{\mathsection}$ &  & 54.1 / 53.7$^{\mathsection}$ &  &  & \textcolor{red}{70.1}$^{\mathsection}$ & 55 / 54.7$^{\mathsection}$ &  &  & \textcolor{red}{62.2} & 37.8 &  &  &  &  & \tabularnewline
\cmidrule{2-18} \cmidrule{3-18} \cmidrule{4-18} \cmidrule{5-18} \cmidrule{6-18} \cmidrule{7-18} \cmidrule{8-18} \cmidrule{9-18} \cmidrule{10-18} \cmidrule{11-18} \cmidrule{12-18} \cmidrule{13-18} \cmidrule{14-18} \cmidrule{15-18} \cmidrule{16-18} \cmidrule{17-18} \cmidrule{18-18} 
 & DaSiamRPN \cite{Zhu18ECCV} & 63.8$^{\mathsection}$ &  &  & \textcolor{green}{44.6} &  & 65.8$^{\mathsection}$ & 58.6 / 58.5$^{\mathsection}$ & 41.1 &  &  & 32.6 &  & \textcolor{blue}{41.5}$^{\mathsection}$ &  & \textcolor{green}{61.7} & 60.7$^{\mathsection}$\tabularnewline
\cmidrule{2-18} \cmidrule{3-18} \cmidrule{4-18} \cmidrule{5-18} \cmidrule{6-18} \cmidrule{7-18} \cmidrule{8-18} \cmidrule{9-18} \cmidrule{10-18} \cmidrule{11-18} \cmidrule{12-18} \cmidrule{13-18} \cmidrule{14-18} \cmidrule{15-18} \cmidrule{16-18} \cmidrule{17-18} \cmidrule{18-18} 
 & ACT \cite{Chen18ECCV_Actor} &  &  &  &  & \textcolor{green}{65.7} & 64.3 &  & 27.5 & 66.3 &  &  &  &  &  &  & \tabularnewline
\cmidrule{2-18} \cmidrule{3-18} \cmidrule{4-18} \cmidrule{5-18} \cmidrule{6-18} \cmidrule{7-18} \cmidrule{8-18} \cmidrule{9-18} \cmidrule{10-18} \cmidrule{11-18} \cmidrule{12-18} \cmidrule{13-18} \cmidrule{14-18} \cmidrule{15-18} \cmidrule{16-18} \cmidrule{17-18} \cmidrule{18-18} 
 & RTINet \cite{Yao18ECCV} &  &  &  &  & 63.7 & 68.2 &  & 29.8 &  & \textcolor{blue}{60.2} &  &  &  &  &  & \tabularnewline
\cmidrule{2-18} \cmidrule{3-18} \cmidrule{4-18} \cmidrule{5-18} \cmidrule{6-18} \cmidrule{7-18} \cmidrule{8-18} \cmidrule{9-18} \cmidrule{10-18} \cmidrule{11-18} \cmidrule{12-18} \cmidrule{13-18} \cmidrule{14-18} \cmidrule{15-18} \cmidrule{16-18} \cmidrule{17-18} \cmidrule{18-18} 
 & SACF \cite{Zhang18ECCV_Aligned} &  &  &  & 34.3 &  & 69.3 &  & 38.0 & 71.3 &  &  &  &  &  &  & \tabularnewline
\cmidrule{2-18} \cmidrule{3-18} \cmidrule{4-18} \cmidrule{5-18} \cmidrule{6-18} \cmidrule{7-18} \cmidrule{8-18} \cmidrule{9-18} \cmidrule{10-18} \cmidrule{11-18} \cmidrule{12-18} \cmidrule{13-18} \cmidrule{14-18} \cmidrule{15-18} \cmidrule{16-18} \cmidrule{17-18} \cmidrule{18-18} 
ECCV & DRL-IS \cite{Ren18ECCV} &  &  &  &  &  & 67.1 &  &  &  & 59.0 &  &  &  &  &  & \tabularnewline
\cmidrule{2-18} \cmidrule{3-18} \cmidrule{4-18} \cmidrule{5-18} \cmidrule{6-18} \cmidrule{7-18} \cmidrule{8-18} \cmidrule{9-18} \cmidrule{10-18} \cmidrule{11-18} \cmidrule{12-18} \cmidrule{13-18} \cmidrule{14-18} \cmidrule{15-18} \cmidrule{16-18} \cmidrule{17-18} \cmidrule{18-18} 
2018 & DSLT \cite{Lu18ECCV} &  &  &  &  &  & 66.0 & 53.0 & 33.2 & 68.3 & 58.7 &  &  &  &  &  & \tabularnewline
\cmidrule{2-18} \cmidrule{3-18} \cmidrule{4-18} \cmidrule{5-18} \cmidrule{6-18} \cmidrule{7-18} \cmidrule{8-18} \cmidrule{9-18} \cmidrule{10-18} \cmidrule{11-18} \cmidrule{12-18} \cmidrule{13-18} \cmidrule{14-18} \cmidrule{15-18} \cmidrule{16-18} \cmidrule{17-18} \cmidrule{18-18} 
 & Meta-Tracker \cite{Park18ECCV} &  &  &  &  &  & 66.2 &  & 31.7 &  &  &  &  &  &  &  & \tabularnewline
\cmidrule{2-18} \cmidrule{3-18} \cmidrule{4-18} \cmidrule{5-18} \cmidrule{6-18} \cmidrule{7-18} \cmidrule{8-18} \cmidrule{9-18} \cmidrule{10-18} \cmidrule{11-18} \cmidrule{12-18} \cmidrule{13-18} \cmidrule{14-18} \cmidrule{15-18} \cmidrule{16-18} \cmidrule{17-18} \cmidrule{18-18} 
 & RT-MDNet \cite{Jung18ECCV} &  &  &  &  &  & 65.0 & 53.5 &  &  & 56.3 &  &  &  &  &  & \tabularnewline
\cmidrule{2-18} \cmidrule{3-18} \cmidrule{4-18} \cmidrule{5-18} \cmidrule{6-18} \cmidrule{7-18} \cmidrule{8-18} \cmidrule{9-18} \cmidrule{10-18} \cmidrule{11-18} \cmidrule{12-18} \cmidrule{13-18} \cmidrule{14-18} \cmidrule{15-18} \cmidrule{16-18} \cmidrule{17-18} \cmidrule{18-18} 
 & MemTrack \cite{Yang18ECCV} &  &  &  &  &  & 62.6 &  &  & 64.2 &  &  &  &  &  &  & \tabularnewline
\cmidrule{2-18} \cmidrule{3-18} \cmidrule{4-18} \cmidrule{5-18} \cmidrule{6-18} \cmidrule{7-18} \cmidrule{8-18} \cmidrule{9-18} \cmidrule{10-18} \cmidrule{11-18} \cmidrule{12-18} \cmidrule{13-18} \cmidrule{14-18} \cmidrule{15-18} \cmidrule{16-18} \cmidrule{17-18} \cmidrule{18-18} 
 & StructSiam \cite{Zhang18ECCV_Structured} &  &  &  &  &  & 62.1 &  & 26.4 & 63.8 &  &  &  &  & 33.5$^{\mathsection}$ &  & \tabularnewline
\cmidrule{2-18} \cmidrule{3-18} \cmidrule{4-18} \cmidrule{5-18} \cmidrule{6-18} \cmidrule{7-18} \cmidrule{8-18} \cmidrule{9-18} \cmidrule{10-18} \cmidrule{11-18} \cmidrule{12-18} \cmidrule{13-18} \cmidrule{14-18} \cmidrule{15-18} \cmidrule{16-18} \cmidrule{17-18} \cmidrule{18-18} 
 & SiamFC-tri \cite{Dong18ECCV} &  &  &  &  & 53.5 & 59.2 &  &  & 62.9 &  & 21.3 &  &  &  &  & \tabularnewline
\midrule
 & DRT \cite{Sun18CVPR_Reliability} &  &  &  &  &  & \textcolor{blue}{69.9} &  & \textcolor{blue}{44.2} & \textcolor{green}{72.0} &  &  &  &  &  &  & \tabularnewline
\cmidrule{2-18} \cmidrule{3-18} \cmidrule{4-18} \cmidrule{5-18} \cmidrule{6-18} \cmidrule{7-18} \cmidrule{8-18} \cmidrule{9-18} \cmidrule{10-18} \cmidrule{11-18} \cmidrule{12-18} \cmidrule{13-18} \cmidrule{14-18} \cmidrule{15-18} \cmidrule{16-18} \cmidrule{17-18} \cmidrule{18-18} 
 & MCCT \cite{Wang18CVPR} &  &  &  &  &  & 69.5 &  & 39.3 & \textcolor{blue}{71.4} & 59.6 &  &  &  &  &  & \tabularnewline
\cmidrule{2-18} \cmidrule{3-18} \cmidrule{4-18} \cmidrule{5-18} \cmidrule{6-18} \cmidrule{7-18} \cmidrule{8-18} \cmidrule{9-18} \cmidrule{10-18} \cmidrule{11-18} \cmidrule{12-18} \cmidrule{13-18} \cmidrule{14-18} \cmidrule{15-18} \cmidrule{16-18} \cmidrule{17-18} \cmidrule{18-18} 
 & SiamRPN \cite{Li18CVPRSiamRPN} &  &  &  & 35.8 &  & 63.7 &  & 34.4 &  &  &  &  &  &  & \textcolor{blue}{45.4}$^{\mathsection}$ & \tabularnewline
\cmidrule{2-18} \cmidrule{3-18} \cmidrule{4-18} \cmidrule{5-18} \cmidrule{6-18} \cmidrule{7-18} \cmidrule{8-18} \cmidrule{9-18} \cmidrule{10-18} \cmidrule{11-18} \cmidrule{12-18} \cmidrule{13-18} \cmidrule{14-18} \cmidrule{15-18} \cmidrule{16-18} \cmidrule{17-18} \cmidrule{18-18} 
 & STRCF \cite{Li18CVPR_Regularized} &  &  &  &  &  & 68.3 &  & 31.3 &  & 60.1 &  &  &  &  &  & \tabularnewline
\cmidrule{2-18} \cmidrule{3-18} \cmidrule{4-18} \cmidrule{5-18} \cmidrule{6-18} \cmidrule{7-18} \cmidrule{8-18} \cmidrule{9-18} \cmidrule{10-18} \cmidrule{11-18} \cmidrule{12-18} \cmidrule{13-18} \cmidrule{14-18} \cmidrule{15-18} \cmidrule{16-18} \cmidrule{17-18} \cmidrule{18-18} 
 & VITAL \cite{Song18CVPR} &  &  &  &  &  & 68.2 &  & 32.3 & 71.0 &  &  &  &  & 39.0$^{\mathsection}$ &  & \tabularnewline
\cmidrule{2-18} \cmidrule{3-18} \cmidrule{4-18} \cmidrule{5-18} \cmidrule{6-18} \cmidrule{7-18} \cmidrule{8-18} \cmidrule{9-18} \cmidrule{10-18} \cmidrule{11-18} \cmidrule{12-18} \cmidrule{13-18} \cmidrule{14-18} \cmidrule{15-18} \cmidrule{16-18} \cmidrule{17-18} \cmidrule{18-18} 
CVPR & LSART \cite{Sun18CVPR_Spatial} &  &  &  &  &  & 67.2 &  &  &  &  & 32.3 &  &  &  &  & \tabularnewline
\cmidrule{2-18} \cmidrule{3-18} \cmidrule{4-18} \cmidrule{5-18} \cmidrule{6-18} \cmidrule{7-18} \cmidrule{8-18} \cmidrule{9-18} \cmidrule{10-18} \cmidrule{11-18} \cmidrule{12-18} \cmidrule{13-18} \cmidrule{14-18} \cmidrule{15-18} \cmidrule{16-18} \cmidrule{17-18} \cmidrule{18-18} 
2018 & FlowTrack \cite{Zhu18CVPR} &  &  &  & 34.1 &  & 65.5 &  & 33.4 & 68.9 &  &  &  &  &  &  & \tabularnewline
\cmidrule{2-18} \cmidrule{3-18} \cmidrule{4-18} \cmidrule{5-18} \cmidrule{6-18} \cmidrule{7-18} \cmidrule{8-18} \cmidrule{9-18} \cmidrule{10-18} \cmidrule{11-18} \cmidrule{12-18} \cmidrule{13-18} \cmidrule{14-18} \cmidrule{15-18} \cmidrule{16-18} \cmidrule{17-18} \cmidrule{18-18} 
 & RASNet \cite{Wang18CVPR_Residual} &  &  &  & 32.7 &  & 64.1 &  &  & 67.0 &  & 28.1 &  &  &  &  & \tabularnewline
\cmidrule{2-18} \cmidrule{3-18} \cmidrule{4-18} \cmidrule{5-18} \cmidrule{6-18} \cmidrule{7-18} \cmidrule{8-18} \cmidrule{9-18} \cmidrule{10-18} \cmidrule{11-18} \cmidrule{12-18} \cmidrule{13-18} \cmidrule{14-18} \cmidrule{15-18} \cmidrule{16-18} \cmidrule{17-18} \cmidrule{18-18} 
 & SA-Siam \cite{He18CVPR_Twofold} &  &  &  & 31.0 & 61.0 & 65.7 &  & 29.1 & 67.7 &  & 23.6 &  &  &  &  & \tabularnewline
\cmidrule{2-18} \cmidrule{3-18} \cmidrule{4-18} \cmidrule{5-18} \cmidrule{6-18} \cmidrule{7-18} \cmidrule{8-18} \cmidrule{9-18} \cmidrule{10-18} \cmidrule{11-18} \cmidrule{12-18} \cmidrule{13-18} \cmidrule{14-18} \cmidrule{15-18} \cmidrule{16-18} \cmidrule{17-18} \cmidrule{18-18} 
 & TRACA \cite{Choi18CVPR} &  &  &  &  &  & 60.3 &  &  & 65.2 &  &  &  &  &  &  & \tabularnewline
\cmidrule{2-18} \cmidrule{3-18} \cmidrule{4-18} \cmidrule{5-18} \cmidrule{6-18} \cmidrule{7-18} \cmidrule{8-18} \cmidrule{9-18} \cmidrule{10-18} \cmidrule{11-18} \cmidrule{12-18} \cmidrule{13-18} \cmidrule{14-18} \cmidrule{15-18} \cmidrule{16-18} \cmidrule{17-18} \cmidrule{18-18} 
 & MKCF \cite{Tang18CVPR} &  &  & 45.5 &  &  &  &  &  & 64.1 &  &  &  &  &  &  & \tabularnewline
\cmidrule{2-18} \cmidrule{3-18} \cmidrule{4-18} \cmidrule{5-18} \cmidrule{6-18} \cmidrule{7-18} \cmidrule{8-18} \cmidrule{9-18} \cmidrule{10-18} \cmidrule{11-18} \cmidrule{12-18} \cmidrule{13-18} \cmidrule{14-18} \cmidrule{15-18} \cmidrule{16-18} \cmidrule{17-18} \cmidrule{18-18} 
 & HP \cite{Dong18CVPR} &  &  &  &  & 55.4 & 60.1 &  &  & 62.9 &  &  &  &  &  &  & \tabularnewline
\cmidrule{2-18} \cmidrule{3-18} \cmidrule{4-18} \cmidrule{5-18} \cmidrule{6-18} \cmidrule{7-18} \cmidrule{8-18} \cmidrule{9-18} \cmidrule{10-18} \cmidrule{11-18} \cmidrule{12-18} \cmidrule{13-18} \cmidrule{14-18} \cmidrule{15-18} \cmidrule{16-18} \cmidrule{17-18} \cmidrule{18-18} 
 & SINT++ \cite{Wang18CVPR_Adversarial} &  &  &  &  & 62.4 & 57.4 &  &  &  &  &  &  &  &  &  & \tabularnewline
\bottomrule
\end{tabular}

\caption{\label{tab:big_research}{Comparison to all trackers published in CVPR, ICCV and ECCV in 2018 and 2019. Results from original papers, except when marked with $^\mathsection$ which are from later papers, or from running open-source code. Results in \{\textcolor{red}{Red}, \textcolor{green}{Green}, \textcolor{blue}{Blue}\} are the \{\textcolor{red}{Best}, \textcolor{green}{Second}, \textcolor{blue}{Third}\}, respectively. Benchmarks are ordered by performance relative to the best method other than ours  ($\Delta$ to SOTA). Methods are ordered first by conference date, then by most `bests', most `seconds', most `thirds' and finally by approximate `head-to-head' performance. \emph{On all benchmarks Siam R-CNN uses exactly the same network weights and tracking hyper-parameters}, except for those marked with $^\dagger$ which use the `short-term' tracking parameters, and those marked with $^*$ which use weights trained only on GOT-10k.}}
\end{table*}

\end{document}